%% file: main.tex
\documentclass[twoside]{article}

\usepackage[preprint]{aistats2027}
\usepackage{graphicx}
\usepackage{amssymb}
\usepackage{booktabs}
\usepackage{longtable}
\usepackage[table]{xcolor}
\usepackage[most]{tcolorbox}
\graphicspath{{figures/controlled_simulations/}{figures/}}
\usepackage[round]{natbib}
\begin{document}

\runningtitle{Rank Reversals under Temporal Nonstationarity}
\runningauthor{Song and Ding}
\setlength{\footskip}{20pt}
\fancyfoot[C]{\thepage}
\thispagestyle{plain}

\twocolumn[
\aistatstitle{Decomposing Rank Reversals under Temporal Nonstationarity: Alignment, Aggressiveness, and Instability}
\aistatsauthor{
  Deying Song
  \And
  Xueying Ding
}
\aistatsaddress{
  Carnegie Mellon University \\
  Pittsburgh, PA, USA
  \And
  Carnegie Mellon University \\
  Pittsburgh, PA, USA
}
]

\begin{abstract}
Under temporal nonstationarity, model performance rankings reverse across training horizons due to heterogeneous sensitivities to memory. 
While existing literature addresses adaptive lookbacks, effective model complexity, and empirical forecast evaluation in isolation, 
no unified framework explains the structural mechanics driving these rank reversals. 
We bridge this gap by deriving a model-agnostic decomposition of out-of-sample $R^2$ 
into target alignment, miscalibrated aggressiveness, and predictive instability. 
Within this framework, we mathematically characterize complexity phase boundaries, 
establish closed-form optimal-memory equations for ridge regression under local drift, 
and extend these guarantees to complex model architectures via surrogate bounds. 
We validate these theoretical dynamics by benchmarking 14 linear, tree-based, and neural architectures across large-scale financial panels. 
In doing so, our framework demonstrates that empirical model rank reversals are driven by predictable surges in miscalibrated aggressiveness, 
providing a rigorous mechanistic diagnosis of generalization under temporal shift.
\end{abstract}

\input{introduction}

\input{related_work}

\input{theory}

\input{experiments}
\input{conclusion}

\section*{AI Use Statement}%
During the preparation of this work, 
the authors used OpenAI ChatGPT-6 Sol, Claude Sonnet 5.5, and Google Gemini-3.1 Pro 
to conduct literature searches, generate baseline code configurations,
develop proofs and derivations, create visualizations,
generate initial drafts of writing, improve grammatical clarity,
check for technical consistency, and verify the correctness of mathematical derivations.
Following the use of these tools, the authors thoroughly reviewed, edited, and verified all generated output. 
The authors take full responsibility for the technical correctness, integrity, and originality of the final manuscript content.

\bibliography{references}

\input{appendix_theory-1}

\input{appendix_simulations}

\input{appendix_main_experiment}

\end{document}

%% file: introduction.tex
\section{Introduction}
\label{sec:introduction}

Reliable forecasting under temporal nonstationarity 
requires balancing the estimation variance of restricted training windows 
against the bias induced by obsolete historical data \citep{dahlhaus1997fitting,pesaran2007selection,giraitis2013adaptive}. 
This tradeoff is exceptionally severe in return prediction, where predictive signals are notoriously weak and transient \citep{welch2008comprehensive,campbell2008predicting,smith2021break}. 
While extensive literature addresses adaptive memory selection, 
systematically explaining why the relative performance of models reverses 
across different training horizons requires formally linking theoretical dynamics with empirical phenomena. 
To establish this mechanistic link, we address two fundamental questions: 
(1) Which specific forecast components drive the performance differential between models at a fixed memory? 
and (2) Can we mathematically predict, directly from the underlying statistical dynamics, the critical memory threshold at which performance rankings reverse?

Although the literature addresses facets of this problem, existing approaches remain disconnected. 
In nonstationary settings, dynamic-regret analyses balance adaptation against data efficiency via sliding windows and forgetting factors, typically assuming known variation budgets \citep{cheung2019learning,zhao2020simple,yuan2020trading,yang2016tracking,zhao2021improved,baby2023second}.
The constraint is recently relaxed by adaptive-lookback estimators \citep{huang2023stability,han2024assessment,baby2025adaptive}. 
Concurrently, statistical learning theory links generalization bounds to effective complexity \citep{caponnetto2007optimal,hsu2012random,dobriban2018high,bousquet2002stability}. 
Finally, empirical forecast evaluation provides diagnostics for bias and calibration \citep{mincer1969evaluation,theil1966applied,gneiting2023regression,waghmare2026proper}, 
alongside statistical tests for predictive ability \citep{diebold1995comparing,giacomini2006tests,white2000reality,hansen2005test,hansen2011model,hewamalage2021look}. 
\textbf{However, these paradigms operate in isolation.} 
Regret and stability bounds do not predict when high-capacity models will surpass simpler baselines under shifting histories, 
while empirical diagnostics quantify performance gaps without isolating their structural drivers. 
Consequently, no existing framework systematically links training memory and model complexity to explain the rank reversals observed in nonstationary environments.

In this work, we close that gap in three steps.
First, we derive a model-agnostic decomposition of out-of-sample $R^2$ that isolates target alignment from penalties for miscalibrated aggressiveness (excess forecast magnitude) and predictive instability (variance across replicates). 
Second, we analytically characterize how the training horizon governs each component, establishing a complexity phase boundary and a closed-form optimal-memory equation for ridge regression under local drift, with extensions to nonlinear architectures via surrogate transfer bounds. 
Third, we validate these theoretical mechanics through controlled simulations and extensive empirical evaluation, benchmarking 14 linear, tree-based, and neural specifications across both macroeconomic industry portfolios \citep{fama1997industry} and a large-scale, out-of-domain equity panel \citep{zhang2026benchmarking}.

In summary, our primary contributions are:

\begin{list}{$\bullet$}{
\setlength{\itemindent}{0pt}
\setlength{\leftmargin}{1.0em} 
\setlength{\itemsep}{5pt}
\setlength{\parsep}{0pt}
\setlength{\parskip}{0pt}
}
\item \textbf{A mechanistic theory of rank reversals.} We provide a unified theoretical framework that mathematically links training memory and model capacity to out-of-sample generalization, formally proving why and when high-capacity models lose their predictive advantage under distribution shift.
\item \textbf{Empirical diagnosis of performance reversals.} Across large-scale financial panels, we demonstrate that when high-capacity models underperform simpler baselines, it is driven by predictable surges in miscalibrated aggressiveness.
\end{list}

%% file: related_work.tex
\section{Related Work}
\label{sec:related-work}

\textbf{Learning under temporal variation.}
Learning under nonstationarity balances rapid adaptation against historical data efficiency. Sliding windows and restarts bound dynamic regret but typically assume known variation budgets \citep{cheung2019learning,zhao2020simple}, 
while forgetting factors trade static and dynamic regret \citep{yuan2020trading,yang2016tracking,zhao2021improved,baby2023second}. 
Recent advances deploy adaptive lookbacks for unknown shifts \citep{huang2023stability,han2024assessment,baby2025adaptive}. 
Despite these methods, the conditions under which a complex model systematically outperforms a simpler baseline across varying histories remain theoretically underspecified. 
Our complexity phase boundary analytically identifies when a model's approximation advantage offsets its heightened vulnerability to memory-dependent variance and drift.

\textbf{Effective complexity and forecast instability.}
Generalization under finite data depends on effective complexity and algorithmic stability. While spectral dimension and signal strength govern ridge prediction risk \citep{caponnetto2007optimal,hsu2012random,dobriban2018high}, 
regularization also controls sensitivity, rendering raw parameter counts inadequate \citep{hsu2012random,bousquet2002stability}. 
Standard risk bounds, however, do not translate this sensitivity into the realized out-of-sample $R^2$ cost induced by variation across forecast replicates. 
Our finite-sample identity bridges this gap by isolating predictive instability as an explicit additive penalty, proving that replicate ensembling eliminates it entirely.

\textbf{Local change and estimation memory.}
Time-varying regressions show that recent data capture drift while longer histories reduce variance \citep{dahlhaus1997fitting,vogt2012nonparametric}, 
framing window selection as an old-data bias versus sampling error tradeoff \citep{pesaran2007selection,giraitis2013adaptive}. 
Forecasting under instability also calls for comparisons that allow relative model performance to change over time \citep{giacomini2015forecasting,giacomini2016model}.
Although these frameworks formalize the bias-variance tension, 
standard heuristics overlook how regularization structurally filters temporal drift across predictor directions to shift the optimal memory horizon. 
Under a local-drift specification, we derive an exact optimal-memory equation that couples spectral effective dimension, filtered drift, and the bias-drift cross-term.

\textbf{Calibration, scale, and combination.}
Classical diagnostics isolate bias and scale, while modern scoring rules separate calibration, discrimination, and uncertainty \citep{mincer1969evaluation,theil1966applied,gneiting2023regression,waghmare2026proper}. 
Forecast combinations similarly handle structural breaks across predictor sets \citep{diebold1987structural,elliott2004optimal,chen2020timevarying}. 
Nevertheless, existing frameworks do not jointly isolate the impacts of weak target alignment, miscalibrated magnitude, and predictive instability on out-of-sample $R^2$. 
Our score identity achieves this tripartite separation analytically, quantifying the oracle $R^2$ gain available solely through scaling forecast aggressiveness.

\textbf{Comparing and ranking forecasts.}
Accuracy tests evaluate unconditional loss differences or map predictive ability to information sets \citep{diebold1995comparing,giacomini2006tests}. 
To counter data-snooping biases, specialized tests and confidence sets correct for multiple comparisons \citep{white2000reality,hansen2005test,hansen2011model}, 
while rank-stability analyses reveal sensitivity to evaluation design \citep{hewamalage2021look}.  
While these frameworks rigorously quantify statistical uncertainty, they are diagnostic rather than explanatory: they detect performance gaps without isolating the driving forecast properties. 
We augment dependence-aware comparisons with our attribution identity---decomposing gaps into alignment, aggressiveness, and instability---to mechanistically explain memory-induced rank reversals.

\textbf{Return prediction and model comparisons.}
Equity premium forecasts routinely fail to outperform historical means, though shrinkage reduces prediction error \citep{welch2008comprehensive,campbell2008predicting}. 
While nonlinear interactions isolate weak signals \citep{goktas2026tempusbench,capponi2025nonstationarity}, 
structural breaks degrade accuracy \citep{gu2020empirical,smith2021break}. 
Broad benchmarks corroborate this predictability limit, showing that transaction costs and market states fundamentally reorder rankings \citep{bejarano2025open,salykaufmann2026deep,zhang2026benchmarking}. 
Because benchmarks often conflate model selection with disparate tuning budgets, it remains unresolved whether training history alone can invert rankings under strictly matched information sets. 
By modulating memory across linear, tree-based, and neural models, we demonstrate that history alone reverses rankings, tracing these shifts directly to target alignment, aggressiveness, and instability.

%% file: theory.tex
\section{Forecast Components and Complexity}
\label{sec:theory-main}

To formalize how training history shapes prediction, 
we introduce a \textbf{model-agnostic decomposition of out-of-sample $R^2$} 
to mathematically isolate the drivers of rank reversal (Sections~\ref{sec:theory-main-decomposition} and~\ref{sec:theory-main-memory}), 
and establish a \textbf{complexity phase boundary} that yields an \textbf{exact optimal-memory equation for ridge regression under local drift} (Section~\ref{sec:theory-main-ridge}) 
alongside explicit \textbf{transfer bounds for nonlinear architectures} (Section~\ref{sec:theory-main-transfer}). 
Throughout this section, we quantify training history via the memory $h$, defined as the number of most recent observations used to train a model. 
Full derivations and proofs for all theoretical results are in Appendix~\ref{sec:theory}.

\begin{figure}[htb]
\centering
\includegraphics[width=0.95\columnwidth]{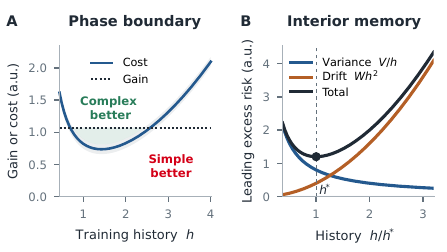}
\caption{Theoretical leading-order dynamics with normalized parameters. 
(A)~The complex class achieves lower risk when its history-independent advantage $\Delta_0$ exceeds the complexity cost $\phi(h)$; bands denote remainder bounds. 
(B)~Variance $V/h$ decays and the drift penalty $Wh^2$ grows with memory $h$, creating an interior risk minimum at $h^*$.}
\label{fig:theory}
\end{figure}

\subsection{Forecast-score decomposition}
\label{sec:theory-main-decomposition}

Let $\mathbf{y} \in \mathbb{R}^n$ represent the true outcomes on a common evaluation sample, assuming a non-zero magnitude $Y_2 = \|\mathbf{y}\|^2 > 0$. 
Suppose a forecasting procedure is fitted $M$ times---varying only the random seeds or training data splits---to produce $M$ replicate forecasts. 
Let $\widehat{\mathbf{y}}_m \in \mathbb{R}^n$ denote the $m$-th replicate, with an ensemble mean $\overline{\mathbf{y}} = \frac{1}{M}\sum_{m=1}^{M} \widehat{\mathbf{y}}_m$.

We evaluate performance using the out-of-sample $R^2$ relative to a naive zero prediction. 
The average score across all replicates is $\overline{R^2} = 1 - \frac{1}{M}\sum_{m=1}^{M} \|\mathbf{y} - \widehat{\mathbf{y}}_m\|^2 / Y_2$. 
To isolate what drives this performance, we characterize the geometry of the forecasts through three components: target alignment $a$, forecast aggressiveness $g$, and predictive instability $u$, defined as
\begin{equation}
a = \frac{\langle\mathbf{y}, \overline{\mathbf{y}}\rangle}{\|\mathbf{y}\|\|\overline{\mathbf{y}}\|}, \quad
g = \frac{\|\overline{\mathbf{y}}\|}{\|\mathbf{y}\|}, \quad
u = \frac{\frac{1}{M}\sum_{m=1}^{M}\|\widehat{\mathbf{y}}_m - \overline{\mathbf{y}}\|^2}{Y_2}.
\label{eq:main-aqu}
\end{equation}
Conceptually, \textbf{alignment $a$ measures direction} (the cosine similarity between the target and the mean forecast), 
\textbf{aggressiveness $g$ measures relative scale} (the overall magnitude of the mean forecast compared to the target), 
and \textbf{instability $u$ measures variance} (the squared dispersion of the individual replicates around their mean, normalized by the target energy $Y_2$).

\paragraph{Replicate-ensemble gain.}
Expanding the squared error and noting that replicate deviations sum to zero ($\sum_{m}(\widehat{\mathbf y}_m-\overline{\mathbf y})=0$) yields an exact decomposition:
\begin{equation}
\overline{R^2}=2ag-g^2-u.
\label{eq:main-identity}
\end{equation}
If $\overline{\mathbf y}=0$, we set $a=0$ by convention.
With this convention, the identity
\(
ag=\langle\mathbf y,\overline{\mathbf y}\rangle / Y_2
\)
holds in all cases.
Performance thus separates into a positive alignment reward ($2ag$), penalized by aggressiveness ($g^2$) and instability ($u$) costs. 
The ensemble mean $\overline{\mathbf y}$ achieves a score of $2ag-g^2$, 
exceeding the average replicate score by $u$. 
Ensembling thus eliminates the instability penalty while preserving the alignment and aggressiveness terms.

\paragraph{Proposition 1 (optimal common scalar).}
Rescaling forecasts $\mathbf{p}_m$ by a factor $\gamma$ yields a mean score $\mathcal{S}(\gamma) = 2\gamma c - \gamma^2 q$, 
where $c = \operatorname{E}_m\langle\mathbf{y}, \mathbf{p}_m\rangle/Y_2$ and $q = \operatorname{E}_m\|\mathbf{p}_m\|^2/Y_2 > 0$. 
On any closed interval $\mathcal{I} \subseteq \mathbb{R}$, the unique maximizer is the projection of $c/q$ onto $\mathcal I$,
\begin{tcolorbox}[
    ams equation,
    colback=gray!5,       
    colframe=blue!50!black,  
    sharp corners,               
    boxrule=0.8pt,          
    boxsep=1pt,             
    left=4pt, right=4pt%
]
\gamma^*_{\mathcal{I}} = \Pi_{\mathcal{I}}(c/q).
\label{eq:main-scalar}
\end{tcolorbox}
\textbf{In essence, this calculates the maximum potential score improvement if the overall magnitude of the predictions is perfectly corrected.}
Unconstrained scaling ($\mathcal{I} = \mathbb{R}$) maximizes the score at $c^2/q$. 
For individual replicates, $\gamma^* = ag/(g^2+u)$. 
For the ensemble mean, $\gamma^* = a/g$. 

\subsection{Complexity phase boundary}
\label{sec:theory-main-memory}

\paragraph{Rank reversal.}
Suppose we train two models, $A$ and $B$, using the $h$ most recent observations and evaluate them on the same target, $\mathbf{y}$. 
For each model $k \in \{A, B\}$, we can break down its average prediction score, $\overline{R^2_k}(h)$, 
into three intuitive components defined in~\eqref{eq:main-aqu}: alignment $a_k(h)$, aggressiveness $g_k(h)$, and instability $u_k(h)$.

Using the identity from~\eqref{eq:main-identity}, we can calculate the exact performance gap, $\Delta_{AB}(h)$, between the two models:
\begin{align}
\Delta_{AB}(h)&=\overline{R^2_A}(h)-\overline{R^2_B}(h)\notag\\
&=2\{a_Ag_A-a_Bg_B\}-\{g_A^2-g_B^2\}\notag\\
&\quad-\{u_A-u_B\}.
\label{eq:main-gap}
\end{align}
Model $A$ outperforms model $B$ at memory $h$ whenever this gap is positive ($\Delta_{AB}(h) > 0$). 
A rank reversal occurs exactly when the performance gap changes sign, meaning $\Delta_{AB}(h_1)\Delta_{AB}(h_2) < 0$.

\paragraph{Complexity phase boundary.}
Let $\mathcal M_0$ be a simple model class and $\mathcal M_1$ a more complex class.
Let $\mathcal L_k(h)$ denote the excess loss of class $k\in\{0,1\}$ at memory $h$, that is, the expected squared error against the current conditional mean.
We assume this loss expands as
\begin{equation}
\mathcal L_k(h)=\mathcal A_k+B_k+\frac{V_k}{h}+K_kh+W_kh^2+e_k(h),
\label{eq:main-class-risk}
\end{equation}
with a known remainder bound $|e_k(h)|\le r_k(h)$. 
Here, $\mathcal A_k$ is the optimal approximation error, and $B_k$ is a static, memory-independent bias. 
The terms $V_k/h$, $K_kh$, and $W_kh^2$ capture the core memory tradeoff: 
estimation variance (which decays with $h$) versus linear and quadratic drift penalties (which grow as older data is included).

Setting $\Delta_0=(\mathcal A_0+B_0)-(\mathcal A_1+B_1)$ isolates the history-independent advantage of the complex class. 
Defining the coefficient differences $\Delta V=V_1-V_0$, $\Delta K=K_1-K_0$, and $\Delta W=W_1-W_0$ yields its history-dependent cost, $\phi(h)=\Delta V/h+\Delta Kh+\Delta Wh^2$ (Fig.~\ref{fig:theory}A).

\paragraph{Proposition 2 (phase boundary).}
The complex class has strictly lower loss at memory $h$ 
if its history-independent advantage overcomes the complexity cost and remainder bounds, meaning $\Delta_0>\phi(h)+r_0(h)+r_1(h)$. 
Conversely, the simple class performs better if $\Delta_0<\phi(h)-r_0(h)-r_1(h)$. 

Setting aside the remainders, suppose the complex class incurs higher variance ($\Delta V>0$) and quadratic drift ($\Delta W>0$) penalties. 
The cost function $\phi(h)$ is then strictly convex (U-shaped), reaching a unique minimum $\phi_{\min}$ 
at a memory $h_{\rm cost}$ that solves $2\Delta W h_{\rm cost}^3+\Delta Kh_{\rm cost}^2=\Delta V$. 
If $\Delta_0<\phi_{\min}$, the complex class never performs better. 
If $\Delta_0>\phi_{\min}$, it outperforms the simple class 
within an optimal memory window bounded by the two positive roots of $\Delta Wh^3+\Delta Kh^2-\Delta_0h+\Delta V=0$. 

When linear drift differences vanish ($\Delta K=0$), 
this memory scales as $h_{\rm cost}=(\Delta V/2\Delta W)^{1/3}$ with minimum cost $\phi_{\min}=3\cdot2^{-2/3}(\Delta V)^{2/3}(\Delta W)^{1/3}$. 
Because varying the training memory dynamically shifts this variance-drift tradeoff, parameter count alone does not determine the prevailing model. 
\textbf{In summary, the complex class only performs better within a specific window of training memory where its baseline advantage outweighs the escalating penalties of variance and data drift.}
(Detailed proofs and degenerate cases are in Propositions~\ref{res:phase} and~\ref{res:phase-leading} of Appendix~\ref{sec:theory-complexity}).

\subsection{Exact prediction risk and optimal memory for ridge regression}
\label{sec:theory-main-ridge}

We first bound the prediction risk of ridge regression under coefficient drift. 
Consider a forecast date $t$ and a training history of $h$ observations $(x_{t-j},y_{t-j})$ for $j=1,\dots,h$, with predictors $x_{t-j}\in\mathbb{R}^p$. 
We model nonstationarity as a local linear drift: $y_{t-j}=x_{t-j}^\top\beta_{t-j}+\varepsilon_{t-j}$, 
where the historical coefficients evolve as $\beta_{t-j}=\beta_t-j\dot\beta_t+r_{t,j}$. 
Here, $\beta_t$ represents the current target coefficient, $\dot\beta_t$ the constant per-period drift, $r_{t,j}$ the nonlinear drift remainder, and $\varepsilon_{t-j}$ the observation noise.

Define the sample second-moment matrix $S_h=h^{-1}\sum_jx_{t-j}x_{t-j}^\top$, the ridge inverse $A_h=(S_h+\lambda I)^{-1}$ for penalty $\lambda>0$, the ridge filter $G_h=A_hS_h$, and the mean temporal lag $\mu_h=(h+1)/2$. 
We measure prediction differences using the local design norm $\|v\|_{S_h}^2=v^\top S_hv$. 
Conditioning on the realized training design $X_h$, we assume the score vector $z=h^{-1}\sum_jx_{t-j}\varepsilon_{t-j}$ satisfies $\operatorname{Var}(z\mid X_h)\preceq\kappa_h\sigma_t^2S_h/h$. 
Here $\sigma_t^2$ is the error variance and $\kappa_h\ge1$ is a dependence factor, 
the assumption does not require $z$ to have a strict conditional mean of zero.

The conditional risk $\mathcal E_t^X(h)=\operatorname{E}[\|\widehat\beta-\beta_t\|_{S_h}^2\mid X_h]$ of the ridge estimator $\widehat\beta$ partitions into four interpretable components:
\begin{align*}
B_{\rm reg}&=\|(I-G_h)\beta_t\|_{S_h}^2,\quad
D^2=\|G_h\dot\beta_t\|_{S_h}^2,\\
d_{2,h}&=\operatorname{tr}[S_h^2(S_h+\lambda I)^{-2}],
\end{align*}
and a remainder $\rho^2$ that captures coefficient curvature, time-imbalance in the design, and the conditional mean of $z$. 

Applying an exact bias-variance identity, the risk satisfies the following bound for any $\eta,\zeta>0$ (Proposition~\ref{res:ridge} in Appendix~\ref{sec:theory-ridge}):
\begin{tcolorbox}[
    ams align, 
    colback=gray!5,       
    colframe=blue!50!black,    
    sharp corners,%
    boxrule=0.8pt,%
    boxsep=1pt,%
    left=4pt, right=4pt
]
\mathcal E_t^X(h)&\le(1+\eta)(1+\zeta)B_{\rm reg}\notag\\
&\quad+(1+\eta)(1+\zeta^{-1})\mu_h^2D^2\notag\\
&\quad+(1+\eta^{-1})\rho^2+\kappa_h\sigma_t^2d_{2,h}/h.
\label{eq:main-ridge-bound}
\end{tcolorbox}
Crucially, this bound is \emph{direction-aware}. 
The variance decays as $1/h$ via the spectral dimension $d_{2,h}$, while the drift penalty grows as $h^2$ via $\mu_h^2D^2$. 
Because the temporal drift $\dot\beta_t$ enters exclusively through the filter $G_h$, 
ridge regression naturally attenuates drift in low-variance predictor directions.

To isolate the exact memory dependence, 
assume the local second-moment matrix remains constant across candidate histories ($S_h=S$), 
the remainder term vanishes ($\rho=0$), 
and the conditional score variance scales inversely with memory as $\operatorname{Var}(z\mid X_h)=\kappa\sigma_t^2S/h$ for a constant $\kappa>0$. 
Define the stable ridge filter $G=(S+\lambda I)^{-1}S$, the effective spectral dimension $d_2=\operatorname{tr}[S^2(S+\lambda I)^{-2}]$, and the inner product $\langle u,v\rangle_S=u^\top Sv$. 
The conditional risk then reduces to four fundamental components: the regularization bias $B_t=\|(I-G)\beta_t\|_S^2$, 
the filtered drift energy $Q_t=\|G\dot\beta_t\|_S^2$, 
the bias-drift cross-term $J_t=\langle(I-G)\beta_t,G\dot\beta_t\rangle_S$, 
and the variance scale $V_t=\kappa\sigma_t^2d_2$:
\begin{equation}
\mathcal E_t(h)=B_t+2\mu_hJ_t+\mu_h^2Q_t+V_t/h.
\label{eq:main-risk}
\end{equation}
Equation~\eqref{eq:main-risk} matches the form of Equation~\eqref{eq:main-class-risk} 
with $V_k=V_t$, $K_k=J_t+Q_t/2$, and $W_k=Q_t/4$ (Fig.~\ref{fig:theory}B).

\paragraph{Proposition 3 (optimal memory for ridge regression).}
Assume strictly positive variance and drift energy ($V_t>0$ and $Q_t>0$), and extend Equation~\eqref{eq:main-risk} to real $h>0$. 
The continuous risk function possesses a unique positive minimizer $h_t^*$, which solves the optimality condition:
\begin{tcolorbox}[
    ams equation,
    colback=gray!5,       
    colframe=blue!50!black,  
    sharp corners,               
    boxrule=0.8pt,          
    boxsep=1pt,             
    left=4pt, right=4pt%
]
V_t/(h_t^*)^2=J_t+\tfrac{h_t^*+1}{2}Q_t.
\label{eq:main-memory}
\end{tcolorbox}
When the bias-drift cross-term vanishes ($J_t=0$), the solution satisfies $(h_t^*)^2(h_t^*+1)=2V_t/Q_t$, converging asymptotically to the cube-root rule $h_t^*\approx(2V_t/Q_t)^{1/3}$ for large training windows. 
Nonzero cross-terms structurally shift the optimal window: a negative $J_t$ lengthens the ideal memory, whereas a positive $J_t$ shortens the ideal memory. 
(Detailed proofs appear in Appendices~\ref{sec:theory-ridge} and~\ref{sec:theory-memory}).

\subsection{Surrogate approximations and transfer bounds for trees and neural networks}
\label{sec:theory-main-transfer}

Because architectures like boosted trees, RNNs, LSTMs, and Transformers are highly nonlinear in the underlying parameters, 
analyzing the exact memory dynamics requires an intermediate linear surrogate. 
We construct the theoretical surrogate by freezing the feature map prior to fitting the final readout layer (Propositions~\ref{res:frozen} and~\ref{res:boost} in Appendix~\ref{sec:theory-fixed-representations}). 
For tree-based models, the frozen map corresponds to a fixed tree topology; for neural networks, the frozen map represents the encoder producing the final hidden state or the pooled token state. 
Applying a ridge regression readout to the fixed features ensures the surrogate risk follows the exact complexity expansion in Equation~\eqref{eq:main-class-risk}, governed by the variance component $V_k/h$ and the quadratic drift component $W_kh^2$.

To quantify the performance gap between the theoretical surrogate and the actual model $k$ at memory $h$, let $g_{k,h}$ denote the surrogate forecast and $f_{k,h}$ denote the fully trained forecast. 
Define $\mathcal L_k^0(h)$ and $\mathcal L_k(h)$ as the respective prediction risks against the current conditional mean. 
Finally, let $\delta_{k,h}^2$ represent the mean squared distance between the trained forecast $f_{k,h}$ and the surrogate forecast $g_{k,h}$, measured under the evaluation norm.

\paragraph{Proposition 4 (transfer bounds).}
The prediction risks of the fully trained model and the linear surrogate differ by at most $\epsilon_{k,h}=2\delta_{k,h}\sqrt{\mathcal L_k^0(h)}+\delta_{k,h}^2$.
Consequently, the trained complex class achieves strictly lower risk than the trained simple class if the history-independent advantage $\Delta_0$ exceeds the combined complexity cost and remainder bounds: $\Delta_0>\phi(h)+r_0(h)+r_1(h)+\epsilon_{1,h}+\epsilon_{0,h}$.
Conversely, the simple class achieves lower risk if $\Delta_0<\phi(h)-r_0(h)-r_1(h)-\epsilon_{1,h}-\epsilon_{0,h}$.

Furthermore, assuming a uniform transfer error bound $\epsilon_{k,h}\le\epsilon$ over an interval of candidate histories, and strictly positive surrogate coefficients $V_k,W_k>0$, the optimal memory for the trained risk lies within a distance of $\sqrt{2\epsilon/W_k}$ from the surrogate optimum.
The surrogate optimum solves $2W_kh^3+K_kh^2=V_k$, where $K_k$ represents the linear drift coefficient from Equation~\eqref{eq:main-class-risk} (Proposition~\ref{res:transfer} and Corollary~\ref{res:architecture} in Appendix~\ref{sec:theory-architecture}).

Because the transfer error $\epsilon$ enters additively and scales linearly with the perturbation $\delta$ (for small $\delta$), the phase boundary and optimal memory expressions preserve the fundamental scaling dynamics of ridge regression: 
a $1/h$ decay for estimation variance and an $h^2$ growth for quadratic drift (Fig.~\ref{fig:theory}).
The various nonlinear architectures differ only in how the perturbation bound $\delta$ scales:
\begin{list}{$\bullet$}{
\setlength{\itemindent}{0pt}
\setlength{\leftmargin}{1.0em} 
\setlength{\itemsep}{5pt}
\setlength{\parsep}{0pt}
\setlength{\parskip}{0pt}
}
\item \textbf{Boosted trees.} For greedy trees built on a finite threshold grid with a strictly positive split margin $\Delta_{\rm split}>0$, the \textbf{transfer error scales as $\epsilon\propto e^{-\alpha h}$}, where $\alpha\propto\Delta_{\rm split}^2$ (Proposition~\ref{res:xgb}). The selection penalty decays exponentially as history length $h$ increases, shifting the optimal memory toward longer windows.
\item \textbf{RNN.} Provided the recurrent weights maintain an operator norm $\chi<1$, the perturbation $\delta$ scales linearly with the movement of the network weights and readout. The sequence scaling factor $(1-\chi^L)/(1-\chi)\le1/(1-\chi)$ remains bounded, \textbf{independent of the context length $L$} (Proposition~\ref{res:rnn}).
\item \textbf{LSTM.} For Long Short-Term Memory networks, the perturbation $\delta$ scales linearly with the movement of the gating mechanisms and the readout. The sequence scaling factor expands to $\sum_{k<L}\varrho^k$, governed by the Lipschitz constant $\varrho$ of the cell state update. The scaling factor remains bounded for $\varrho<1$, but \textbf{grows exponentially with context length $L$ if $\varrho\ge1$} (Proposition~\ref{res:lstm}).
\item \textbf{Transformer.} For a single attention block, the perturbation $\delta$ scales linearly with the movement of the attention logits, value matrices, output map, and readout. The scaling factor relies on $\sqrt{N_{\rm head}}$ and exhibits \textbf{no explicit dependence on the context length $L$} (Proposition~\ref{res:transformer}).
\end{list}
For exact-greedy trees, the transfer bound does not decay with history length $h$ (Complete conditions and proofs appear in Appendix~\ref{sec:theory-architecture}).

%% file: experiments.tex
\section{Experiments}
\label{sec:experiments}

\subsection{Controlled simulation}
\label{sec:experiments-simulation}

\paragraph{Design.}
We generate eight Gaussian covariates, $x_1,\dots,x_8$, with variances $q^{k-1}$ for spectral decay $q\in\{1,0.45\}$ (Appendix~\ref{app:sim-method}, covariates). 
The response combines a linear signal, a nonlinear interaction ($\gamma$ times the standardized product $x_1x_2$), and Gaussian noise, with interaction strength $\gamma\in\{0,0.4,0.8\}$ and SNR $\in\{0.25,2\}$ (Appendix~\ref{app:sim-method}, signal). 
To simulate nonstationarity, the first coefficient of an observation aged $a$ periods shifts by $-0.008a$ (gradual drift), $-0.65$ for $a>48$ (abrupt break), or $0$ (zero drift); test targets use unshifted current coefficients (Appendix~\ref{app:sim-method}, drift). 
We compare linear ridge on the 8 base covariates against quadratic ridge on 44 expanded features (including $x_1x_2$), testing penalties $\lambda\in\{0.03,0.5\}$ and memories $h\in\{16,32,64,128,256\}$ (Appendix~\ref{app:sim-method}, models). 
Each factorial condition uses 24 independent worlds, each containing 16 test targets and three training histories (Appendix~\ref{app:sim-method}, replicates). 
We score out-of-sample $R^2$ against the zero forecast, analyzing the performance gap $\Delta R^2=R^2_{\rm quad}-R^2_{\rm lin}$ with paired 95\% intervals across worlds (Appendix~\ref{app:sim-method}, score and contrasts). 

\begin{figure}[htb]
\centering
\includegraphics[width=0.9\columnwidth]{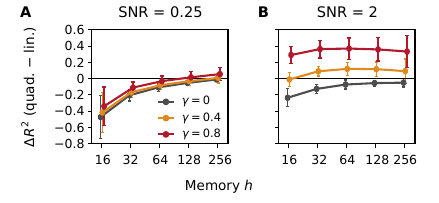}
\vspace{-1em}
\caption{Out-of-sample performance gap $\Delta R^2$ (quadratic minus linear) across training memories under gradual drift ($q=0.45$, $\lambda=0.03$). Error bars denote paired 95\% intervals.} 
\label{fig:sim-main}
\end{figure}

\paragraph{Results.}
Across 360 comparisons, the quadratic model outperforms the linear baseline in 0 of 120 cases at $\gamma=0$, 46 at $\gamma=0.4$, and 78 at $\gamma=0.8$ (Appendix~\ref{app:sim-results}, Figs.~\ref{fig:sim-phase-zero}, \ref{fig:sim-phase-gradual}, and~\ref{fig:sim-phase-abrupt}). 
Under gradual drift ($\gamma=0.8$), $\Delta R^2$ is positive for all memories at SNR 2; at SNR 0.25, it remains negative for $h\le64$ and turns positive only at $h\ge128$, corroborating the phase boundary of Section~\ref{sec:theory-main-memory} (Fig.~\ref{fig:sim-main}; Appendix~\ref{app:sim-results}, Fig.~\ref{fig:sim-phase-gradual}). 
In a balanced linear design, optimal memory falls from 128 to 32 as drift rises from 0.002 to 0.016; the $-0.67$ log-log slope tightly matches the predicted $-2/3$ from Section~\ref{sec:theory-main-ridge} (Appendix~\ref{app:sim-results}, Fig.~\ref{fig:sim-scaling}). 
At equal raw drift, linear ridge ($\lambda=0.03$) $R^2$ drops from $0.57$ ($h=16$) to $-2.37$ ($h=256$) if the drifting coordinate variance is 1, but rises from $0.60$ to $0.67$ if the variance is $0.004$, confirming ridge attenuates drift in low-variance directions (Appendix~\ref{app:sim-results}, Fig.~\ref{fig:sim-directional}). 

\subsection{Empirical dynamics of forecast components}
\label{sec:experiments-main}

\paragraph{Design.}
Following established protocols \citep{capponi2025nonstationarity}, we forecast daily excess returns for 17 U.S. industry portfolios \citep{fama1997industry}. 
The dataset spans September 1987 to November 2016, dedicating January 1990 onwards strictly to out-of-sample evaluation. 
The daily information set comprises 129 predictors partitioned into four structural categories: 
(1) 94 feature-based portfolios capturing firm-level attributes such as price trends, liquidity, and volatility \citep{gu2020empirical}; 
(2) three standard market factors \citep{fama1993common}; 
(3) 15 monthly structural series, including macroeconomic hidden states extracted from 178 underlying economic indicators \citep{chen2024deep}; and 
(4) the previous day's 17 target returns to capture autoregressive dynamics.

Six tabular models, fixed- and tuned-penalty ridge, random forest, XGBoost, small MLP, and medium MLP, process single-day predictors. 
Seven temporal models ingest the trailing $L\in\{20,60\}$ days: fixed- and tuned-penalty ridge, random forest, XGBoost, MLPs, an RNN, an LSTM, and a Transformer (Appendix~\ref{app:stage5-method}, Model specifications).
Each model trains on a rolling 16-month, rolling 64-month, or expanding historical window, with hyperparameters tuned on immediately preceding data (Appendix~\ref{app:stage5-method}, Hyperparameter selection).
We evaluate six random-seed replicates per configuration, scoring out-of-sample $R^2$ against the zero forecast and decomposing performance gaps via identity~\eqref{eq:main-identity} (Appendix~\ref{app:stage5-method}, Replicate statistics).

\begin{figure}[htb]
\centering
\includegraphics[width=0.9\columnwidth]{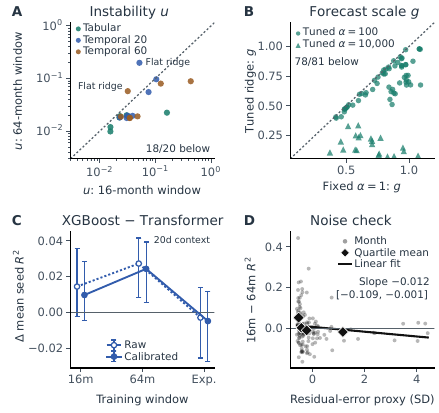}
\caption{Empirical dynamics of forecast components. 
Model classes are color-coded: tabular (teal), 20-day temporal (blue), and 60-day temporal (brown). 
(A)~Restricted training memory systematically exacerbates predictive instability $u$. 
(B)~Stronger ridge regularization effectively attenuates the aggressiveness penalty $g$. 
(C)~Ex-post scalar calibration preserves target alignment, failing to resolve fundamental performance gaps between models. 
(D)~Elevated ambient noise systematically diminishes the out-of-sample advantage of short training memory.}
\label{fig:main-forecast}
\end{figure}

\begin{table*}[htb]
\caption{Memory-induced rank reversals and the empirical recovery of predictive instability. Expanding the training horizon (16m to Exp.) systematically inverts model rankings, favoring heavily regularized or temporally structured architectures over static, complex baselines at longer histories. Additionally, replicate ensembling strictly improves upon mean predictive performance, quantifying the exact out-of-sample cost of finite-sample instability. Shading identifies the optimal architecture within the tabular and 60-day temporal branches; RMSE is reported in percentage points of daily excess returns.}
\label{tab:main-results}
\centering
\begingroup\scriptsize\setlength{\tabcolsep}{1.5pt}
\begin{tabular}{@{}l*{13}{c}@{}}
\toprule
 & \multicolumn{6}{c}{Tabular} & \multicolumn{7}{c}{Temporal, 60-day context} \\
\cmidrule(lr){2-7}\cmidrule(lr){8-14}
 & \begin{tabular}[t]{@{}c@{}}Ridge,\\$\alpha{=}1$\end{tabular} & \begin{tabular}[t]{@{}c@{}}Tuned\\ridge\end{tabular} & \begin{tabular}[t]{@{}c@{}}Random\\forest\end{tabular} & \begin{tabular}[t]{@{}c@{}}XGBoost\end{tabular} & \begin{tabular}[t]{@{}c@{}}Small\\MLP\end{tabular} & \begin{tabular}[t]{@{}c@{}}Medium\\MLP\end{tabular} & \begin{tabular}[t]{@{}c@{}}Flattened\\ridge\end{tabular} & \begin{tabular}[t]{@{}c@{}}Flattened\\XGBoost\end{tabular} & \begin{tabular}[t]{@{}c@{}}Fixed\\flattened\\MLP\end{tabular} & \begin{tabular}[t]{@{}c@{}}Tuned\\flattened\\MLP\end{tabular} & \begin{tabular}[t]{@{}c@{}}Vanilla\\RNN\end{tabular} & \begin{tabular}[t]{@{}c@{}}LSTM\end{tabular} & \begin{tabular}[t]{@{}c@{}}Small\\Transformer\end{tabular} \\
\midrule
\multicolumn{14}{@{}l}{\emph{Mean $R^2\uparrow$}} \\
16m & $-$0.0909 & 0.3515 & \cellcolor[HTML]{C8E6C9}0.3549 & 0.3316 & 0.2869 & 0.2786 & \phantom{$-$}0.0772 & \cellcolor[HTML]{C8E6C9}0.2812 & $-$0.8216 & $-$0.2223 & 0.2776 & 0.1913 & 0.2712 \\
64m & \phantom{$-$}0.3256 & \cellcolor[HTML]{C8E6C9}0.3668 & 0.3374 & 0.3419 & 0.3231 & 0.3189 & \phantom{$-$}0.1078 & \cellcolor[HTML]{C8E6C9}0.3223 & $-$0.0124 & \phantom{$-$}0.0002 & 0.2989 & 0.2472 & 0.2969 \\
Exp. & \phantom{$-$}0.3386 & \cellcolor[HTML]{C8E6C9}0.3602 & 0.2953 & 0.3232 & 0.3301 & 0.3270 & $-$0.2835 & 0.3148 & \phantom{$-$}0.0881 & \phantom{$-$}0.1158 & 0.3169 & 0.2885 & \cellcolor[HTML]{C8E6C9}0.3195 \\
\midrule
\multicolumn{14}{@{}l}{\emph{Ensemble $R^2\uparrow$}} \\
16m & \phantom{$-$}0.0706 & 0.3671 & \cellcolor[HTML]{C8E6C9}0.3705 & 0.3621 & 0.3096 & 0.3018 & \phantom{$-$}0.1092 & 0.3161 & $-$0.3891 & $-$0.0970 & 0.3123 & 0.2153 & \cellcolor[HTML]{C8E6C9}0.3191 \\
64m & \phantom{$-$}0.3481 & \cellcolor[HTML]{C8E6C9}0.3767 & 0.3494 & 0.3619 & 0.3414 & 0.3383 & \phantom{$-$}0.1654 & \cellcolor[HTML]{C8E6C9}0.3402 & \phantom{$-$}0.0765 & \phantom{$-$}0.0804 & 0.3167 & 0.2661 & 0.3160 \\
Exp. & \phantom{$-$}0.3477 & \cellcolor[HTML]{C8E6C9}0.3644 & 0.3079 & 0.3392 & 0.3423 & 0.3425 & $-$0.0295 & 0.3253 & \phantom{$-$}0.1690 & \phantom{$-$}0.1822 & 0.3276 & 0.3042 & \cellcolor[HTML]{C8E6C9}0.3342 \\
\midrule
\multicolumn{14}{@{}l}{\emph{Mean RMSE (pp) $\downarrow$}} \\
16m & 1.4675 & 1.1315 & \cellcolor[HTML]{C8E6C9}1.1286 & 1.1487 & 1.1865 & 1.1934 & 1.3498 & \cellcolor[HTML]{C8E6C9}1.1913 & 1.8962 & 1.5534 & 1.1942 & 1.2636 & 1.1995 \\
64m & 1.1539 & \cellcolor[HTML]{C8E6C9}1.1181 & 1.1438 & 1.1399 & 1.1560 & 1.1596 & 1.3272 & \cellcolor[HTML]{C8E6C9}1.1567 & 1.4137 & 1.4049 & 1.1765 & 1.2191 & 1.1782 \\
Exp. & 1.1427 & \cellcolor[HTML]{C8E6C9}1.1239 & 1.1795 & 1.1559 & 1.1500 & 1.1527 & 1.5918 & 1.1631 & 1.3418 & 1.3212 & 1.1613 & 1.1851 & \cellcolor[HTML]{C8E6C9}1.1591 \\
\midrule
\multicolumn{14}{@{}l}{\emph{Ensemble RMSE (pp) $\downarrow$}} \\
16m & 1.3546 & 1.1178 & \cellcolor[HTML]{C8E6C9}1.1148 & 1.1222 & 1.1675 & 1.1741 & 1.3262 & 1.1620 & 1.6560 & 1.4716 & 1.1652 & 1.2446 & \cellcolor[HTML]{C8E6C9}1.1594 \\
64m & 1.1345 & \cellcolor[HTML]{C8E6C9}1.1093 & 1.1334 & 1.1224 & 1.1403 & 1.1430 & 1.2836 & \cellcolor[HTML]{C8E6C9}1.1413 & 1.3503 & 1.3474 & 1.1615 & 1.2037 & 1.1621 \\
Exp. & 1.1348 & \cellcolor[HTML]{C8E6C9}1.1202 & 1.1690 & 1.1422 & 1.1395 & 1.1394 & 1.4256 & 1.1541 & 1.2808 & 1.2707 & 1.1522 & 1.1720 & \cellcolor[HTML]{C8E6C9}1.1465 \\
\bottomrule
\end{tabular}
\endgroup
\end{table*}

\paragraph{Results.}
Evaluated by mean out-of-sample $R^2$, random forest performance degrades from $0.3549$ under a 16-month memory to $0.2953$ under an expanding history, whereas tuned ridge regression improves from $0.3515$ to $0.3602$ (Table~\ref{tab:main-results}).
The complex tree-based model performs better at short memories where its static approximation advantage dominates, but the simpler ridge model performs better as history expands and variance-drift penalties shift, consistent with results in Section~\ref{sec:theory-main-memory}.
A parallel reversal emerges among the 60-day temporal architectures: flattened XGBoost achieves the highest mean $R^2$ at 16 and 64 months ($0.2812$ and $0.3223$), but the small Transformer strictly overtakes it under expanding history ($0.3195$).
Finally, quantifying the empirical magnitude of predictive instability, replicate ensembling recovers substantial performance costs, boosting the 16-month random forest $R^2$ to $0.3705$ and the expanding-history Transformer to $0.3342$ (Table~\ref{tab:main-results}; full results in Appendix Table~\ref{tab:stage5-all}).

As predicted by the variance term $V/h$ (Eq.~\eqref{eq:main-risk}), restricted memory exacerbates instability: $u$ is strictly larger at 16 than 64 months in 18 of 20 architecture--context pairs (Fig.~\ref{fig:main-forecast}A; Appendix Table~\ref{tab:stage5-all}). 
Ridge systematically attenuates forecast magnitude and variance (Section~\ref{sec:theory-main-ridge}). Compared to a penalty of 1, tuned ridge (penalty 100 or 10,000) lowers aggressiveness $g$ in 78 of 81 year-window pairs and instability $u$ in all 81 (Fig.~\ref{fig:main-forecast}B). 
Because scalar transformations preserve target alignment (Section~\ref{sec:theory-main-decomposition}), calibration cannot resolve fundamental disparities. A historically fitted $[0,1]$ scalar shifts the 20-day XGBoost--Transformer gap by at most $0.005$ (Fig.~\ref{fig:main-forecast}C). 

Optimal memory expands with noise and contracts with filtered drift (Eq.~\eqref{eq:main-memory}). 
Empirically, elevated noise (recent ridge-fit squared error) lowers the 16-month advantage by $0.012$ $R^2$ per standard deviation (95\% interval $[-0.109,-0.001]$; Fig.~\ref{fig:main-forecast}D). 

\subsection{Generalization to independent data}%
\label{sec:experiments-independent}

\paragraph{Design.}
To evaluate firm-level prediction, we extend our framework to daily CRSP equity data spanning 2018--2024, forecasting next-day total returns using the stock universe and 24 feature representations established by \citep{zhang2026benchmarking}. 
We refit all models monthly from January 2021 through December 2024 across three training memories: a rolling 252-day window, a rolling 756-day window, and an expanding historical window. 
Three tabular models (ridge, XGBoost, and an MLP) process single-day predictors, while six temporal architectures ingest historical contexts of the trailing 20 or 60 days. 
Hyperparameters remain strictly fixed across evaluations, yielding two paired-seed replicates per architecture--context pair.

\paragraph{Results.}
On the independent CRSP stock-return panel, XGBoost strictly dominates the fixed architectures across all nine context--memory combinations (Table~\ref{tab:stage6-best}). 
In a separate 2022--2024 evaluation, applying an ex-post scalar calibration (fitted over the preceding 12 months) collapses nine XGBoost--neural performance gaps from the \(0.20\)--\(1.03\) range to under \(0.00015\). 
Because this convergence is achieved almost entirely by attenuating forecast magnitudes toward zero, it confirms that the initial performance disparities stemmed from miscalibrated aggressiveness rather than fundamental differences in target alignment.

\begin{table}[t]
\centering
\footnotesize
\setlength{\tabcolsep}{3pt}
\begin{tabular}{@{}lccc@{}}
\toprule
Context & 252d & 756d & Exp. \\
\midrule
Tabular      & $-0.00189$ & $-0.00145$ & $-0.00129$ \\
Temporal 20d & $-0.00198$ & $-0.00132$ & $-0.00076$ \\
Temporal 60d & $-0.00199$ & $-0.00135$ & $-0.00096$ \\
\bottomrule
\end{tabular}
\caption{Mean pooled out-of-sample $R^2$ against a zero-return baseline (January 2021--December 2024).}
\label{tab:stage6-best}
\end{table}

%% file: conclusion.tex
\section{Conclusion}
\label{sec:conclusion}

This paper investigates the fundamental mechanisms governing the heterogeneous sensitivity of predictive architectures to training memory under distribution shift. 
To this end, we derive a model-agnostic decomposition of out-of-sample $R^2$ that partitions predictive performance into target alignment, an aggressiveness penalty, and finite-sample predictive instability. 
Building upon this identity, we formally characterize the complexity phase boundary---delineating the memory regimes where high-capacity models overcome simpler baselines---and establish a closed-form optimal memory equation for ridge regression under local drift. 
Controlled simulations rigorously validate both the predicted phase boundaries and the derived $-2/3$ asymptotic scaling of optimal memory with respect to drift velocity. 
Crucially, in empirical financial forecasting, our framework mechanically explains rank reversals across lookback windows.
The identity isolates these inversions to shifting tensions between alignment and aggressiveness, while confirming that replicate ensembling cleanly recovers the precise theoretical penalty of predictive instability. 
While prior literature has traditionally treated the estimation--drift tradeoff, model benchmarking, and adaptive memory selection as disparate domains, 
our primary contribution is structural: by unifying these fields through a single deterministic identity, 
we trace any macroscopic shift in model ranking directly to a quantified, mechanistic forecast property.

\textbf{Computational and economic implications}. 
Aggressively restricting the training window is an inadequate heuristic for nonstationarity; when ambient noise dominates filtered drift, heavily regularized linear models with expanding histories can match high-capacity learners at a fraction of the cost. 
Our decomposition therefore serves as a practical diagnostic, allowing practitioners to evaluate the three forecast components before funding expensive architecture searches. 
Conversely, while replicate ensembling eliminates the instability penalty, it scales training costs linearly. 
Ultimately, the framework quantifies the exact marginal predictive return of each additional unit of compute.

\textbf{Limitations and future work}.
While our closed-form derivations apply exactly to ridge regression under local drift, theoretical guarantees for high-capacity architectures currently rely on surrogate transfer bounds. 
Additionally, because memory is parameterized strictly via uniform truncation windows, the framework does not address exponentially weighted or adaptive histories. 
Future research could generalize the theory to accommodate dynamic forgetting factors, leveraging the proposed three-component decomposition for online, adaptive memory selection. 
Further natural extensions include tightening the nonlinear transfer bounds, integrating compute-aware evaluation metrics, and validating these complexity phase boundaries across diverse forecasting horizons and alternative asset classes.

%% file: appendix_theory-1.tex
\clearpage
\appendix
\thispagestyle{plain}

\onecolumn
\aistatstitle{Decomposing Rank Reversals under Temporal Nonstationarity: Alignment, Aggressiveness, and Instability}

\newcounter{appres}
\renewcommand{\theappres}{A.\arabic{appres}}
\newcommand{\appresult}[2]{\refstepcounter{appres}\paragraph{#1~\theappres\ (#2).}}

\section{Theoretical Results and Proofs}
\label{sec:theory}

This appendix gives the derivations behind Section~\ref{sec:theory-main}.
It has four subsections.
Each subsection expands the subsection of Section~\ref{sec:theory-main} with the same number, in the same order and with the same logic.
We restate the claims of Section~\ref{sec:theory-main}, add the definitions and assumptions that it omits, and prove every claim.
We also include further results that Section~\ref{sec:theory-main} does not state.

In Section~\ref{sec:theory-app-decomposition}, we expand Section~\ref{sec:theory-main-decomposition}.
We prove Proposition~1 of Section~\ref{sec:theory-main-decomposition} in Proposition~\ref{res:scalar}, together with the constrained scaling rules that Section~\ref{sec:theory-main-decomposition} omits.
These results hold in every finite sample.
They need no stationarity assumption and no linear-model assumption.

In Section~\ref{sec:theory-app-memory}, we expand Section~\ref{sec:theory-main-memory}.
We prove Proposition~2 of Section~\ref{sec:theory-main-memory} in Propositions~\ref{res:phase} and~\ref{res:phase-leading}.
The phase boundary needs the risk expansion~\eqref{eq:main-class-risk}.

In Section~\ref{sec:theory-app-ridge}, we expand Section~\ref{sec:theory-main-ridge}.
We bound the risk of ridge regression under coefficient drift and serial dependence, which proves~\eqref{eq:main-ridge-bound} (Proposition~\ref{res:ridge}).
We then solve for the memory that minimizes an exactly solvable version of this risk, which proves Proposition~3 of Section~\ref{sec:theory-main-ridge} (Corollary~\ref{res:interior-memory}).
Each result states the conditions that it needs.

In Section~\ref{sec:theory-app-transfer}, we expand Section~\ref{sec:theory-main-transfer}.
We extend the ridge bound to the surrogate, that is, to frozen feature maps (Proposition~\ref{res:frozen}) and to boosted trees with fixed splits (Proposition~\ref{res:boost}).
We derive an exact coordinate-wise phase boundary for the surrogate (Proposition~\ref{res:spectral}).
We then bound the gap between the surrogate and the trained model.
This proves Proposition~4 of Section~\ref{sec:theory-main-transfer} (Proposition~\ref{res:transfer} and Corollary~\ref{res:architecture}), and it gives the conditions for boosted trees, RNNs, LSTMs, and Transformers (Propositions~\ref{res:xgb}--\ref{res:transformer}).

Throughout, $\|\cdot\|$ and $\langle\cdot,\cdot\rangle$ are the Euclidean norm and inner product.
For symmetric matrices, $A\preceq B$ means that $B-A$ is positive semidefinite.
We write $\operatorname{E}$ and $\operatorname{Var}$ for expectation and variance, $\operatorname{tr}$ for the trace, and $\mathrm R$ for the real line.
The index $t$ is the forecast date, and the memory $h$ is the number of most recent observations that train a model.
We number the results of this appendix A.1, A.2, and so on.
We cite the results of Section~\ref{sec:theory-main} by their number in that section.

\subsection{Forecast-score decomposition, ensembling, and scalar calibration}
\label{sec:theory-app-decomposition}

\appresult{Proposition}{optimal common scalar}\label{res:scalar}
Assume $q>0$.
Then $\mathcal S(\gamma)=2\gamma c-\gamma^2q$.
Let $\mathcal I\subseteq\mathrm R$ be any nonempty closed interval, bounded or unbounded.
On $\mathcal I$, the score has the unique maximizer
\begin{equation}
\boxed{\gamma^*_{\mathcal I}
  =\Pi_{\mathcal I}\!
   \left(\frac{c}{q}\right),}
\label{eq:scalar-optimum}
\end{equation}
where $\Pi_{\mathcal I}$ is the projection onto the interval.
The interval sets the calibration rule.
Signed scaling uses $\mathcal I=\mathrm R$ and gives $c/q$.
Sign-preserving scaling uses $\mathcal I=[0,\infty)$ and gives $\max\{0,c/q\}$.
Shrinkage-only scaling uses $\mathcal I=[0,1]$ and gives $\min\{1,\max\{0,c/q\}\}$.
For unconstrained scaling, the maximum score and the gain over $\gamma=1$ are
\begin{equation}
\max_{\gamma}\mathcal S(\gamma)=\frac{c^2}{q},
\qquad
\mathcal S(\gamma^*)-\mathcal S(1)
=\frac{(q-c)^2}{q}.
\label{eq:scalar-gain}
\end{equation}
If $q=0$, every forecast is zero.
Then the score is zero for every $\gamma$, and the optimizer is not identified.

\paragraph{Proof.}
Expanding the squared error gives $\mathcal S(\gamma)=2\gamma c-\gamma^2q$.
For $q>0$, completing the square gives
$\mathcal S(\gamma)=c^2/q
-q(\gamma-c/q)^2$.
This is a concave parabola with vertex $c/q$.
Hence its maximizer on a closed interval is the projection of the vertex onto that interval.
At $\gamma=1$ the score is $2c-q$.
Subtracting it from $c^2/q$ gives \eqref{eq:scalar-gain}.
\hfill\rule{0.5em}{0.5em}

We now apply this result to three cases.
First, consider a single nonzero replicate $\mathbf p_m$.
Define $a_m=\langle\mathbf y,\mathbf p_m\rangle/
(\|\mathbf y\|\|\mathbf p_m\|)$ and
$g_m=\|\mathbf p_m\|/\|\mathbf y\|$.
Then $c=a_mg_m$ and $q=g_m^2$, so the unconstrained optimum is $a_m/g_m$.
It follows that the optimal scale shrinks the forecast if $g_m>a_m$ and expands it if $g_m<a_m$.
Second, consider the average replicate score under one common scale.
Here $c=ag$ and $q=g^2+u$, which gives
\begin{equation}
\gamma^*_{\mathrm{rep}}=\frac{ag}{g^2+u},
\qquad
\max_{\gamma}\overline{R^2}(\gamma)
=\frac{a^2g^2}{g^2+u}.
\label{eq:replicate-calibration}
\end{equation}
Third, consider the ensemble mean with $g>0$.
Here $c=ag$ and $q=g^2$.
Hence $\gamma^*_{\mathrm{ens}}=a/g$, and the maximum score is $a^2$.
The gap between the two maxima is
\begin{equation}
\frac{a^2u}{g^2+u}.
\label{eq:calibration-instability-gap}
\end{equation}
This gap is the loss that instability causes even after optimal rescaling.

The optima above are in-sample oracle diagnostics, because $c$ and $q$ use the evaluation outcomes.
A deployable scaling must use only information that is available before each forecast date.

\subsection{Rank reversal and the complexity phase boundary}
\label{sec:theory-app-memory}
\label{sec:theory-complexity}

We now give the details of the phase boundary in Section~\ref{sec:theory-main-memory}.
Let $\mathcal M_0$ be a simple model class and $\mathcal M_1$ a more complex class.
Let $\mathcal F_k$ be the set of predictors that class $k\in\{0,1\}$ can fit.
We call the classes nested if $\mathcal F_0\subseteq\mathcal F_1$.
Let $\mathcal L_k(h)$ be the excess loss of class $k$ at memory $h$.
This is the expected squared error of the fitted forecast against the current conditional mean, so the irreducible noise drops out.
We assume the leading expansions
\begin{align}
\mathcal L_k(h)
&=\mathcal A_k+B_k+\frac{V_k}{h}
 +K_kh+W_kh^2+e_k(h),\notag\\
|e_k(h)|&\le r_k(h),
\label{eq:class-risk}
\end{align}
for $k\in\{0,1\}$.
These are the expansions~\eqref{eq:main-class-risk} of Section~\ref{sec:theory-main-memory}.
The terms have the following meaning.
The oracle approximation error $\mathcal A_k$ is the smallest error that a predictor in $\mathcal F_k$ can attain against the current conditional mean.
The term $B_k$ is a bias that does not depend on the history.
The term $V_k$ scales with the noise variance and the effective dimension.
The terms $K_k$ and $W_k$ are the coefficients of the linear and quadratic drift terms after filtering.
Regularization and model type can change both $K_k$ and $W_k$, and $K_k$ may be negative.
The remainder $e_k(h)$ has the known bound $r_k(h)$.
We show in Section~\ref{sec:theory-memory} that the ridge risk has this form with explicit coefficients.
In the nested case, $\mathcal F_0\subseteq\mathcal F_1$ implies $\mathcal A_1\le\mathcal A_0$.
It does not order the other terms.
Set
\begin{align}
\Delta_0&=(\mathcal A_0+B_0)-(\mathcal A_1+B_1),\notag\\
\Delta V&=V_1-V_0,\quad \Delta K=K_1-K_0,\notag\\
\Delta W&=W_1-W_0.
\label{eq:complexity-deltas}
\end{align}
The complex class gains $\Delta_0$ in the history-independent loss.
It pays $\Delta V/h+\Delta Kh+\Delta Wh^2$ in the history-dependent terms.

We prove Proposition~2 of Section~\ref{sec:theory-main-memory} in two parts.
We first state the conditions under which each class wins (Proposition~\ref{res:phase}).
The remainder bounds in these conditions form the band of the illustration (Fig.~\ref{fig:theory}A).
We then analyze the leading approximation (Proposition~\ref{res:phase-leading}).

\appresult{Proposition}{complexity phase boundary}\label{res:phase}
The complex class has strictly smaller loss whenever
\begin{equation}
\boxed{\Delta_0>\frac{\Delta V}{h}
+\Delta Kh+\Delta Wh^2+r_1(h)+r_0(h).}
\label{eq:flexible-wins}
\end{equation}
The simple class has strictly smaller loss whenever
\begin{equation}
\Delta_0<\frac{\Delta V}{h}+\Delta Kh+\Delta Wh^2
-r_1(h)-r_0(h).
\label{eq:simple-wins}
\end{equation}
Between these two inequalities, the leading expansion alone does not determine the ranking.

\paragraph{Proof.}
We subtract \eqref{eq:class-risk} for $k=0$ from \eqref{eq:class-risk} for $k=1$:
\begin{equation*}
\mathcal L_1-\mathcal L_0
=-\Delta_0+\frac{\Delta V}{h}+\Delta Kh+\Delta Wh^2
+e_1-e_0.
\end{equation*}
Since $|e_1-e_0|\le r_1+r_0$, the difference is negative under \eqref{eq:flexible-wins} and positive under \eqref{eq:simple-wins}.
\hfill\rule{0.5em}{0.5em}

\appresult{Proposition}{leading-approximation phase boundary}\label{res:phase-leading}
We now ignore the remainders.
Suppose $\Delta V>0$ and $\Delta W>0$.
The complexity cost $\phi(h)=\Delta V/h+\Delta Kh+\Delta Wh^2$ is then strictly convex in $h>0$.
Its unique minimizer $h_{\mathrm{cost}}$ and its minimum $\phi_{\min}$ satisfy
\begin{align}
2\Delta W h_{\mathrm{cost}}^3
+\Delta K h_{\mathrm{cost}}^2-\Delta V&=0,\notag\\
\phi_{\min}&=\phi(h_{\mathrm{cost}}).
\label{eq:phase-minimum}
\end{align}
If $\Delta K=0$, these reduce to
$h_{\mathrm{cost}}=(\Delta V/2\Delta W)^{1/3}$ and
$\phi_{\min}=3\cdot2^{-2/3}
(\Delta V)^{2/3}(\Delta W)^{1/3}$.
In this leading approximation, the complex class wins at memory $h$ if and only if $\Delta_0>\phi(h)$ (Fig.~\ref{fig:theory}A).
The phase boundaries are the positive roots of
\begin{equation}
\Delta Wh^3+\Delta Kh^2-\Delta_0h+\Delta V=0.
\label{eq:phase-roots}
\end{equation}
Three cases follow.
If $\Delta_0<\phi_{\min}$, the complex class never wins.
If $\Delta_0=\phi_{\min}$, the two curves touch at $h_{\mathrm{cost}}$.
If $\Delta_0>\phi_{\min}$, the complex class wins on an interval between the two positive roots of \eqref{eq:phase-roots}.
Only the part of this interval with admissible $h\ge1$ is relevant.

\paragraph{Proof.}
The second derivative is $\phi''(h)=2\Delta V/h^3+2\Delta W>0$, so $\phi$ is strictly convex on $h>0$.
Hence the first derivative $\phi'(h)=-\Delta V/h^2+\Delta K+2\Delta Wh$ is strictly increasing.
It tends to $-\infty$ as $h\to0^+$ and to $+\infty$ as $h\to\infty$, so it has exactly one zero $h_{\mathrm{cost}}$.
Multiplying $\phi'(h_{\mathrm{cost}})=0$ by $h_{\mathrm{cost}}^2$ gives the cubic in \eqref{eq:phase-minimum}.
Next, the complex class wins at memory $h$ if and only if $\Delta_0>\phi(h)$.
Multiplying this inequality by $h>0$ gives $\Delta Wh^3+\Delta Kh^2-\Delta_0h+\Delta V<0$.
The left side equals $h(\phi(h)-\Delta_0)$, so its zeros solve $\phi(h)=\Delta_0$, which is \eqref{eq:phase-roots}.
Since $\phi$ is strictly convex and $\phi(h)\to\infty$ as $h\to0^+$ and as $h\to\infty$, the equation $\phi(h)=\Delta_0$ has no solution if $\Delta_0<\phi_{\min}$.
It has the single solution $h_{\mathrm{cost}}$ if $\Delta_0=\phi_{\min}$, and two positive solutions if $\Delta_0>\phi_{\min}$.
In the last case, $\phi(h)<\Delta_0$ holds exactly between the two solutions.
This gives the three cases.
Finally, let $\Delta K=0$.
Then $h_{\mathrm{cost}}^3=\Delta V/(2\Delta W)$, so $\Delta V/h_{\mathrm{cost}}=2\Delta Wh_{\mathrm{cost}}^2$.
Hence $\phi_{\min}=3\Delta Wh_{\mathrm{cost}}^2$, and $h_{\mathrm{cost}}^2=(\Delta V/2\Delta W)^{2/3}$ gives $\phi_{\min}=3\cdot2^{-2/3}(\Delta V)^{2/3}(\Delta W)^{1/3}$.
\hfill\rule{0.5em}{0.5em}

Two degenerate cases with $\Delta K=0$ and $\Delta_0>0$ are simple.
If $\Delta W=0$ and $\Delta V>0$, the complex class wins for $h>\Delta V/\Delta_0$.
If $\Delta V=0$ and $\Delta W>0$, it wins for $h<\sqrt{\Delta_0/\Delta W}$.
In the first case, $\phi(h)=\Delta V/h$, and in the second case, $\phi(h)=\Delta Wh^2$.
The U-shaped pattern above requires $\Delta W>0$.
However, $\Delta W$ can be negative if regularization or model type filters drifting directions more strongly in the complex class.
In that case, the U-shaped characterization does not apply.
Thus parameter count alone decides neither side of the comparison in \eqref{eq:flexible-wins} and \eqref{eq:simple-wins}.

\subsection{Exact prediction risk and optimal memory for ridge regression}
\label{sec:theory-app-ridge}

We give the details of Section~\ref{sec:theory-main-ridge} here.
We first bound the risk of ridge regression under drift and serial dependence (Section~\ref{sec:theory-ridge}).
We then make the memory dependence exact and prove Proposition~3 of Section~\ref{sec:theory-main-ridge} (Section~\ref{sec:theory-memory}).

\subsubsection{Direction-aware ridge risk under temporal dependence}
\label{sec:theory-ridge}

We now study ridge regression when the coefficients drift, as in Section~\ref{sec:theory-main-ridge}.
Let $t$ be the forecast date.
The model trains on the $h$ observations that precede $t$, indexed by $j=1,\ldots,h$.
Observation $t-j$ has predictor vector $x_{t-j}\in\mathrm R^p$ and outcome $y_{t-j}\in\mathrm R$.
We assume
\begin{equation}
y_{t-j}=x_{t-j}^{\top}\beta_{t-j}+\varepsilon_{t-j},
\qquad
\beta_{t-j}=\beta_t-j\dot\beta_t+r_{t,j}.
\label{eq:local-drift}
\end{equation}
Here $\beta_s\in\mathrm R^p$ is the coefficient vector at date $s$, and $\varepsilon_s$ is the regression error.
The vector $\dot\beta_t$ is the per-period drift of the coefficients near date $t$.
The term $r_{t,j}$ is the remainder that this linear drift leaves out.
The estimation target is the current coefficient $\beta_t$.
We condition on the realized training design $X_h=(x_{t-1},\ldots,x_{t-h})$.
We treat the coefficient path in \eqref{eq:local-drift} as fixed, or as a function of $X_h$.

We define the sample second-moment matrix $S_h$, the ridge inverse $A_{h,\lambda}$, the ridge filter $G_{h,\lambda}$, and the mean lag $\mu_h$:
\begin{align}
S_h&=\frac1h\sum_{j=1}^h x_{t-j}x_{t-j}^{\top},
\quad A_{h,\lambda}=(S_h+\lambda I)^{-1},\notag\\
G_{h,\lambda}&=A_{h,\lambda}S_h,
\qquad \mu_h=\frac{h+1}{2},
\label{eq:ridge-operators}
\end{align}
where $\lambda>0$ is the ridge penalty.
The ridge estimator is
\begin{equation}
\widehat\beta_{t,h,\lambda}
=\arg\min_b\left\{\frac1h\sum_{j=1}^h
 (y_{t-j}-x_{t-j}^{\top}b)^2+\lambda\|b\|^2\right\}.
\label{eq:ridge-objective}
\end{equation}
Three further quantities enter the estimation error:
\begin{align}
H_h&=\frac1h\sum_{j=1}^h jx_{t-j}x_{t-j}^{\top}-\mu_hS_h,
\label{eq:design-imbalance}\\
q_{t,h}&=\frac1h\sum_{j=1}^h
 x_{t-j}x_{t-j}^{\top}r_{t,j},
\qquad
z_{t,h}=\frac1h\sum_{j=1}^h x_{t-j}\varepsilon_{t-j}.
\label{eq:curvature-score}
\end{align}
We call $H_h$ the time-weighted design imbalance.
It vanishes if every lag carries the same second-moment matrix.
We call $q_{t,h}$ the curvature aggregate, because it sums the design-weighted remainders $r_{t,j}$.
We call $z_{t,h}$ the score.
Its conditional mean is $\bar z_{t,h}=\operatorname{E}[z_{t,h}\mid X_h]$.

We assume that the centered conditional score variance obeys
\begin{equation}
\operatorname{Var}(z_{t,h}\mid X_h)
\preceq\frac{\kappa_h\sigma_t^2}{h}S_h.
\tag{A1}\label{eq:score-variance}
\end{equation}
Here $\sigma_t^2>0$ is the error variance near date $t$, and $\kappa_h\ge1$ is a dependence factor.
We use \eqref{eq:score-variance} to bound the centered long-run score variance.
We do not set the conditional score mean to zero.
The stronger condition $\bar z_{t,h}=0$ is strict score exogeneity at the window level.
It is strong because $X_h$ contains regressors dated after some of the errors.
Predetermined predictors, lagged returns, and persistent predictors with return-correlated innovations need not satisfy it \citep{stambaugh1999predictive}.
For this reason, we keep $\bar z_{t,h}$ in the bound.
If the errors are independent of the whole design, conditionally homoskedastic, and conditionally serially uncorrelated, then $\kappa_h=1$.
Serial dependence can require a bounded inflation factor $\kappa_h>1$.

For a vector $v$, write $\|v\|_{S_h}^2=v^{\top}S_hv$.
This local design norm measures the squared prediction difference, averaged over the training rows.
We split the estimation risk into four components:
\begin{align}
B_{\mathrm{reg},t}(h,\lambda)
&=\|(I-G_{h,\lambda})\beta_t\|_{S_h}^2,\notag\\
d_{2,h}(\lambda)
&=\operatorname{tr}\!\left[S_h^2(S_h+\lambda I)^{-2}\right],\notag\\
D_{t,h}^2(\lambda)
&=\|G_{h,\lambda}\dot\beta_t\|_{S_h}^2,\notag\\
\rho_{t,h}^2(\lambda)
&=\|A_{h,\lambda}
  (q_{t,h}-H_h\dot\beta_t+\bar z_{t,h})\|_{S_h}^2.
\label{eq:ridge-components}
\end{align}
The term $B_{\mathrm{reg},t}$ is the regularization bias, the part of $\beta_t$ that ridge shrinks away.
The term $d_{2,h}$ is the ridge-filtered spectral dimension, which scales the variance.
The term $D_{t,h}^2$ is the drift energy that the ridge filter passes through.
The term $\rho_{t,h}^2$ is a remainder.
It records three departures from the ideal case: curvature in the coefficient path, time imbalance in the realized design, and a nonzero conditional score mean.

In Section~\ref{sec:theory-main-ridge}, we suppress the penalty $\lambda$ and the date $t$ in the notation.
We write $A_h$, $G_h$, $z$, $B_{\rm reg}$, $D^2$, $\rho^2$, and $\mathcal E_t^X(h)$ there for $A_{h,\lambda}$, $G_{h,\lambda}$, $z_{t,h}$, $B_{\mathrm{reg},t}(h,\lambda)$, $D_{t,h}^2(\lambda)$, $\rho_{t,h}^2(\lambda)$, and $\mathcal E_t^X(h,\lambda)$.
We keep the full notation here.

\appresult{Proposition}{conditional ridge identity and bound}\label{res:ridge}
The conditional excess prediction risk in the local design norm, $\mathcal E_t^X(h,\lambda)$, satisfies the exact identity
\begin{align}
\mathcal E_t^X(h,\lambda)
&:=\operatorname{E}\!\left[
 \|\widehat\beta_{t,h,\lambda}-\beta_t\|_{S_h}^2
 \mid X_h\right]\notag\\
&=\Big\|(I-G_{h,\lambda})\beta_t
 +\mu_hG_{h,\lambda}\dot\beta_t\notag\\
&\qquad
 -A_{h,\lambda}
  (q_{t,h}-H_h\dot\beta_t+\bar z_{t,h})
 \Big\|_{S_h}^2\notag\\
&\quad+\operatorname{tr}\!\left[
 S_hA_{h,\lambda}\operatorname{Var}(z_{t,h}\mid X_h)
 A_{h,\lambda}\right].
\label{eq:ridge-exact}
\end{align}
Under \eqref{eq:score-variance}, it also satisfies, for every $\eta,\zeta>0$,
\begin{align}
\mathcal E_t^X(h,\lambda)
&\le (1+\eta)(1+\zeta)
 B_{\mathrm{reg},t}(h,\lambda)\notag\\
&\quad +(1+\eta)(1+\zeta^{-1})
 \mu_h^2D_{t,h}^2(\lambda)\notag\\
&\quad +(1+\eta^{-1})\rho_{t,h}^2(\lambda)
 +\frac{\kappa_h\sigma_t^2}{h}d_{2,h}(\lambda).
\label{eq:ridge-bound}
\end{align}

\paragraph{Proof.}
First we expand the estimator.
The ridge normal equations and \eqref{eq:local-drift} give
\begin{align*}
\widehat\beta_{t,h,\lambda}
&=A_{h,\lambda}\bigl[S_h\beta_t
 -(\mu_hS_h+H_h)\dot\beta_t
 +q_{t,h}+z_{t,h}\bigr].
\end{align*}
Subtracting $\beta_t$ yields
\begin{align*}
\widehat\beta_{t,h,\lambda}-\beta_t
&=-(I-G_{h,\lambda})\beta_t
 -\mu_hG_{h,\lambda}\dot\beta_t\\
&\quad+A_{h,\lambda}
 (q_{t,h}-H_h\dot\beta_t+\bar z_{t,h})\\
&\quad+A_{h,\lambda}(z_{t,h}-\bar z_{t,h}).
\end{align*}
The first three terms are functions of $X_h$.
The last term has conditional mean zero given $X_h$.
Hence the conditional risk equals the squared norm of the first three terms plus the variance of the last term.
This is \eqref{eq:ridge-exact}.
The identity needs no bound on the variance.
We use \eqref{eq:score-variance} only in the final step.

To prove \eqref{eq:ridge-bound}, we use the inequality
$\|v+w\|^2\le(1+c)\|v\|^2+(1+c^{-1})\|w\|^2$, which holds for every $c>0$.
We apply it twice.
The first application, with $c=\eta$, separates the remainder from the leading bias.
The second application, with $c=\zeta$, separates the regularization bias from the filtered drift.
For the variance term, $A_{h,\lambda}S_hA_{h,\lambda}$ is positive semidefinite.
Hence \eqref{eq:score-variance} and the cyclicity of the trace bound the variance term.
Because $S_h$ and $A_{h,\lambda}$ commute, the bound equals
\begin{equation*}
\frac{\kappa_h\sigma_t^2}{h}
\operatorname{tr}\!\left[S_h^2(S_h+\lambda I)^{-2}\right].
\end{equation*}
Collecting the terms proves \eqref{eq:ridge-bound}.
\hfill\rule{0.5em}{0.5em}

Two features of \eqref{eq:ridge-bound} matter for the choice of memory (Fig.~\ref{fig:theory}B).
First, the variance term scales with the ridge-filtered dimension $d_{2,h}(\lambda)$ and decays like $h^{-1}$.
If $\sup_h\kappa_h<\infty$, serial dependence preserves this $h^{-1}$ order.
Second, the obsolete-data term scales with $\mu_h^2D_{t,h}^2(\lambda)$, which uses the filtered drift $G_{h,\lambda}\dot\beta_t$.
Ridge shrinks drift in heavily shrunk or low-variance directions.
Thus such drift adds little to the risk in the local prediction norm.

We now connect the local norm to next-period risk.
Let $\Sigma_t$ be the second-moment matrix of the predictor at the forecast date, conditional on $X_h$.
We measure next-period excess prediction risk by
$\operatorname{E}[\|\widehat\beta_{t,h,\lambda}-\beta_t\|_{\Sigma_t}^2\mid X_h]$.
Suppose $\Sigma_t\preceq c_+S_h$ for a constant $c_+>0$.
Then this risk is at most $c_+\mathcal E_t^X(h,\lambda)$.
Hence $c_+$ times the right side of \eqref{eq:ridge-bound} also bounds the next-period risk.
If the covariance is locally stable, one would expect $S_h$, $d_{2,h}$, and $D_{t,h}$ to approach their local population counterparts.
Finally, exact local linearity, a time-balanced design, and $\bar z_{t,h}=0$ together imply $\rho_{t,h}=0$.

\subsubsection{Optimal memory and minimized risk}
\label{sec:theory-memory}

The bound \eqref{eq:ridge-bound} controls the risk from above.
It does not identify the risk-minimizing memory.
An exact memory calculation needs stronger assumptions than \eqref{eq:score-variance}.
Hence we study a simplified family of candidate histories.
All histories share the same local second-moment matrix $S$, coefficient $\beta_t$, drift $\dot\beta_t$, and penalty $\lambda$.
We also assume $\rho_{t,h}=0$, that is, $q_{t,h}-H_h\dot\beta_t+\bar z_{t,h}=0$.
This holds, for example, if $q_{t,h}$, $H_h$, and $\bar z_{t,h}$ all vanish.
The centered score covariance equals exactly $\kappa\sigma_t^2S/h$ for a constant $\kappa>0$.
These assumptions define a simplified comparison.
We do not claim them for arbitrary nested historical windows.
Write $G=(S+\lambda I)^{-1}S$ and $d_2(\lambda)=\operatorname{tr}[S^2(S+\lambda I)^{-2}]$.
Let $\langle u,v\rangle_S=u^{\top}Sv$.
Define
\begin{equation}
\begin{aligned}
B_t&=\|(I-G)\beta_t\|_S^2,\\
J_t&=\langle(I-G)\beta_t,G\dot\beta_t\rangle_S,\\
Q_t&=\|G\dot\beta_t\|_S^2,\qquad
V_t=\kappa\sigma_t^2d_2(\lambda).
\end{aligned}
\label{eq:memory-exact-components}
\end{equation}
Here $B_t$ is the regularization bias, $Q_t$ is the filtered drift energy, $J_t$ is their cross term, and $V_t$ is the variance scale.
Under these assumptions, \eqref{eq:ridge-exact} reduces to the exact interpolation
\begin{equation}
\mathcal E_t^\circ(h,\lambda)
=B_t+2\mu_hJ_t+\mu_h^2Q_t+\frac{V_t}{h},
\qquad \mu_h=\frac{h+1}{2},
\label{eq:memory-stable-risk}
\end{equation}
where $\mathcal E_t^\circ(h,\lambda)$ is the conditional risk under these assumptions.
This is the risk~\eqref{eq:main-risk} of Section~\ref{sec:theory-main-ridge}, which we write there as $\mathcal E_t(h)$.

The risk~\eqref{eq:memory-stable-risk} has the form~\eqref{eq:class-risk} of Section~\ref{sec:theory-complexity}.
Indeed, $2\mu_hJ_t=(h+1)J_t$ and $\mu_h^2Q_t=(h+1)^2Q_t/4$.
Collecting the terms in $h$, we can take
\begin{align}
B_k&=B_t+J_t+Q_t/4,\qquad V_k=V_t,\notag\\
K_k&=J_t+Q_t/2,\qquad W_k=Q_t/4,
\label{eq:ridge-class-coefficients}
\end{align}
and we treat any oracle approximation error separately in $\mathcal A_k$.
These are the coefficients that Section~\ref{sec:theory-main-ridge} states.
This choice keeps the known linear term $K_kh$ explicit.
A quadratic-only surrogate would move this term into $e_k$.
It would then leave a remainder of order $h$, which can make a narrow ranking margin uninformative.

We restate Proposition~3 of Section~\ref{sec:theory-main-ridge} and prove it.

\appresult{Corollary}{cross-term-aware interior memory}\label{res:interior-memory}
Assume $V_t>0$ and $Q_t>0$.
Extend \eqref{eq:memory-stable-risk} to real $h>0$.
The extended risk is strictly convex and has a unique positive minimizer $h_t^*$ (Fig.~\ref{fig:theory}B).
It satisfies
\begin{equation}
\boxed{\frac{V_t}{(h_t^*)^2}
=J_t+\frac{h_t^*+1}{2}Q_t.}
\label{eq:optimal-memory}
\end{equation}
If $J_t=0$, this condition reduces to the exact finite-$h$ cubic
\begin{equation}
(h_t^*)^2(h_t^*+1)=\frac{2V_t}{Q_t}.
\label{eq:memory-cubic}
\end{equation}
If $h_t^*$ is large and $J_t$ is negligible relative to $h_t^*Q_t$, then $h_t^*\sim(2V_t/Q_t)^{1/3}$.
The sign of $J_t$ shifts the optimum.
A negative $J_t$ lengthens the optimal history, and a positive $J_t$ shortens it.
The cube-root formula is only approximate.
It fails for general ridge shrinkage and for short histories.

\paragraph{Proof.}
Differentiating \eqref{eq:memory-stable-risk} gives
$-V_t/h^2+J_t+(h+1)Q_t/2$.
Its derivative is $2V_t/h^3+Q_t/2>0$, so the first derivative is strictly increasing.
It tends to $-\infty$ as $h\to0^+$ and to $+\infty$ as $h\to\infty$.
Hence it crosses zero exactly once on $(0,\infty)$, and this zero is the unique minimizer.
Setting $J_t=0$ and multiplying the first-order condition by $2h^2$ gives \eqref{eq:memory-cubic}.
\hfill\rule{0.5em}{0.5em}

Integer histories need a final step.
For admissible histories $h\ge1$, we clip the continuous optimum to the permitted interval before we compare neighboring integers.
In particular, we use $h=1$ if $h_t^*<1$.
If $Q_t=0$, then $J_t=0$ by the Cauchy--Schwarz inequality in the $S$-inner product.
In that case, any positive $V_t$ favors the longest available history.
Regularization changes $V_t$, $Q_t$, and $J_t$ jointly.

A simpler surrogate keeps only the leading terms.
We drop $J_t$ and replace $\mu_h^2$ by $h^2/4$.
The memory-dependent part then equals $V_t/h+W_th^2$ with $W_t=Q_t/4$.
Its minimum is
\begin{equation}
\min_{h>0}\left\{\frac{V_t}{h}+W_th^2\right\}
=3\cdot2^{-2/3}V_t^{2/3}W_t^{1/3}.
\label{eq:minimized-memory-risk}
\end{equation}
This value is the minimum of the surrogate, not of \eqref{eq:memory-stable-risk}.

\subsection{Surrogate approximations and transfer bounds for trees and neural networks}
\label{sec:theory-app-transfer}

We give the details of Section~\ref{sec:theory-main-transfer} here.
Boosted trees, RNNs, LSTMs, and Transformers are not linear in their parameters.
Hence we analyze a surrogate that fixes the feature map before we fit the readout.
We bound the risk of this surrogate for frozen feature maps and fixed-topology boosting (Section~\ref{sec:theory-fixed-representations}), and we derive its exact spectral phase boundary (Section~\ref{sec:theory-spectral-phase}).
We then bound the gap between the surrogate and the trained model (Section~\ref{sec:theory-architecture}).

\subsubsection{Frozen feature maps and fixed-tree boosting}
\label{sec:theory-fixed-representations}

The ridge argument of Section~\ref{sec:theory-ridge} extends exactly to a feature map that we fix before we observe the outcomes that fit its readout.
Here ``fixed'' means that the parameters of the map do not change during the readout fit.
The output of the map may still depend on the input sequence.
We need this condition because a feature map that we select on the same responses as its readout is not fixed.
For trees, the feature map is the tree topology.
For the neural models, it is the encoder that produces the final hidden state or the pooled token state.

Let $v_s$ be the raw input at date $s$.
It can be a vector or a sequence of length $L$.
Let $\varphi$ map an input into $\mathrm R^p$, and call $\varphi(v_s)$ the feature vector.
We may choose $\varphi$ from an independent or an earlier sample.
Let $\mathcal G_h$ be a sigma-field that contains the chosen map $\varphi$ and the current design $(v_{t-1},\ldots,v_{t-h})$.
We condition on $\mathcal G_h$ and write $z_j=\varphi(v_{t-j})$.
We assume that the coefficient path, the approximation residuals, and all matrices below are $\mathcal G_h$-measurable.
We also assume
\begin{align}
y_{t-j}&=z_j^{\top}\theta_{t-j}+b_{t,j}+\varepsilon_{t-j},\notag\\
\theta_{t-j}&=\theta_t-j\dot\theta_t+r_{t,j},
\qquad j=1,\ldots,h.
\label{eq:frozen-feature-model}
\end{align}
Here $\theta_s\in\mathrm R^p$ is the feature coefficient at date $s$, $\dot\theta_t$ is its drift per period, and $r_{t,j}$ is the remainder of the linear drift.
The term $b_{t,j}$ is the approximation residual, the part of the conditional mean that the features cannot represent.

We now make $\theta_s$ and $b_{t,j}$ unique.
Let $P_s$ be the distribution of the input $v_s$, and let $m_s(v)=\operatorname{E}[y_s\mid v_s=v]$ be the conditional mean function at date $s$.
Write $\operatorname{E}_s$ for expectation under $P_s$, and assume $\operatorname{E}_s[\varphi\varphi^\top]\succ0$.
We define $\theta_s$ as the unique $L_2(P_s)$ projection coefficient of $m_s$ onto $\varphi$.
We define $b_s(v)=m_s(v)-\varphi(v)^\top\theta_s$.
Then $\operatorname{E}_s[\varphi b_s]=0$ and $b_{t,j}=b_{t-j}(v_{t-j})$.
This orthogonality holds in the population.
It need not hold in a realized historical design.

We define the feature second-moment matrix, the ridge inverse and filter, the time imbalance, the aggregate bias, and the score:
\begin{align}
S_h^\varphi&=\frac1h\sum_j z_jz_j^{\top},
\quad A_h^\varphi=(S_h^\varphi+\lambda I)^{-1},\notag\\
G_h^\varphi&=A_h^\varphi S_h^\varphi,
\quad H_h^\varphi=\frac1h\sum_j jz_jz_j^{\top}-\mu_hS_h^\varphi,\notag\\
q_h^\varphi&=\frac1h\sum_j z_j
  (z_j^{\top}r_{t,j}+b_{t,j}),\notag\\
s_h^\varphi&=\frac1h\sum_j z_j\varepsilon_{t-j},
\qquad \mu_h=\frac{h+1}{2},\notag\\
\bar s_h^\varphi&=
 \operatorname{E}[s_h^\varphi\mid\mathcal G_h].
\label{eq:frozen-feature-operators}
\end{align}
We fit the readout $\widehat\theta_h$ by ridge regression of $y_{t-j}$ on $z_j$ with penalty $\lambda\|\theta\|^2$ and $\lambda>0$.
That is, $\widehat\theta_h$ minimizes $h^{-1}\sum_{j=1}^h(y_{t-j}-z_j^{\top}\theta)^2+\lambda\|\theta\|^2$.
We assume, with $\kappa_h$ and $\sigma_t^2$ as in \eqref{eq:score-variance},
\begin{equation}
\operatorname{Var}(s_h^\varphi\mid\mathcal G_h)
\preceq\frac{\kappa_h\sigma_t^2}{h}S_h^\varphi.
\label{eq:frozen-score-condition}
\end{equation}
We do not assume a zero conditional score mean.
The term $\bar s_h^\varphi$ carries any whole-window score bias, including the bias that persistent endogenous predictors induce.
Let $M_t\succeq0$ be any $\mathcal G_h$-measurable matrix that defines the prediction norm, and write $\|v\|_{M_t}^2=v^{\top}M_tv$.

\appresult{Proposition}{frozen-feature-map ridge bound}\label{res:frozen}
Assume the model \eqref{eq:frozen-feature-model}, the projection identification above, and \eqref{eq:frozen-score-condition}.
The conditional risk of the fitted readout for the represented target $\varphi(\cdot)^{\top}\theta_t$ is then exactly
\begin{align}
\mathcal E_t^\varphi(h,\lambda;M_t)
&:=\operatorname{E}\!\left[
 \|\widehat\theta_h-\theta_t\|_{M_t}^2
 \mid\mathcal G_h\right]\notag\\
&=\Big\|(I-G_h^\varphi)\theta_t
 +\mu_hG_h^\varphi\dot\theta_t\notag\\
&\qquad-A_h^\varphi
   (q_h^\varphi-H_h^\varphi\dot\theta_t+\bar s_h^\varphi)
  \Big\|_{M_t}^2\notag\\
&\quad+\operatorname{tr}\!\left[
 M_tA_h^\varphi
 \operatorname{Var}(s_h^\varphi\mid\mathcal G_h)
 A_h^\varphi\right].
\label{eq:frozen-exact}
\end{align}
Set
\begin{align}
B_h^\varphi&=\|(I-G_h^\varphi)\theta_t\|_{M_t}^2,
\quad D_{h,\varphi}^2=\|G_h^\varphi\dot\theta_t\|_{M_t}^2,\notag\\
\rho_{h,\varphi}^2&=\|A_h^\varphi
 (q_h^\varphi-H_h^\varphi\dot\theta_t
  +\bar s_h^\varphi)\|_{M_t}^2,\notag\\
d_h^\varphi&=\operatorname{tr}\!\left[
 M_tA_h^\varphi S_h^\varphi A_h^\varphi\right].
\label{eq:frozen-risk-components}
\end{align}
Then, for every $\eta,\zeta>0$,
\begin{align}
\mathcal E_t^\varphi(h,\lambda;M_t)
&\le(1+\eta)(1+\zeta)B_h^\varphi\notag\\
&\quad+(1+\eta)(1+\zeta^{-1})
 \mu_h^2D_{h,\varphi}^2\notag\\
&\quad+(1+\eta^{-1})\rho_{h,\varphi}^2
 +\frac{\kappa_h\sigma_t^2}{h}d_h^\varphi.
\label{eq:frozen-bound}
\end{align}

\paragraph{Proof.}
The ridge normal equations give
\begin{align*}
\widehat\theta_h-\theta_t
&=-(I-G_h^\varphi)\theta_t
 -\mu_hG_h^\varphi\dot\theta_t\\
&\quad+A_h^\varphi
 (q_h^\varphi-H_h^\varphi\dot\theta_t)
 +A_h^\varphi s_h^\varphi.
\end{align*}
We center $s_h^\varphi$ at $\bar s_h^\varphi$ and take the conditional squared $M_t$-norm.
This proves \eqref{eq:frozen-exact}.
Two applications of Young's inequality bound the squared bias, as in the proof of the ridge bound in Section~\ref{sec:theory-ridge}.
The covariance order in \eqref{eq:frozen-score-condition} bounds the trace by $\kappa_h\sigma_t^2d_h^\varphi/h$.
This proves \eqref{eq:frozen-bound}.
\hfill\rule{0.5em}{0.5em}

Two choices of $M_t$ are useful.
If $M_t=S_h^\varphi$, the variance term becomes
$\operatorname{tr}[(S_h^\varphi)^2(S_h^\varphi+\lambda I)^{-2}]$.
If $M_t$ is the next-period feature covariance, the bound measures risk in the norm that matters for forecasting.
The bound concerns the represented target only.
A bound for the full conditional mean must also include the approximation residual at the forecast date.
To see this, take a $\mathcal G_h$-measurable evaluation design $Z_*\in\mathrm R^{n_*\times p}$ with $n_*$ rows of feature vectors.
Let $b_*\in\mathrm R^{n_*}$ be the approximation residuals at these rows, and set $M_t=Z_*^{\top}Z_*/n_*$.
For any $\tau>0$, the prediction error for the full conditional mean satisfies
\begin{align}
&\operatorname{E}\!\left[
 \|Z_*(\widehat\theta_h-\theta_t)-b_*\|^2/n_*
 \mid\mathcal G_h\right]\notag\\
&\quad\le(1+\tau)\mathcal E_t^\varphi(h,\lambda;M_t)
 +(1+\tau^{-1})\|b_*\|^2/n_*.
\label{eq:frozen-full-risk}
\end{align}
This follows from the same two-term norm inequality $\|v+w\|^2\le(1+\tau)\|v\|^2+(1+\tau^{-1})\|w\|^2$.

\appresult{Corollary}{tree and sequence feature maps}\label{res:frozen-maps}
Assume that the feature map does not depend on the readout outcomes and that \eqref{eq:frozen-score-condition} holds.
Then the bound \eqref{eq:frozen-bound} of Proposition~\ref{res:frozen} holds for each of the following maps:
\begin{itemize}
\item \emph{Prechosen trees.}
Let $\varphi_{\rm tree}(x)$ concatenate the leaf indicators of prechosen tree partitions.
Joint ridge fitting of the additive leaf coefficients then obeys \eqref{eq:frozen-bound}.
This fit is a fixed-tree surrogate.
It differs from the usual greedy XGBoost fit.
\item \emph{Frozen RNN or LSTM.}
Let $\varphi_{\rm RNN}(v)$ or $\varphi_{\rm LSTM}(v)$ be the final hidden state of the encoder.
A ridge readout on this state obeys \eqref{eq:frozen-bound}.
It has its own covariance spectrum and coefficient drift.
\item \emph{Frozen Transformer.}
Let $\varphi_{\rm Tr}(v)$ be the final-token state or a pooled state of the encoder.
A ridge readout on this state obeys the same bound.
Attention can depend on the input sequence, but the fitted parameters stay frozen.
\end{itemize}
We distinguish two lengths.
The memory $h$ is the number of training dates that fit the readout.
The context length $L$ is the number of time steps in the input sequence that constructs each feature vector.

Suppose that, over the candidate histories, we can approximate $d_h^\varphi$, $D_{h,\varphi}^2$, $\kappa_h$, and the remaining terms by constants.
The memory-dependent part of the bound then has the form $V_\varphi/h+K_\varphi h+W_\varphi h^2$.
This is the form~\eqref{eq:class-risk} that Section~\ref{sec:theory-main-transfer} uses for the surrogate risk.
Here $V_\varphi$ is proportional to $\kappa\sigma_t^2d^\varphi$, and $W_\varphi$ is proportional to $D_{h,\varphi}^2$.
If $\mu_h^2$ is the only source of time dependence in the bound, then $K_\varphi=2W_\varphi$ exactly.
For positive $V_\varphi$ and $W_\varphi$, this form has a unique positive continuous minimizer.
It solves
\begin{equation}
\boxed{2W_\varphi(h_\varphi^*)^3
+K_\varphi(h_\varphi^*)^2-V_\varphi=0.}
\label{eq:frozen-optimal-memory}
\end{equation}
For admissible integer histories $h\ge1$, we clip the continuous solution to the history interval and compare the neighboring integers.
This procedure minimizes the stated surrogate.
It does not minimize the risk of a retrained encoder.

We can also analyze the stagewise squared-loss updates of boosting for fixed tree topologies.
This avoids refitting all leaf coefficients jointly.
Let $N_{\rm tree}$ be the number of boosting stages.
Tree $k$ has $T_k$ leaves, and $Z_k\in\{0,1\}^{h\times T_k}$ is its matrix of leaf indicators on the $h$ training rows.
We treat each $Z_k$ as fixed given $\mathcal G_h$.
Let $\mathsf W_k$ be a nonnegative diagonal matrix of fixed row-subsampling weights, and set $\mathsf W_k=I$ for full-row fitting.
Let $\lambda_k>0$ be the $\ell_2$ leaf penalty of stage $k$, and let $0<\nu\le1$ be the learning rate.
We assume a fixed zero initial prediction and no $\ell_1$ penalty.
Define the stage-$k$ leaf smoother $P_k$ and the boosting smoother $T_h$ by
\begin{equation}
P_k=Z_k(Z_k^{\top}\mathsf W_kZ_k+\lambda_k I)^{-1}
 Z_k^{\top}\mathsf W_k,
\label{eq:boost-leaf-operator}
\end{equation}
\begin{equation}
T_h=I-\prod_{k=N_{\rm tree}}^{1}(I-\nu P_k).
\label{eq:fixed-boost-smoother}
\end{equation}
The product in \eqref{eq:fixed-boost-smoother} runs from $k=N_{\rm tree}$ on the left to $k=1$ on the right.
Let $\mathbf y_h=(y_{t-1},\ldots,y_{t-h})^{\top}$.
The fitted values at the training inputs are $T_h\mathbf y_h$.

Let $m_t=(m_t(v_{t-1}),\ldots,m_t(v_{t-h}))^{\top}$ be the current regression function evaluated at the historical inputs.
Let $\mathsf L_h=\operatorname{diag}(1,\ldots,h)$ be the matrix of lags.
We assume that the historical conditional means obey
\begin{equation}
\operatorname{E}[\mathbf y_h\mid\mathcal G_h]
=m_t-\mathsf L_h\dot m_t+r_h,
\label{eq:boost-local-drift}
\end{equation}
where we treat the drift vector $\dot m_t$ and the remainder $r_h$ as constants given $\mathcal G_h$.
The remainder $r_h$ includes any distortion of the conditional mean that comes from conditioning on the whole historical design.
Hence $r_h=0$ is a strong whole-window restriction.
It can fail with persistent endogenous predictors or lagged returns.
Define the centered residual $\varepsilon_h=\mathbf y_h-
\operatorname{E}[\mathbf y_h\mid\mathcal G_h]$.
We assume
\begin{equation}
\operatorname{E}[\varepsilon_h\mid\mathcal G_h]=0,
\qquad
\Omega_h:=\operatorname{Var}(\varepsilon_h\mid\mathcal G_h)
\preceq\kappa_h\sigma_t^2I.
\label{eq:boost-noise-condition}
\end{equation}
The mean-zero equality holds by definition of $\varepsilon_h$.
It does not impose strict exogeneity on the original return disturbance.
We can take $\kappa_h=1$, for example, if the centered errors are conditionally serially uncorrelated and conditionally homoskedastic with variance at most $\sigma_t^2$ given $\mathcal G_h$.
Independence from the design and homoskedasticity alone do not suffice, because off-diagonal covariances can remain.
We use the scaled norm $\|v\|_h^2=h^{-1}\|v\|^2$.

\appresult{Proposition}{fixed-topology boosting bound}\label{res:boost}
Assume \eqref{eq:boost-leaf-operator}--\eqref{eq:boost-noise-condition}.
The conditional local-design risk of the boosted fit is then exactly
\begin{align}
\mathcal E_t^{\rm boost}(h)
&:=\operatorname{E}\!\left[
 \|T_h\mathbf y_h-m_t\|_h^2
 \mid\mathcal G_h\right]\notag\\
&=\|(I-T_h)m_t+T_h\mathsf L_h\dot m_t-T_hr_h\|_h^2\notag\\
&\quad+\frac1h\operatorname{tr}(T_h\Omega_hT_h^{\top}).
\label{eq:boost-exact}
\end{align}
Define
\begin{align}
B_h^{\rm boost}&=\|(I-T_h)m_t\|_h^2,\notag\\
D_{h,\rm boost}^2&=\|T_h\mathsf L_h\dot m_t/\mu_h\|_h^2,\notag\\
\rho_{h,\rm boost}^2&=\|T_hr_h\|_h^2,
\quad d_h^{\rm boost}=\operatorname{tr}(T_hT_h^{\top}).
\label{eq:boost-components}
\end{align}
Then, for every $\eta,\zeta>0$,
\begin{align}
\mathcal E_t^{\rm boost}(h)
&\le(1+\eta)(1+\zeta)B_h^{\rm boost}\notag\\
&\quad+(1+\eta)(1+\zeta^{-1})
 \mu_h^2D_{h,\rm boost}^2\notag\\
&\quad+(1+\eta^{-1})\rho_{h,\rm boost}^2
 +\frac{\kappa_h\sigma_t^2}{h}d_h^{\rm boost}.
\label{eq:boost-bound}
\end{align}

\paragraph{Proof.}
Let $\widehat m_k$ be the vector of fitted values after stage $k$, with $\widehat m_0=0$.
At stage $k$, squared-loss minimization on the fixed leaves gives the fitted increment $\nu P_k(\mathbf y_h-\widehat m_{k-1})$.
Hence the residual obeys
\begin{equation*}
\mathbf y_h-\widehat m_k
=(I-\nu P_k)(\mathbf y_h-\widehat m_{k-1}).
\end{equation*}
Iterating from $\widehat m_0=0$ yields $\widehat m_{N_{\rm tree}}=T_h\mathbf y_h$.
We substitute \eqref{eq:boost-local-drift} and obtain
\begin{equation*}
T_h\mathbf y_h-m_t
=-(I-T_h)m_t-T_h\mathsf L_h\dot m_t+T_hr_h+T_h\varepsilon_h.
\end{equation*}
The noise term has conditional mean zero, which gives \eqref{eq:boost-exact}.
Two Young inequalities bound the squared bias.
The condition \eqref{eq:boost-noise-condition} bounds the variance by $\kappa_h\sigma_t^2d_h^{\rm boost}/h$.
This proves \eqref{eq:boost-bound}.
\hfill\rule{0.5em}{0.5em}

The bound above measures risk on the training inputs.
Forecasting needs the risk on a fixed common evaluation design with $n_*$ rows.
Write $\|v\|_*^2=\|v\|^2/n_*$ for the scaled norm on this design.
Let $Z_{*,k}$ be the matrix of leaf indicators that tree $k$ assigns to the evaluation rows.
Define
\begin{align}
T_{*,h}
&=\nu\sum_{k=1}^{N_{\rm tree}}
Z_{*,k}(Z_k^\top \mathsf W_kZ_k+\lambda_kI)^{-1}\notag\\
&\quad\times Z_k^\top \mathsf W_k
\prod_{\ell=k-1}^{1}(I-\nu P_\ell).
\label{eq:boost-eval-operator}
\end{align}
Here the empty product equals $I$.
This operator maps the training response to the forecast of the fixed-tree model on the evaluation design.
We call this forecast the reference forecast.
Let $m_{t,*}$ be the current mean vector on this design.
The same calculation as above gives the unclipped reference risk
\begin{align}
L_{*,h}^{0,\mathrm{unclip}}
&=\operatorname{E}[
\|T_{*,h}\mathbf y_h-m_{t,*}\|_*^2
\mid\mathcal G_h]\notag\\
&=\|T_{*,h}(m_t-\mathsf L_h\dot m_t+r_h)-m_{t,*}\|_*^2\notag\\
&\quad+n_*^{-1}
\operatorname{tr}(T_{*,h}\Omega_hT_{*,h}^\top).
\label{eq:boost-eval-risk}
\end{align}
Under \eqref{eq:boost-noise-condition}, the variance term is at most $\kappa_h\sigma_t^2d_{*,h}/h$, where
$d_{*,h}=(h/n_*)\operatorname{tr}(T_{*,h}T_{*,h}^\top)$.
For any $\eta,\zeta>0$, two Young inequalities then give the explicit upper bound
\begin{align}
U_{*,h}^{\rm fixed}
&:=(1+\eta)(1+\zeta)
\|m_{t,*}-T_{*,h}m_t\|_*^2\notag\\
&\quad+(1+\eta)(1+\zeta^{-1})
\|T_{*,h}\mathsf L_h\dot m_t\|_*^2\notag\\
&\quad+(1+\eta^{-1})\|T_{*,h}r_h\|_*^2
+\frac{\kappa_h\sigma_t^2}{h}d_{*,h}.
\label{eq:boost-eval-bound}
\end{align}
Hence $L_{*,h}^{0,\mathrm{unclip}}\le
U_{*,h}^{\rm fixed}$.

The boosting operator $T_h$ filters the drift in \eqref{eq:boost-bound}.
Consider a single fixed tree with $\mathsf W_1=I$ and $\nu=1$.
Then $d_h^{\rm boost}=\sum_{\ell}
[n_\ell/(n_\ell+\lambda_1)]^2$, where $n_\ell$ is the number of training rows in leaf $\ell$.
The leaf penalty in \eqref{eq:boost-leaf-operator} acts on the unnormalized squared-loss scale.
It differs from the normalized ridge objective \eqref{eq:ridge-objective}.
Suppose we hold $\lambda_1$ fixed and $n_\ell$ grows in proportion to $h$.
Then each shrinkage factor $n_\ell/(n_\ell+\lambda_1)$ tends to one as $h$ grows.
Short histories thus receive more shrinkage, which makes boosting less aggressive there.
Consequently, $T_h$ and $d_h^{\rm boost}$ change with the memory.
If we scale the penalty as $\lambda_k=h\bar\lambda_k$, the relative penalty stays stable once the leaf proportions stabilize.
This scaling is an additional condition.
It does not follow from a fixed tree topology.

For several trees, $T_h$ need not be symmetric.
Hence the relevant variance dimension is $\operatorname{tr}(T_hT_h^{\top})$, not a ridge eigenvalue sum.
Suppose $d_h^{\rm boost}$ and $D_{h,\rm boost}^2$ stay stable over the candidate windows, and we control the other terms.
Then the memory-dependent part of the bound has the form $V_{\rm boost}/h+K_{\rm boost}h+W_{\rm boost}h^2$.
If $\mu_h^2$ is the only source of time dependence, then $K_{\rm boost}=2W_{\rm boost}$, and the positive continuous minimizer solves
\begin{equation}
2W_{\rm boost}h^3+K_{\rm boost}h^2
-V_{\rm boost}=0.
\label{eq:boost-memory-cubic}
\end{equation}
This result does not cover outcome-dependent greedy splits, $\ell_1$ leaf penalties, or an outcome-dependent initial prediction.
A fully retrained XGBoost model has all three features.
Analyzing them needs additional assumptions.
We treat greedy splits in Section~\ref{sec:theory-architecture}.

\subsubsection{Spectral phase boundaries for fixed feature maps}
\label{sec:theory-spectral-phase}

The bounds above compare the memory-dependent costs in aggregate.
We now compare models coordinate by coordinate in an orthogonal fixed-feature setting.
This refines Proposition~2 of Section~\ref{sec:theory-main-memory} for the surrogate.
Let $\psi_1,\ldots,\psi_p$ be features that we fix before we use the readout outcomes.
Let $V$ be a generic input drawn from the current input distribution $P_t$, and write $\operatorname{E}_t$ for expectation under $P_t$.
We assume that the features are orthogonal under $P_t$:
$\operatorname{E}_t[\psi_i\psi_k]=s_i\mathbf 1\{i=k\}$ with $s_i>0$.
We write the current conditional mean as $m_t=m_{\perp,t}+\sum_{i=1}^p\theta_{i,t}\psi_i$.
Here $\theta_{i,t}$ is the coefficient of mode $i$, and $m_{\perp,t}$ is orthogonal to every $\psi_i$.
The prediction norm is $\|f\|_t^2=\operatorname{E}_{V\sim P_t}[f(V)^2]$.
We assume that the training normal equations are diagonal with the same $s_i$.
Let $\dot\theta_{i,t}$ be the per-period drift of $\theta_{i,t}$.
We define the training score of mode $i$ and assume its conditional moments:
\begin{align}
U_{i,h}&=\frac1h\sum_{j=1}^h
 \psi_i(v_{t-j})y_{t-j},\notag\\
\operatorname{E}[U_{i,h}\mid\mathcal G_h]
 &=s_i(\theta_{i,t}-\mu_h\dot\theta_{i,t}),\notag\\
\operatorname{Var}(U_{i,h}\mid\mathcal G_h)
 &=\frac{\kappa\sigma_t^2s_i}{h},
\qquad \mu_h=\frac{h+1}{2}.
\label{eq:orthogonal-score}
\end{align}
These assumptions encode four conditions: locally linear coefficient drift, temporal balance of the design, an exact coordinate score variance, and zero conditional score bias given the whole design.
The last condition is strict score exogeneity.
A long-run variance bound does not imply it.
All four conditions are stronger than the inequalities that we used for the frozen-feature-map and boosting bounds.

For a subset $I$ of modes, we fit the ridge coefficients $\widehat\theta_i=U_{i,h}/(s_i+\lambda)$ with $\lambda\ge0$ for $i\in I$, and we set all other coefficients to zero.
Put $\omega_i=s_i/(s_i+\lambda)$, the ridge gain of mode $i$.
Put $b_{i,h}=(1-\omega_i)\theta_{i,t}+\mu_h\omega_i\dot\theta_{i,t}$, the bias of coordinate $i$.

\appresult{Proposition}{exact spectral complexity boundary}\label{res:spectral}
Assume \eqref{eq:orthogonal-score}.
The conditional squared error of the fitted forecast for the current conditional mean, excluding irreducible future noise, is
\begin{align}
\mathcal L_I(h)
&=\|m_{\perp,t}\|_t^2
+\sum_{i\notin I}s_i\theta_{i,t}^2\notag\\
&\quad+\sum_{i\in I}\left[
 s_ib_{i,h}^2
 +\frac{\kappa\sigma_t^2}{h}\omega_i^2\right].
\label{eq:spectral-exact-risk}
\end{align}
Consider nested mode sets $I_S\subset I_F$ that we fit with the same $\lambda$, and let $\mathcal N=I_F\setminus I_S$ be the added modes.
The richer set $I_F$ has strictly smaller error if and only if
\begin{align}
\sum_{i\in\mathcal N}s_i\theta_{i,t}^2
&>\sum_{i\in\mathcal N}
\left(s_ib_{i,h}^2
+\frac{\kappa\sigma_t^2}{h}\omega_i^2\right).
\label{eq:spectral-phase}
\end{align}
Now fit all included modes without penalty, so that $\omega_i=1$ for $i\in\mathcal N$.
Define $d_{\mathcal N}=|\mathcal N|$,
$C_{\mathcal N}=\sum_{i\in\mathcal N}s_i\theta_{i,t}^2$,
and $Q=\sum_{i\in\mathcal N}s_i\dot\theta_{i,t}^2$.
Then \eqref{eq:spectral-phase} reduces exactly to
\begin{equation}
\boxed{C_{\mathcal N}>
\frac{\kappa\sigma_t^2d_{\mathcal N}}{h}
             +\mu_h^2Q.}
\label{eq:spectral-unpenalized-phase}
\end{equation}
The left side $C_{\mathcal N}$ is the approximation error that the added modes remove.
The right side has two costs of fitting them.
The estimation variance $\kappa\sigma_t^2d_{\mathcal N}/h$ falls with $h$.
The drift bias $\mu_h^2Q$ grows with $h$.

\paragraph{Proof.}
For an included mode, \eqref{eq:orthogonal-score} gives
$\operatorname{E}[\widehat\theta_i\mid\mathcal G_h]
=\omega_i(\theta_{i,t}-\mu_h\dot\theta_{i,t})$ and
$s_i\operatorname{Var}(\widehat\theta_i\mid\mathcal G_h)
=\kappa\sigma_t^2\omega_i^2/h$.
Orthogonality makes the squared prediction error a sum of coordinate errors, plus $\|m_{\perp,t}\|_t^2$.
An omitted mode contributes $s_i\theta_{i,t}^2$.
This proves \eqref{eq:spectral-exact-risk}.
The common coordinates cancel in the nested comparison, which gives \eqref{eq:spectral-phase} and \eqref{eq:spectral-unpenalized-phase}.
\hfill\rule{0.5em}{0.5em}

\appresult{Corollary}{exact interior-memory equation}\label{res:spectral-memory}
Assume that $s_i$, $\lambda$, $\kappa\sigma_t^2$, and the coefficient path in Proposition~\ref{res:spectral} are the same across candidate histories.
Define
\begin{align}
V_I&=\kappa\sigma_t^2\sum_{i\in I}\omega_i^2,\notag\\
J_I&=\sum_{i\in I}s_i(1-\omega_i)\omega_i
 \theta_{i,t}\dot\theta_{i,t},\notag\\
Q_I&=\sum_{i\in I}s_i\omega_i^2\dot\theta_{i,t}^2.
\label{eq:spectral-memory-coefficients}
\end{align}
The part of \eqref{eq:spectral-exact-risk} that depends on $h$ equals $V_I/h+2\mu_hJ_I+\mu_h^2Q_I$.
Assume $V_I>0$ and $Q_I>0$.
The continuous extension of this part has a unique positive minimizer.
It solves
\begin{equation}
\boxed{\frac{V_I}{h^2}
=J_I+\frac{h+1}{2}Q_I.}
\label{eq:spectral-memory-equation}
\end{equation}
If we fit all modes in $I$ without penalty, then $J_I=0$ and $Q_I=\sum_{i\in I}s_i\dot\theta_{i,t}^2$.
The minimizer then solves the cubic below, and it obeys the asymptotic form when it is large:
\begin{equation}
\boxed{h^2(h+1)=\frac{2\kappa\sigma_t^2|I|}{Q_I}},
\qquad
h^*\sim\left(\frac{2\kappa\sigma_t^2|I|}
{Q_I}\right)^{1/3}.
\label{eq:spectral-memory-cubic}
\end{equation}

We now apply this result to the cost of the added modes.
Suppose we fit the added modes in \eqref{eq:spectral-unpenalized-phase} without penalty, and suppose $d_{\mathcal N},Q>0$.
Their complexity cost is $\kappa\sigma_t^2d_{\mathcal N}/h+\mu_h^2Q$.
Its minimizer solves \eqref{eq:spectral-memory-cubic} with $|I|$ replaced by $d_{\mathcal N}$ and $Q_I$ replaced by $Q$.
The complex class wins on an interval of positive histories if and only if $C_{\mathcal N}$ exceeds the minimum of this cost.
If $Q=0$ and $d_{\mathcal N},C_{\mathcal N}>0$, the complex class wins for all $h>\kappa\sigma_t^2d_{\mathcal N}/C_{\mathcal N}$.
For integer histories, we must check the grid directly.

\paragraph{Proof.}
We expand the square in \eqref{eq:spectral-exact-risk}.
Its $h$-dependent terms are $V_I/h+2\mu_hJ_I+\mu_h^2Q_I$, up to a constant.
Differentiation gives
$-V_I/h^2+J_I+(h+1)Q_I/2$.
Its derivative is
$2V_I/h^3+Q_I/2>0$, so it has exactly one zero, which is the minimum.
The unpenalized formulas follow by setting $\omega_i=1$.
\hfill\rule{0.5em}{0.5em}

The results above apply to boosted trees with fixed splits.
Take one prechosen squared-loss tree with unpenalized leaf means.
Refine it into a nested tree with more leaves.
An orthogonal contrast basis for the refinement makes $d_{\mathcal N}$ the number of added leaf contrasts.
The quantity $C_{\mathcal N}$ is the reduction in current approximation error, and $Q$ is the drift energy in these contrasts.
Under the score and design conditions above, \eqref{eq:spectral-unpenalized-phase} is thus a boundary for fixed-partition XGBoost-type models.

For several fixed boosting trees, we use the general noncommuting operator bound of Proposition~\ref{res:boost}.
A spectral specialization needs three further conditions.
First, $\mathsf W_k=I$, so that every $P_k$ is symmetric.
Second, the $P_k$ commute.
Third, their common orthonormal eigenvectors equal the normalized training-sample evaluations of the modes $\psi_i$.
Commutation alone does not identify these modes.
Let $p_{k,i}$ be the joint eigenvalue of $P_k$ for mode $i$.
The boosting smoother then has the modal gains
$\omega_i^{\rm boost}=1-\prod_{k=1}^{N_{\rm tree}}(1-\nu p_{k,i})$.
If the prediction norm and the score moments use this joint mode basis and satisfy \eqref{eq:orthogonal-score}, the same calculation of bias, drift, and variance applies to these modes.

The same results cover frozen neural encoders.
For a frozen RNN or LSTM, we can orthogonalize the final-hidden-state features and use them as the $\psi_i$.
For a frozen Transformer, we can orthogonalize the pooled-state or selected-token features and use them as the $\psi_i$.
Attention may stay input-dependent.
Each model family has its own $(s_i,\theta_{i,t},\dot\theta_{i,t})$.
Hence each family has its own boundary \eqref{eq:spectral-phase} and its own optimum \eqref{eq:spectral-memory-equation}.
The comparison covers nested sets of readout modes within one frozen family.
It does not order different, nonnested encoders.
We fix the input-context length of an encoder and vary only the training history $h$.
Finally, the cross term $J_I$ explains why a simple cube-root rule is only approximate.
Ridge shrinkage and partial boosting gains both give $0<\omega_i<1$, which can make $J_I\neq0$.

\subsubsection{Transfer bounds for boosted trees and neural encoders}
\label{sec:theory-architecture}

We now bound the gap between the surrogate and the trained model, as in Section~\ref{sec:theory-main-transfer}.
We prove Proposition~4 of Section~\ref{sec:theory-main-transfer} in two steps.
First, we bound the risk gap without model-specific conditions (Proposition~\ref{res:transfer}).
Second, we derive the phase boundary and the memory for trained models (Corollary~\ref{res:architecture}).
Between these steps, we give sufficient conditions for each model family.

\appresult{Proposition}{transfer under forecast perturbation}\label{res:transfer}
Let $k$ index a model.
Let $g_{k,h}$ be its fixed-feature surrogate forecast and $f_{k,h}$ its retrained forecast at memory $h$.
We evaluate both against the same current conditional mean $m_t$ in the evaluation norm $\|\cdot\|_t$ of Section~\ref{sec:theory-spectral-phase}.
Let $\mathcal G$ be common pre-fit information.
Define the surrogate risk, the retrained risk, and the squared perturbation by
\begin{align*}
\mathcal L_k^0(h)&=\operatorname{E}[\|g_{k,h}-m_t\|_t^2\mid\mathcal G],\\
\mathcal L_k(h)&=\operatorname{E}[\|f_{k,h}-m_t\|_t^2\mid\mathcal G],\\
\delta_{k,h}^2&=
\operatorname{E}[\|f_{k,h}-g_{k,h}\|_t^2\mid\mathcal G].
\end{align*}
Then
\begin{equation}
\left|\mathcal L_k(h)-\mathcal L_k^0(h)\right|
\le 2\delta_{k,h}\sqrt{\mathcal L_k^0(h)}+\delta_{k,h}^2
=:\epsilon_{k,h}.
\label{eq:transfer-risk}
\end{equation}
Consequently, if a surrogate ranking gap exceeds $\epsilon_{1,h}+\epsilon_{0,h}$, the retrained forecasts have the same ranking sign.
Next, consider the memory.
Let $\mathcal H\subseteq[1,\infty)$ be an admissible interval of histories.
Suppose $|\mathcal L_k(h)-F_0(h)|\le\epsilon$ for all $h\in\mathcal H$, where $F_0(h)=C_0+V/h+Kh+Wh^2$ with $V,W>0$.
Let $h_{\rm actual}$ minimize $\mathcal L_k$ over $\mathcal H$, and let $h_0^*$ minimize $F_0$ over $\mathcal H$.
If $h_0^*$ is interior, it is the unique positive root of $2Wh^3+Kh^2-V=0$.
Otherwise it is the corresponding endpoint of $\mathcal H$.
Then
\begin{equation}
|h_{\rm actual}-h_0^*|
\le\sqrt{2\epsilon/W}.
\label{eq:transfer-memory}
\end{equation}
In Section~\ref{sec:theory-main-transfer}, $V$, $K$, and $W$ are the surrogate coefficients $V_k$, $K_k$, and $W_k$.

\paragraph{Proof.}
We expand the difference of the squared norms and apply the Cauchy--Schwarz inequality.
This proves \eqref{eq:transfer-risk}.
For the memory claim, we apply the uniform error bound twice.
It shows that the surrogate risk $F_0$ at $h_{\rm actual}$ exceeds its minimum by at most $2\epsilon$.
The surrogate has second derivative $2V/h^3+2W\ge2W$.
Because $h_0^*$ minimizes $F_0$ over the interval $\mathcal H$, strong convexity gives $F_0(h)-F_0(h_0^*)\ge W(h-h_0^*)^2$ for every $h\in\mathcal H$.
Hence $W(h_{\rm actual}-h_0^*)^2\le2\epsilon$, which is \eqref{eq:transfer-memory}.
\hfill\rule{0.5em}{0.5em}

The transfer error $\epsilon_{k,h}$ enters additively, and it grows linearly in $\delta_{k,h}$ when $\delta_{k,h}$ is small, because $\delta_{k,h}^2$ is of second order.
This transfer result is conditional.
We do not show that $\delta_{k,h}$ is small for XGBoost with outcome-selected splits.
We also do not show it for recurrent or attention encoders whose parameters we train on the forecast loss.
Before we can transfer the fixed-feature boundary to these algorithms, we need an independent bound on the change in splits, the movement of the features, or the effect of the training algorithm.
We now state sufficient conditions that bound the perturbation $\delta_{k,h}$ of Proposition~\ref{res:transfer}.
Each condition concerns a specific version of a training procedure.
A model name alone does not imply any of them.
Prior work gives learning bounds for regularized boosting, recurrent networks, and Transformers under different complexity restrictions
\citep{cortes2019regularized,chen2020recurrent,
trauger2024transformers}.
We add explicit forecast-perturbation terms to the memory and phase comparisons of this appendix.

We start with greedy boosted trees, the first family in Section~\ref{sec:theory-main-transfer}.
We follow the split-and-leaf rules of \citet{chen2016xgboost}.
Consider squared-loss stagewise boosting with $N_{\rm tree}$ greedy, axis-aligned trees.
The model has a fixed learning rate, fixed row weights, $\ell_2$ leaf penalties, a zero initial prediction, and no $\ell_1$ penalty.
Candidate thresholds lie on a prespecified finite grid, for example a fixed number of histogram bins per feature.
Tree depth and $N_{\rm tree}$ stay bounded across candidate histories.
Hence the total number of candidate gains along a reference path has a uniform bound.

We define a reference tree sequence from population gains.
Let $\mathcal G_h^0$ contain the training and common evaluation designs, the candidate grid, the row weights, the penalty choices, and the conditional target law.
It excludes the realized readout errors.
Index the candidate gains at the reference decisions by $s$.
At each reference decision, we compute every normalized empirical gain $\widehat G_s$ from the \emph{random reference residual}.
This residual results from applying all earlier reference leaf updates to the training outcomes.
We define the population gain $G_s=\operatorname{E}[\widehat G_s\mid\mathcal G_h^0]$.
Hence later-stage population gains average over the noisy residuals that earlier reference updates generate.
We choose recursively the candidate with the largest population gain, including the choice to stop.
These choices define the reference topology, which is $\mathcal G_h^0$-measurable.
Let $\widetilde f_h$ be the forecast that fits leaf updates on this topology.
Let $f_h$ be the forecast that uses greedy, outcome-dependent splits on the same grid.
We normalize all gains by $h$.
We express any minimum-split-loss threshold on this same scale, or we set it to zero.

\appresult{Proposition}{finite-grid greedy XGBoost with stable splits}\label{res:xgb}
Conditional on $\mathcal G_h^0$, assume the following.
\begin{itemize}
\item[(i)] \emph{Margin.}
At each reference split, the best population gain exceeds every competing gain by at least $\Delta_{\rm split}>0$.
\item[(ii)] \emph{Count.}
The reference decisions involve at most $N_{\rm gain}$ candidate gains in total, including the stop choice.
\item[(iii)] \emph{Concentration.}
For each such gain and every $\vartheta>0$,
\begin{equation}
\Pr\!\left(
 |\widehat G_s-G_s|\ge\vartheta\mid\mathcal G_h^0
\right)
\le 2\exp(-c h\vartheta^2/b_G^2),
\label{eq:xgb-gain-tail}
\end{equation}
where $c>0$ and $b_G>0$ are constants that are common to all gains.
\end{itemize}
The event in (iii) covers every decision that the greedy algorithm reaches while it agrees with the reference path.
We clip both final forecasts to $[-B,B]$, and we assume that the current conditional mean on the fixed evaluation design lies in $[-B,B]$.
Let $L_{*,h}^{0,\mathrm{clip}}$ and
$L_{*,h}^{\mathrm{greedy,clip}}$ be the conditional mean-squared risks of the clipped reference forecast and the clipped greedy forecast in the common norm $\|\cdot\|_*$ of Section~\ref{sec:theory-fixed-representations}.
Then
\begin{align}
\left|L_{*,h}^{\mathrm{greedy,clip}}
-L_{*,h}^{0,\mathrm{clip}}\right|
&\le\epsilon_{\rm XGB}(h),\notag\\
\epsilon_{\rm XGB}(h)
&:=8B^2N_{\rm gain}
\exp\!\left(-\frac{ch\Delta_{\rm split}^2}{4b_G^2}\right).
\label{eq:xgb-clipped-transfer}
\end{align}
Clipping also contracts squared error to a target in $[-B,B]$, so
\begin{align}
L_{*,h}^{\mathrm{greedy,clip}}
&\le L_{*,h}^{0,\mathrm{clip}}
+\epsilon_{\rm XGB}(h)\notag\\
&\le U_{*,h}^{\rm fixed}(h)
+\epsilon_{\rm XGB}(h),
\label{eq:xgb-greedy-bound}
\end{align}
where $U_{*,h}^{\rm fixed}$ is the fixed-evaluation upper bound \eqref{eq:boost-eval-bound}.
The second inequality in \eqref{eq:xgb-greedy-bound} is one-sided, because clipping can only lower the reference risk.

\paragraph{Proof.}
By the margin condition, deviations below $\Delta_{\rm split}/2$ preserve the best split at every reference decision.
Induction over nodes and boosting stages then makes the greedy and reference topologies, leaf updates, and forecasts identical.
The union bound and \eqref{eq:xgb-gain-tail} give
\begin{equation*}
\Pr(f_h\ne\widetilde f_h\mid\mathcal G_h^0)
\le 2N_{\rm gain}\exp(-c h\Delta_{\rm split}^2/(4b_G^2)).
\end{equation*}
Clipped squared error against a target in $[-B,B]$ lies in $[0,4B^2]$.
Hence a disagreement changes the conditional risk by at most $4B^2$ times this probability, in either direction.
Clipping cannot increase the squared error to such a target.
The bound \eqref{eq:boost-eval-bound} controls the unclipped reference risk.
This proves both claims.
\hfill\rule{0.5em}{0.5em}

We now study the memory implied by this bound.
Suppose the stable part of $U_{*,h}^{\rm fixed}$ equals $C_0+V/h+Kh+Wh^2$ with $V,W>0$.
Suppose also that the split-margin constants $B,N_{\rm gain},\Delta_{\rm split},c,b_G$ do not change across candidate histories.
Then the upper bound in \eqref{eq:xgb-greedy-bound} equals
$C_0+V/h+Kh+Wh^2+C_{\rm sel} e^{-\alpha h}$, where
$C_{\rm sel}=8B^2N_{\rm gain}$ and
$\alpha=c\Delta_{\rm split}^2/(4b_G^2)$.
This function is strictly convex.
Its unique continuous minimizer solves
\begin{equation}
2Wh+K-\frac{V}{h^2}
-\alpha C_{\rm sel} e^{-\alpha h}=0
\label{eq:xgb-selection-memory}
\end{equation}
and exceeds the positive root of $2Wh^3+Kh^2-V=0$, because the derivative of the selection term is negative at that root.
Thus a positive split-selection penalty moves the minimizer of this \emph{upper-bound surrogate} toward longer memory.
We do not claim that the minimizer of the actual greedy risk moves by the same amount.
This is the boosted-tree statement of Section~\ref{sec:theory-main-transfer}, where $\alpha\propto\Delta_{\rm split}^2$.

The result has limits.
The gain-tail condition \eqref{eq:xgb-gain-tail} is an assumption.
It does not follow from the XGBoost objective.
Bounded gains, a minimum leaf occupancy, and a suitable dependence condition may supply it, but we do not prove this.
Exact-greedy trees that consider every observed threshold behave differently.
Their candidate count generally grows with $h$, and adjacent gains have vanishing margins.
For them, $\epsilon_{\rm XGB}(h)$ need not decay, and the constant-parameter memory surrogate fails without new $h$-dependent control.
To extend the local-design bound to next-period inputs, we need the fixed-evaluation operator \eqref{eq:boost-eval-operator} and corresponding design assumptions.

We next turn to recurrent networks, the second family in Section~\ref{sec:theory-main-transfer}.
Consider an input sequence $x_1,\ldots,x_L$ of fixed length $L$.
We fix the reference encoder using independent or earlier outcomes.
Its hidden states are $s_\ell^0=\sigma(W_0s_{\ell-1}^0+U_0x_\ell)$.
The trained encoder has hidden states $s_\ell=\sigma(Ws_{\ell-1}+Ux_\ell)$.
Both encoders start at zero.
Here $x_\ell$ is the input at step $\ell$, $W_0$ and $W$ are the recurrent weight matrices, and $U_0$ and $U$ are the input weight matrices.
The function $\sigma$ is the activation.
We write $\|\cdot\|_{\rm op}$ for the operator norm.
Suppose the activation $\sigma$ is $1$-Lipschitz and $\sigma(0)=0$.

\appresult{Proposition}{contractive trained RNN}\label{res:rnn}
Assume the following bounds.
They hold almost surely given the common pre-fit sigma-field $\mathcal G$, uniformly over evaluation inputs.
Their upper bounds are $\mathcal G$-measurable.
\begin{align*}
\|x_\ell\|&\le B_x,\\
\|W_0\|_{\rm op},\|W\|_{\rm op}&\le\chi_{\rm RNN}<1,\\
\|U_0\|_{\rm op},\|U\|_{\rm op}&\le b,\\
\|W-W_0\|_{\rm op}&\le e_W,\\
\|U-U_0\|_{\rm op}&\le e_U.
\end{align*}
These bounds control the input norm ($B_x$), the recurrent gain ($\chi_{\rm RNN}$), the input weights ($b$), and the movement of the two weight matrices ($e_W$ and $e_U$).
Let the reference forecast be
$g_h(v)=a_0^\top s_L^0(v)$, where $v$ is the input sequence and $a_0$ is the ridge readout fitted on the frozen features.
Let the trained forecast be
$f_h(v)=a^\top s_L(v)$ with readout $a$.
Assume almost surely given $\mathcal G$ that
$\|a-a_0\|\le e_a$ and $\|a\|\le R_{\rm out}$,
with $\mathcal G$-measurable upper bounds.
Set $\Gamma_{{\rm RNN},L}=(1-\chi_{\rm RNN}^L)/(1-\chi_{\rm RNN})$ and
$R_L=bB_x\Gamma_{{\rm RNN},L}$.
Then, uniformly in $v$,
\begin{align}
|f_h(v)-g_h(v)|
&\le \delta_{\rm RNN},\notag\\
\delta_{\rm RNN}
&:=e_aR_L\notag\\
&\quad+R_{\rm out}\Gamma_{{\rm RNN},L}
(e_WR_L+e_UB_x).
\label{eq:rnn-movement}
\end{align}
Let $U_{\rm RNN}^{0}(h)$ be the bound on the conditional risk of the reference forecast from Proposition~\ref{res:frozen}, including any approximation residual via \eqref{eq:frozen-full-risk}.
Then the conditional risk $\mathcal L_{\rm RNN}(h)$ of the trained forecast obeys
\begin{equation}
\mathcal L_{\rm RNN}(h)
\le U_{\rm RNN}^{0}(h)
+2\delta_{\rm RNN}\sqrt{U_{\rm RNN}^{0}(h)}
+\delta_{\rm RNN}^2.
\label{eq:rnn-risk-transfer}
\end{equation}

\paragraph{Proof.}
The reference state obeys
$\|s_\ell^0\|\le\chi_{\rm RNN}\|s_{\ell-1}^0\|+bB_x$, hence
$\|s_\ell^0\|\le bB_x\Gamma_{{\rm RNN},\ell}\le R_L$.
The state difference satisfies
\begin{equation*}
\|s_\ell-s_\ell^0\|
\le\chi_{\rm RNN}\|s_{\ell-1}-s_{\ell-1}^0\|
+e_WR_L+e_UB_x.
\end{equation*}
Iteration gives
$\|s_L-s_L^0\|\le\Gamma_{{\rm RNN},L}(e_WR_L+e_UB_x)$.
We add and subtract $a^\top s_L^0$ to obtain \eqref{eq:rnn-movement}.
The pointwise bound controls the conditional $L_2$ perturbation in Proposition~\ref{res:transfer}.
This gives \eqref{eq:rnn-risk-transfer}.
\hfill\rule{0.5em}{0.5em}

The contraction $\chi_{\rm RNN}<1$ keeps the perturbation bounded as $L$ grows, because $\Gamma_{{\rm RNN},L}\le(1-\chi_{\rm RNN})^{-1}$.
The result stays conditional.
It needs bounds on the movement of the trained weights and readout.
We do not claim that gradient descent supplies such bounds.
This is the RNN statement of Section~\ref{sec:theory-main-transfer}, where we write $\chi$ for $\chi_{\rm RNN}$.

We turn to LSTMs, the third family in Section~\ref{sec:theory-main-transfer}.
Consider two LSTM encoders on the same inputs $x_1,\ldots,x_L$.
We fix the reference encoder using independent or earlier outcomes.
The trained encoder has all its parameters trained on the forecast loss.
For either encoder, let $s_\ell\in\mathrm R^p$ be the hidden state and $c_\ell\in\mathrm R^p$ the cell state.
Write $z_\ell=(s_\ell,c_\ell)$ for the augmented state.
The updates are
\begin{equation*}
c_\ell=f_\ell\odot c_{\ell-1}
+i_\ell\odot \widetilde c_\ell,\qquad
s_\ell=o_\ell\odot\tanh(c_\ell),
\end{equation*}
where $\odot$ is the elementwise product, $f_\ell$, $i_\ell$, and $o_\ell$ are the forget, input, and output gates, and $\widetilde c_\ell$ is the candidate cell state.
The sigmoid gates lie in $[0,1]^p$, and $\widetilde c_\ell\in[-1,1]^p$.
Both encoders start at zero.
We assume that the forget gates satisfy $\|f_\ell\|_\infty\le\rho_f<1$.

\appresult{Proposition}{trained LSTM with controlled gates}\label{res:lstm}
Compare the two encoders at any common admissible state and input.
Assume that the forget, input, candidate, and output gate outputs of the two encoders differ by at most $e_f,e_i,e_{\rm cand},e_o$, respectively, in Euclidean norm.
Assume also that the one-step map of the trained encoder, which sends $z_{\ell-1}$ to $z_\ell$ for a given input, is $\varrho$-Lipschitz in its state argument, with finite $\varrho\ge0$.
These movement and Lipschitz bounds hold almost surely given $\mathcal G$, uniformly over evaluation inputs.
Their upper bounds are $\mathcal G$-measurable.
Let $g_h(v)=a_0^\top s_L^0(v)$ use the frozen ridge readout, and let $f_h(v)=a^\top s_L(v)$ use the trained readout.
Assume $\|a-a_0\|\le e_a$ and $\|a\|\le R_{\rm out}$
almost surely given $\mathcal G$, with $\mathcal G$-measurable upper bounds.
Put
\begin{align*}
C_f&=(1-\rho_f)^{-1},\\
K_{\rm gate}&=2(C_fe_f+e_i+e_{\rm cand})+e_o,\\
\Gamma_{{\rm LSTM},L}&=\sum_{k=0}^{L-1}\varrho^k.
\end{align*}
Then
\begin{align}
|f_h(v)-g_h(v)|
&\le\delta_{\rm LSTM},\notag\\
\delta_{\rm LSTM}
&:=e_a\sqrt p
+R_{\rm out}\Gamma_{{\rm LSTM},L}K_{\rm gate}.
\label{eq:lstm-movement}
\end{align}
Let $U_{\rm LSTM}^{0}(h)$ be the risk bound for the frozen readout from Proposition~\ref{res:frozen} and \eqref{eq:frozen-full-risk}.
Consequently, the risk of the trained forecast is at most
\begin{equation}
U_{\rm LSTM}^{0}(h)
+2\delta_{\rm LSTM}\sqrt{U_{\rm LSTM}^{0}(h)}
+\delta_{\rm LSTM}^2.
\label{eq:lstm-risk-transfer}
\end{equation}

\paragraph{Proof.}
The forget-gate condition and the bounded input gate and candidate give $\|c_\ell\|_\infty\le C_f$.
In addition, $\|s_\ell\|\le\sqrt p$.
At a common state, perturbing the gates changes the cell update by at most
$E_c=C_fe_f+e_i+e_{\rm cand}$.
Since $\tanh$ is $1$-Lipschitz, the hidden update changes by at most $e_o+E_c$.
The Euclidean change in the augmented state is thus at most $2E_c+e_o=K_{\rm gate}$.
We insert and subtract one full state update at the same state and obtain
\begin{equation*}
\|z_\ell-z_\ell^0\|
\le \varrho\|z_{\ell-1}-z_{\ell-1}^0\|+K_{\rm gate}.
\end{equation*}
We iterate this inequality, and then we add and subtract
$a^\top s_L^0$ to get \eqref{eq:lstm-movement}.
We then apply Proposition~\ref{res:transfer} to obtain \eqref{eq:lstm-risk-transfer}.
\hfill\rule{0.5em}{0.5em}

The forget-gate bound controls the size of the cell state.
It does not by itself imply a contraction of the full state, which the special case $\varrho<1$ requires.
We must verify the full-state Lipschitz constant from all recurrent and gate weights.
Alternatively, we can replace it by a Jacobian product of finite length $L$.
If $\varrho\ge1$, the factor $\Gamma_{{\rm LSTM},L}$ and hence the bound can grow rapidly with the context length.
This is the LSTM statement of Section~\ref{sec:theory-main-transfer}, where $\varrho$ is the Lipschitz constant of the state update.

We finish with attention, the fourth family in Section~\ref{sec:theory-main-transfer}.
Consider a one-block Transformer readout with $N_{\rm head}$ attention heads and a fixed input-context length $L$.
We fix its reference attention map using independent or earlier outcomes.
For head $r$ and token $\ell$, let $\xi_{r\ell}^0$ and $v_{r\ell}^0$ be the reference attention logit and value vector.
Let $\xi_{r\ell}$ and $v_{r\ell}$ be their trained counterparts.
The value vector $v_{r\ell}$ carries two subscripts, and the input $v$ carries none.
Define the head outputs
\begin{equation*}
z_r^0=\sum_{\ell=1}^{L}
 \operatorname{softmax}(\xi_r^0)_\ell v_{r\ell}^0,
\quad
z_r=\sum_{\ell=1}^{L}
 \operatorname{softmax}(\xi_r)_\ell v_{r\ell}.
\end{equation*}

\appresult{Proposition}{trained multihead attention block}\label{res:transformer}
Assume the following bounds.
They hold almost surely given $\mathcal G$, uniformly over heads, tokens, and evaluation inputs.
Their upper bounds are $\mathcal G$-measurable.
\begin{equation*}
\|v_{r\ell}^0\|\le B_v,\quad
\|v_{r\ell}-v_{r\ell}^0\|\le e_v,\quad
|\xi_{r\ell}-\xi_{r\ell}^0|\le e_\xi.
\end{equation*}
Let $\varphi^0=O_0[z_1^0;\ldots;z_{N_{\rm head}}^0]$ and
$\varphi=O[z_1;\ldots;z_{N_{\rm head}}]$ be the feature vectors.
Here $O_0$ and $O$ are the output projection matrices of the reference and trained blocks, and $[\cdot\,;\cdot]$ stacks the head outputs.
Assume $\|O_0\|_{\rm op},\|O\|_{\rm op}\le N_O$ and
$\|O-O_0\|_{\rm op}\le e_O$ almost surely given $\mathcal G$, with $\mathcal G$-measurable upper bounds.
The frozen forecast is $g_h(v)=a_0^\top\varphi^0(v)$ with
ridge readout $a_0$.
The trained forecast is
$f_h(v)=a^\top\varphi(v)$.
Assume $\|a-a_0\|\le e_a$ and $\|a\|\le R_{\rm out}$
almost surely given $\mathcal G$, with
$\mathcal G$-measurable upper bounds.
Then
\begin{align}
|f_h(v)-g_h(v)|
&\le\delta_{\rm Tr},\notag\\
\delta_{\rm Tr}
&:=e_aN_O\sqrt{N_{\rm head}}\,B_v\notag\\
&\quad+R_{\rm out}\sqrt{N_{\rm head}}\,e_OB_v\notag\\
&\quad+R_{\rm out}\sqrt{N_{\rm head}}\,N_O
(e_v+B_ve_\xi).
\label{eq:transformer-movement}
\end{align}
Let $U_{\rm Tr}^{0}(h)$ be the risk bound for the frozen attention map from Proposition~\ref{res:frozen} and \eqref{eq:frozen-full-risk}.
Then the risk of the trained one-block model is at most
\begin{equation}
U_{\rm Tr}^{0}(h)
+2\delta_{\rm Tr}\sqrt{U_{\rm Tr}^{0}(h)}
+\delta_{\rm Tr}^2.
\label{eq:transformer-risk-transfer}
\end{equation}
The perturbation bound has no explicit factor of $L$.
It does need a uniform bound on the logit and value perturbations over all $L$ tokens.

\paragraph{Proof.}
Let $\pi$ be a probability vector on the $L$ tokens, and let $D_\pi^{\rm sm}=\operatorname{diag}(\pi)-\pi\pi^\top$ be the softmax derivative.
For any vector $w\in\mathrm R^L$, we have
$\|D_\pi^{\rm sm}w\|_1=\operatorname{E}_\pi|w-\operatorname{E}_\pi w|\le\|w\|_\infty$.
This follows from the Cauchy--Schwarz inequality and
$\operatorname{Var}_\pi(w)\le\|w\|_\infty^2$.
We integrate along the line segment between the two logit vectors and obtain
\begin{equation*}
\|\operatorname{softmax}(\xi)
-\operatorname{softmax}(\xi^0)\|_1
\le\|\xi-\xi^0\|_\infty.
\end{equation*}
Hence $\|z_r^0\|\le B_v$ and
$\|z_r-z_r^0\|\le e_v+B_ve_\xi$.
Stacking the heads multiplies these Euclidean bounds by at most $\sqrt{N_{\rm head}}$.
We perturb $O$ first and the readout second, which proves \eqref{eq:transformer-movement}.
We then apply Proposition~\ref{res:transfer} to obtain \eqref{eq:transformer-risk-transfer}.
\hfill\rule{0.5em}{0.5em}

Consider dot-product logits $\xi_{r\ell}=q_r^\top k_{r\ell}/\sqrt d$ with query $q_r$ and keys $k_{r\ell}\in\mathrm R^d$.
Suppose the query and key are linear images of tokens with norm at most $B_x$.
Suppose also that the query and key matrices of both the trained and reference blocks have operator norm at most $N_{QK}$.
If the query matrices differ by at most $e_Q$ and the key matrices by at most $e_K$ in operator norm, then
$e_\xi\le B_x^2N_{QK}(e_Q+e_K)/\sqrt d$.
If the value matrices differ by at most $e_V$ in operator norm, then $e_v\le B_xe_V$.
For additional Transformer blocks, feedforward layers, and residual paths, we can compose their bounded Lipschitz and perturbation constants.
The bound \eqref{eq:transformer-movement} itself holds for one block only.
This is the Transformer statement of Section~\ref{sec:theory-main-transfer}.

We can combine each of Propositions~\ref{res:xgb}--\ref{res:transformer} with Proposition~\ref{res:transfer} in two ways.
First, suppose a fixed-surrogate complexity margin is strict and exceeds the two relevant transfer errors.
Then the model ranking does not change.
Second, suppose the transfer error is uniform over the candidate histories and the fixed-surrogate risk has the strongly convex memory form above.
Then \eqref{eq:transfer-memory} localizes the retrained optimum.
Both implications are conditional on the model-specific constants above.
We do not claim that an arbitrary trained model satisfies them.

\appresult{Corollary}{model-conditional phase boundary and memory}\label{res:architecture}
Fix one common evaluation target, one design or distribution, one prediction norm, and one pre-fit sigma-field $\mathcal G$ for all models and histories.
If XGBoost enters the comparison, we take $\|\cdot\|_*$ on the fixed evaluation design as the prediction norm of every model.
Let $\mathcal L_k(h)$ be the risk of the fully trained model $k$, and let $\mathcal L_k^0(h)$ be the risk of its fixed surrogate.
Define the transfer error $\epsilon_k(h)$ as follows.
For a finite-grid greedy boosted-tree model that satisfies Proposition~\ref{res:xgb}, set
\begin{equation*}
\epsilon_{\rm XGB}(h)
=8B^2N_{\rm gain}\exp(
-c h\Delta_{\rm split}^2/(4b_G^2)).
\end{equation*}
For a trained RNN, LSTM, or one-block Transformer, set, respectively,
\begin{equation*}
\epsilon_k(h)=
2\delta_k\sqrt{U_k^0(h)}+\delta_k^2,
\end{equation*}
with $\delta_k$ given by
\eqref{eq:rnn-movement}, \eqref{eq:lstm-movement}, or
\eqref{eq:transformer-movement}, and with $U_k^0$ the surrogate risk bound of the corresponding proposition.
We assume that the gain-tail and movement bounds that define $\epsilon_k$ also hold given this common $\mathcal G$.
For XGBoost, we assume that the fixed-evaluation moment and gain-tail conditions hold with
$\mathcal G_h=\mathcal G_h^0=\mathcal G$ for every candidate $h$.

We also assume that the fixed-surrogate risks of classes $\mathcal M_0$ and $\mathcal M_1$ have expansions of the form \eqref{eq:class-risk}, with $e_k$ the surrogate remainder bounded by $r_k(h)$.
For XGBoost, the fixed-surrogate risk is the \emph{clipped} reference risk $L_{*,h}^{0,\mathrm{clip}}$ of \eqref{eq:xgb-clipped-transfer}.
The unclipped risk \eqref{eq:boost-eval-risk} gives only a one-sided upper bound after clipping.
Hence we cannot derive the two-sided expansion of $L_{*,h}^{0,\mathrm{clip}}$ that we need from Proposition~\ref{res:boost}.
For XGBoost, this expansion is an additional assumption.

Then the following hold.
\begin{itemize}
\item[(a)] The fully trained complex model $\mathcal M_1$ has smaller risk than the fully trained simple model $\mathcal M_0$ whenever
\begin{align}
\Delta_0&>\frac{\Delta V}{h}+\Delta Kh+\Delta Wh^2\notag\\
&\quad+r_1(h)+r_0(h)
+\epsilon_1(h)+\epsilon_0(h).
\label{eq:architecture-phase}
\end{align}
\item[(b)] The reverse ranking holds when $\Delta_0$ is below the same leading boundary minus all four remainder terms, that is, when
$\Delta_0<\Delta V/h+\Delta Kh+\Delta Wh^2-r_1(h)-r_0(h)-\epsilon_1(h)-\epsilon_0(h)$.
\item[(c)] Fix one model $k$.
Suppose $r_k(h)+\epsilon_k(h)\le\bar\epsilon_k$ uniformly over a continuous admissible history interval, that $V_k,W_k>0$, and that both minimizers below lie in this interval.
Let $h_k^{\rm trained}$ minimize $\mathcal L_k(h)$, and let $h_k^{0*}$ minimize the surrogate expansion $\mathcal A_k+B_k+V_k/h+K_kh+W_kh^2$.
Then
\begin{align}
|h_k^{\rm trained}-h_k^{0*}|
&\le\sqrt{2\bar\epsilon_k/W_k},\notag\\
2W_k(h_k^{0*})^3
+K_k(h_k^{0*})^2-V_k&=0
\label{eq:architecture-memory}
\end{align}
if the surrogate minimum is interior.
At a boundary, $h_k^{0*}$ is the constrained minimum instead.
\end{itemize}
Because we keep the known linear term $K_kh$ explicit, $\bar\epsilon_k$ need not absorb it.
The localization bound is informative only if the remaining uniform error is small on the chosen interval.
Items (a) and (b) are the phase-boundary statement of Proposition~4 in Section~\ref{sec:theory-main-transfer}, and item (c) is its memory statement.

\paragraph{Proof.}
We combine the transfer bound \eqref{eq:xgb-clipped-transfer} for XGBoost with Propositions~\ref{res:rnn}--\ref{res:transformer} for the neural models.
This gives
$|\mathcal L_k(h)-\mathcal L_k^0(h)|\le\epsilon_k(h)$
in the common evaluation norm.
For the neural cases, $\epsilon_k$ increases with $\mathcal L_k^0$, so we can replace $\mathcal L_k^0$ by its bound $U_k^0$.
We then subtract the two fixed-surrogate expansions.
Bounding their remainders and the transfer errors gives \eqref{eq:architecture-phase}.
For the reverse ranking in (b), we repeat the steps with the signs reversed.
For (c), the trained risk differs from
$\mathcal A_k+B_k+V_k/h+K_kh+W_kh^2$ by at most $\bar\epsilon_k$, uniformly on the interval.
The strong-convexity argument in the proof of Proposition~\ref{res:transfer} then gives \eqref{eq:architecture-memory}.
Together with Proposition~\ref{res:transfer}, this proves Proposition~4 of Section~\ref{sec:theory-main-transfer}.
\hfill\rule{0.5em}{0.5em}

%% file: appendix_simulations.tex
\section{Controlled Simulation Results}
\label{app:controlled-simulations}
This appendix gives the design and results for Section~\ref{sec:experiments-simulation}; Figure~\ref{fig:sim-main} presents the main-text gradual-drift comparison.
\subsection{Simulation method}
\label{app:sim-method}
\paragraph{Covariates and signal.} Each observation has $p=8$ covariates $x\in\mathbb{R}^p$ drawn from $N(0,\Sigma)$. The covariance is $\Sigma=\mathrm{diag}(\kappa_1,\dots,\kappa_p)$ with $\kappa_k=q^{k-1}$. The decay parameter $q\in\{1,0.45\}$ sets the spectrum. We call $q=1$ the flat spectrum and $q=0.45$ the decaying spectrum. The conditional mean at drift state $\beta$ is
\begin{equation*}
m_\beta(x)=\sum_{k=1}^{p}\beta_k x_k+\gamma\,\frac{x_1x_2}{\sqrt{\kappa_1\kappa_2}} .
\end{equation*}
The nonlinear strength $\gamma\in\{0,0.4,0.8\}$ scales a standardized product feature with unit variance. The baseline coefficient vector $\beta^{0}$ has a single nonzero entry, $\beta^{0}_2=0.7/\sqrt{\kappa_2}$, so the linear signal has variance $0.7^2$ for every $q$. The response is $y=m_\beta(x)+\varepsilon$ with $\varepsilon\sim N(0,s^2)$. The signal-to-noise ratio is $\mathrm{SNR}=(0.7^2+\gamma^2)/s^2$, which fixes the noise standard deviation $s=\sqrt{(0.7^2+\gamma^2)/\mathrm{SNR}}$ for $\mathrm{SNR}\in\{0.25,2\}$.

\paragraph{Drift.} A training observation of age $a\in\{1,\dots,256\}$ was recorded $a$ observations before the prediction date. Its coefficient vector equals $\beta^{0}$ except in one drifting coordinate $j$ (coordinate 1 in the factorial grid). Under gradual drift, $\beta_j=\beta^0_j-\rho a$ with rate $\rho=0.008$. Under an abrupt break, $\beta_j=\beta^0_j-0.65\cdot\mathbf 1\{a>48\}$. Under zero drift, $\beta=\beta^0$. The test targets use $\beta^{0}$ (age zero). Thus drift measures how far older observations lie from the prediction-date relation. The nonlinear term does not drift.

\paragraph{Models.} The linear model uses the $p=8$ covariates. The quadratic model adds the 36 centered second-order features $x_ix_j-\kappa_i\mathbf 1\{i=j\}$ for $i\le j$, which gives 44 features. Every feature has mean zero. The quadratic span contains the true nonlinear term, so the quadratic model is correctly specified up to drift. Both models are fitted without an intercept by ridge regression. For features $\phi$, the estimator minimizes $h^{-1}\sum_{t}(y_t-w^\top\phi_t)^2+\lambda\|w\|^2$ over the $h$ most recent observations, with penalty $\lambda\in\{0.03,0.5\}$ on the raw feature scale. The memory $h\in\{16,32,64,128,256\}$ is the number of training observations used. The histories are nested, so every memory uses the last $h$ rows of one 256-row history.

\paragraph{Worlds, replicates, and cells.} A \emph{world} fixes 16 test covariate vectors drawn from $N(0,\Sigma)$ and their targets. Within a world, we draw $R=3$ \emph{replicates}. Each replicate is an independent 256-row training history. All models, penalties, and memories are fitted to the same replicates and scored on the same targets. Hence replicate variation measures the effect of training sampling at fixed dates. Each factorial condition uses $W=24$ independent worlds. A \emph{cell} is one combination of design, model, penalty, and memory. It pools $N=16W$ test targets, collected in the vector $y\in\mathbb{R}^N$. Let $f_r\in\mathbb{R}^N$ be the forecasts of replicate $r$ and $\bar f=R^{-1}\sum_r f_r$ their average. Let $\mu\in\mathbb{R}^N$ be the known conditional mean at the test points.

\paragraph{Score and exact decomposition.} The score is the out-of-sample $R^2$ against the zero forecast,
\begin{equation*}
R^2=1-\frac{R^{-1}\sum_{r}\|f_r-y\|^2}{\|y\|^2}.
\end{equation*}
Define the aggressiveness $g=\|\bar f\|/\|y\|$, the alignment $a=y^\top\bar f/(\|y\|\|\bar f\|)$ (the cosine between targets and the average forecast), and the instability $u=R^{-1}\sum_r\|f_r-\bar f\|^2/\|y\|^2$. Expanding the square gives the exact identity
\begin{equation*}
R^2=2ag-g^2-u .
\end{equation*}
We call $2ag$ the alignment reward, $g^2$ the aggressiveness cost, and $u$ the instability cost. We compute all three from unrounded forecasts. The largest reconstruction error over all 844 cells is $8.88\times10^{-16}$. We also report the excess risk $R^{-1}N^{-1}\sum_r\|f_r-\mu\|^2$, which removes observation noise from the comparison.

\paragraph{Spectral proxy.} For each cell we compute a population approximation to the excess risk. Let $v_k$ be the variance of feature $k$. It equals $\kappa_k$ for a linear feature and $\kappa_i\kappa_j(1+\mathbf 1\{i=j\})$ for a quadratic one. Define the ridge filter $\tau_k=v_k/(v_k+\lambda)$ and the variance-effective dimension $d_2(\lambda)=\sum_k\tau_k^2$. The shrinkage bias in feature $k$ is $-\beta^{0}_k(1-\tau_k)$. The drift bias in the drifting coordinate is $-\tfrac{h+1}{2}\rho\,\tau_j$ under gradual drift and $-\tfrac{\max(h-48,0)}{h}\,0.65\,\tau_j$ under the break. Let $B$ be the sum over features of $v_k$ times the squared total bias. Let $A=\gamma^2$ for the linear model, which omits the nonlinear term, and $A=0$ for the quadratic model. The proxy excess risk is
\begin{equation*}
\widehat{\mathrm{ER}}=A+B+(s^2+A)\,d_2(\lambda)/h .
\end{equation*}
The proxy $R^2$ is $1-(s^2+\widehat{\mathrm{ER}})/(0.7^2+\gamma^2+s^2)$. The proxy gain of the quadratic model is $\widehat{\mathrm{ER}}_{\text{linear}}-\widehat{\mathrm{ER}}_{\text{quadratic}}$. This approximation is exact in form for correctly specified gradual linear drift. For the break and for the omitted nonlinearity, it is a diagnostic approximation, because the omitted term enters only as extra residual variance. We assign no proxy to the null design.

\paragraph{Focused designs.} Four further designs share the ridge fits and score definitions above. All use zero nonlinear strength unless noted.
\emph{(i) Ridge scaling.} We use the flat spectrum, $\mathrm{SNR}=1$, gradual drift with $\rho\in\{0.002,0.004,0.008,0.016\}$, linear ridge with $\lambda=0.03$, and 11 memories in $\{16,32,48,64,80,96,112,128,160,192,256\}$. The design has 512 worlds. Training covariates are deterministic and balanced. A block of $2p=16$ rows contains $\pm\sqrt{p\kappa_k}\,e_k$ for each coordinate $k$, placed at positions $k$ and $2p+1-k$. The block is tiled 16 times, and every memory is a whole number of blocks. Hence the training second-moment matrix equals $\Sigma$ exactly. Moreover, the age-weighted second-moment matrix equals the mean age times $\Sigma$, because each sign pair is symmetric in age. Only the noise is redrawn across replicates. The test covariates stay random. We take the observed optimal memory $h^\star$ as the grid memory with the smallest excess risk. We compare it with the proxy minimizer and regress $\log h^\star$ on $\log\rho$. Balancing the squared drift bias, of order $(h\rho)^2$, against the variance, of order $1/h$, predicts a slope of $-2/3$.
\emph{(ii) Directional drift.} We use gradual drift with $\rho=0.012$, $\mathrm{SNR}=2$, $q=0.45$, and the drifting coordinate $j=1$ (variance $\kappa_1=1$) or $j=8$ (variance $\kappa_8=0.45^{7}\approx0.004$). The raw drift is equal, but the filtered drift $\kappa_j(\tau_j\rho)^2$ differs.
\emph{(iii) Stationary weak signal.} We use zero drift, $q=0.45$, $\beta^0_2=0.15/\sqrt{\kappa_2}$, $\gamma=0$, noise standard deviation $s=1$, and 192 worlds.
\emph{(iv) Permutation null.} We use zero drift, $q=0.45$, $\mathrm{SNR}=2$, $\gamma=0.4$, and 96 worlds. In each training history and each test set, we generate the responses from an independent covariate draw and then permute them across rows. Hence responses are independent of the covariates. We set the conditional mean to zero. No forecast should beat zero systematically.
The directional, weak-signal, and null designs each use both models, both penalties, and the five main-grid memories (the scaling design uses linear ridge only). In total, the grid has $36\times20=720$ factorial cells and $40+44+20+20=124$ focused cells, which gives 844.

\paragraph{Random numbers and pairing.} We use NumPy's \texttt{default\_rng} with base seed $s_0=20261003$. The factorial seed is $s_0+1000\,i_{\mathrm{SNR}}+i_q$, where $i_{\mathrm{SNR}}$ and $i_q$ are the zero-based positions of the SNR and decay in their lists. Hence all nine drift--$\gamma$ conditions with the same SNR and spectrum reuse the same standard normal draws, and only the coefficients and noise scale change. These common random numbers pair the contrasts across drift and nonlinear strength. The directional designs use $s_0+2000$ and the scaling designs use $s_0+3000$. The weak-signal and null designs use $s_0+4000+k$, where $k=38$ and $k=39$ is the position of the design in the run order.

\paragraph{Paired contrasts and intervals.} For each of the 360 factorial conditions (drift, SNR, $\gamma$, $q$, $\lambda$, $h$), we compare the two models fitted to the same worlds. Let $\Delta R^2=R^2_{\text{quadratic}}-R^2_{\text{linear}}$. For world $w$, let $\ell_w$ be the mean squared forecast error averaged over replicates and the 16 targets. The paired difference is $d_w=\ell^{\text{linear}}_w-\ell^{\text{quadratic}}_w$. The standard error of $\Delta R^2$ is $\mathrm{sd}(d_w)/\sqrt{W}$ divided by the mean per-world target energy. The interval is $\Delta R^2\pm1.96$ standard errors. The 24 worlds are independent, so this normal interval treats worlds as clusters. We call a contrast \emph{clear} if its interval excludes zero. We call the proxy sign \emph{matching} if $\Delta R^2$ and the proxy gain have the same sign. The intervals are descriptive. They do not adjust for multiplicity or for the search over specifications.

\subsection{Results}
\label{app:sim-results}
\begin{figure}[p]
\centering
\includegraphics[width=0.7\textwidth,height=0.58\textheight,keepaspectratio]{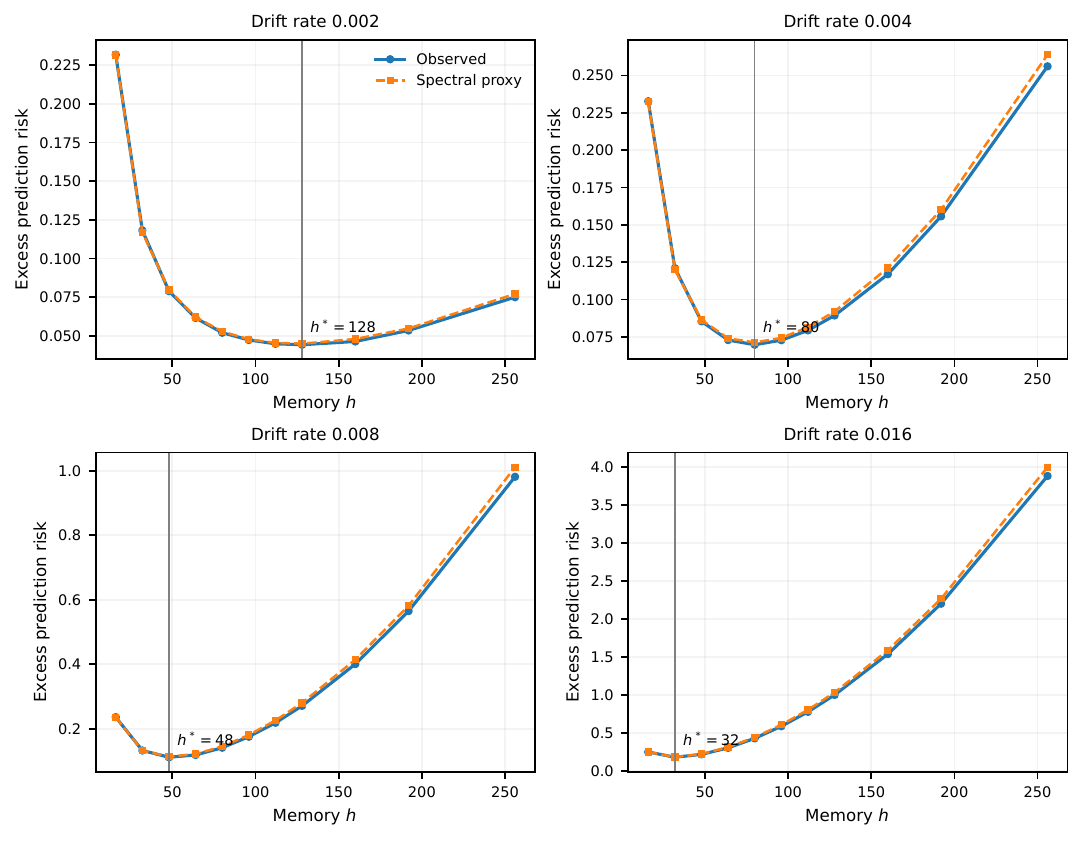}
\caption{Ridge theorem check under a balanced linear design. Each panel varies memory at a fixed coefficient-drift rate. Solid curves are observed excess prediction risk and dashed curves are the spectral approximation. The vertical rule marks the observed optimum; all four agree with the proxy grid optimum, and the log-optimal-memory versus log-drift slope is -0.67.}
\label{fig:sim-scaling}
\end{figure}

\begin{figure}[p]
\centering
\includegraphics[width=0.7\textwidth,height=0.58\textheight,keepaspectratio]{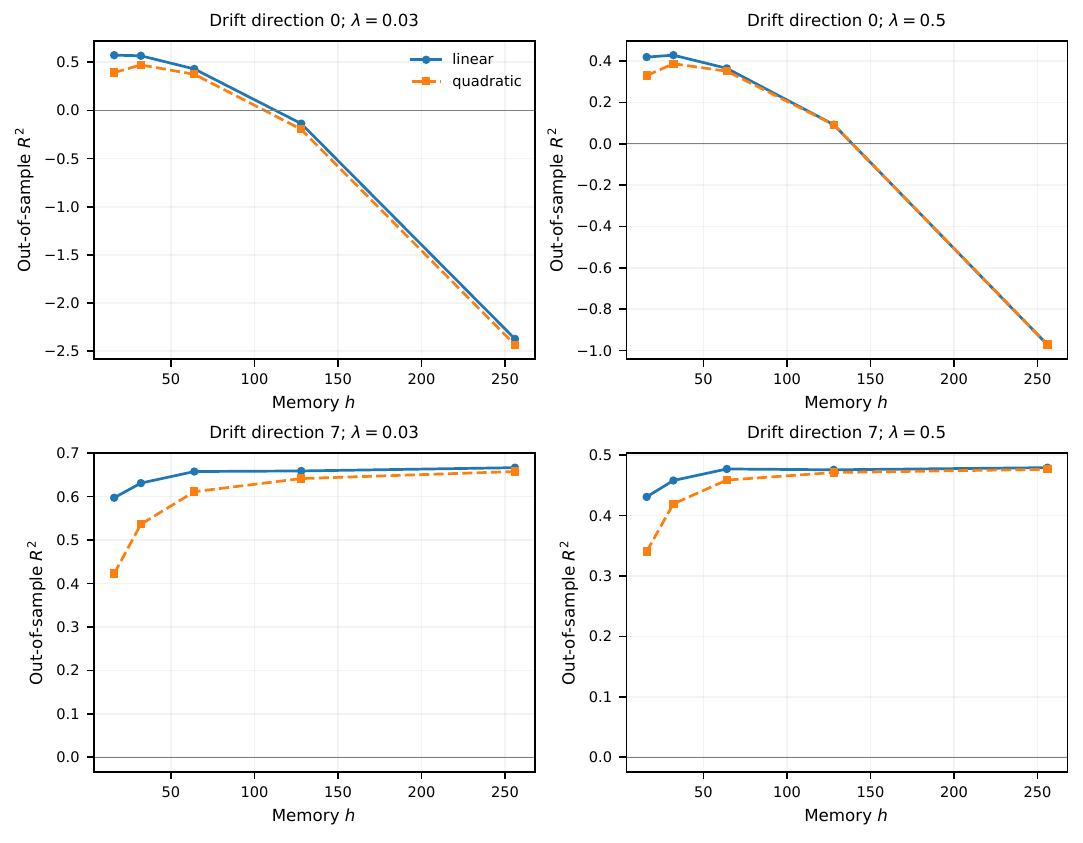}
\caption{Direction-aware drift. Rows put the same raw drift rate in a high-variance coordinate (top) or heavily filtered low-information coordinate (bottom); columns vary ridge penalty. Linear and quadratic scores are shown at every candidate memory. With light regularization, the best linear memory is 32 in the high-variance direction and 256 in the low-information direction.}
\label{fig:sim-directional}
\end{figure}

\begin{figure}[p]
\centering
\includegraphics[width=0.7\textwidth,height=0.58\textheight,keepaspectratio]{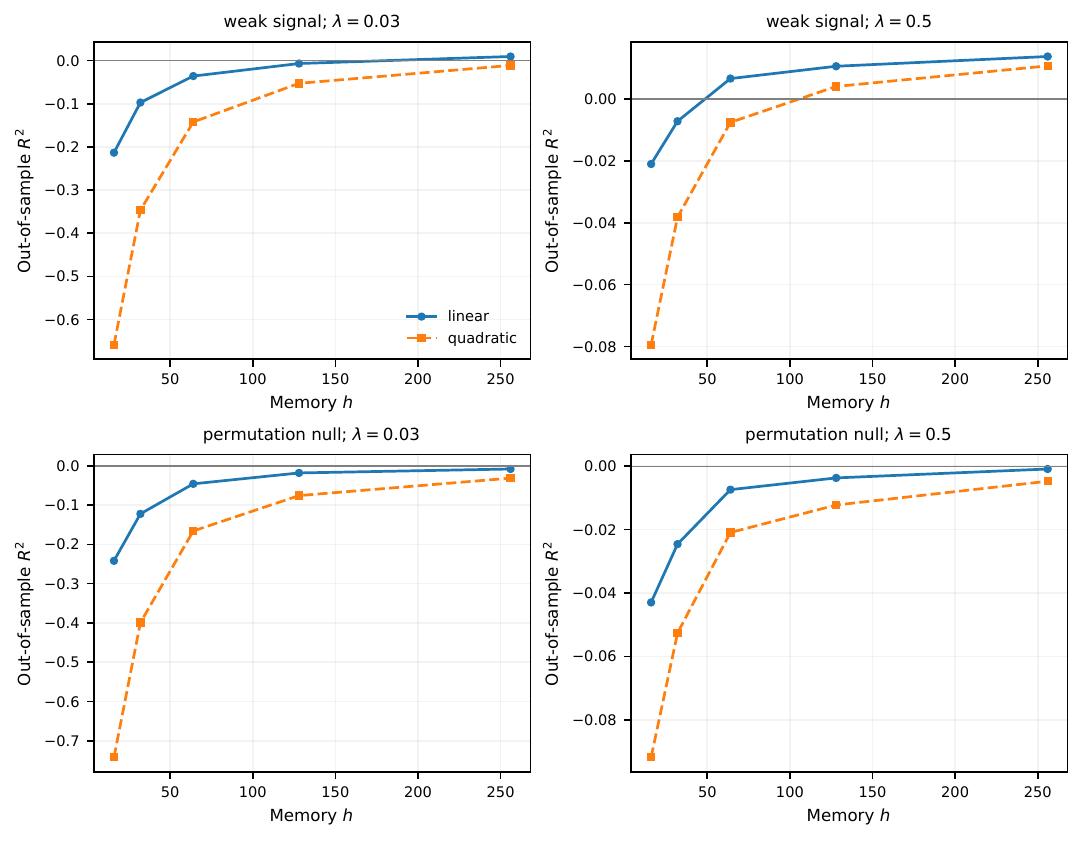}
\caption{Stationary weak-signal and permutation-null controls. Columns compare light and heavy ridge penalties; each panel shows both model classes over every memory. The weak-signal optimum is long-memory, heavily shrunk linear ridge. None of the 20 null specifications exceeds the zero forecast in this run.}
\label{fig:sim-controls}
\end{figure}

\begin{figure}[p]
\centering
\includegraphics[width=0.85\textwidth,height=0.70\textheight,keepaspectratio]{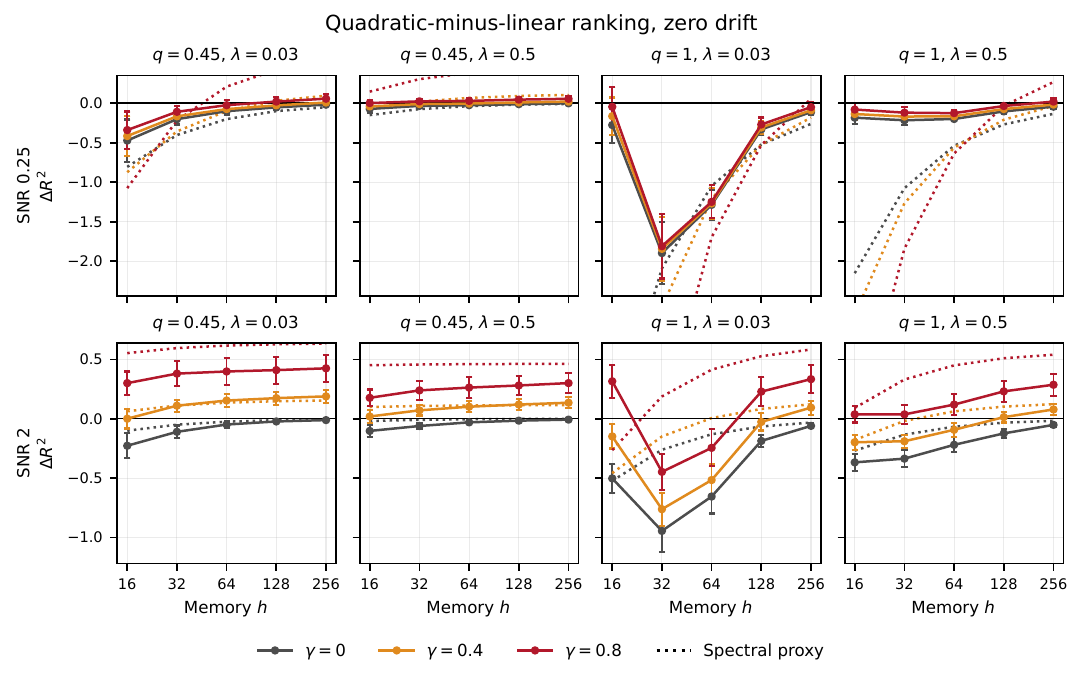}
\caption{Quadratic-minus-linear $R^2$ gap versus memory under zero drift. Rows fix SNR and columns fix covariance decay $q$ and penalty $\lambda$. Solid curves with bars are the observed gap and its paired 95\% interval; colors give nonlinear strength $\gamma$. Dotted curves are the leading spectral proxy. Positive values favor the quadratic model. The vertical range of each row covers the observed intervals, so a proxy curve that leaves the panel has missed the observed gap by a large margin. The intervals are descriptive and unadjusted for multiplicity.}
\label{fig:sim-phase-zero}
\end{figure}

\begin{figure}[p]
\centering
\includegraphics[width=0.85\textwidth,height=0.70\textheight,keepaspectratio]{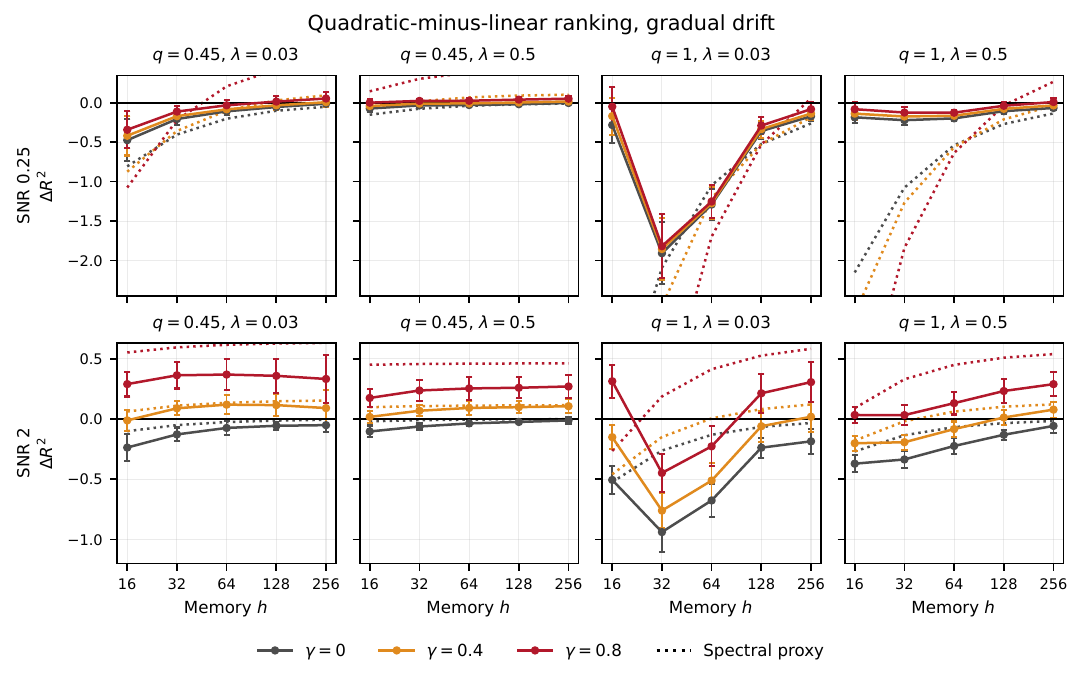}
\caption{Quadratic-minus-linear $R^2$ gap versus memory under gradual drift. Rows fix SNR and columns fix covariance decay $q$ and penalty $\lambda$. Solid curves with bars are the observed gap and its paired 95\% interval; colors give nonlinear strength $\gamma$. Dotted curves are the leading spectral proxy. Positive values favor the quadratic model. The vertical range of each row covers the observed intervals, so a proxy curve that leaves the panel has missed the observed gap by a large margin. The intervals are descriptive and unadjusted for multiplicity.}
\label{fig:sim-phase-gradual}
\end{figure}

\begin{figure}[p]
\centering
\includegraphics[width=0.85\textwidth,height=0.70\textheight,keepaspectratio]{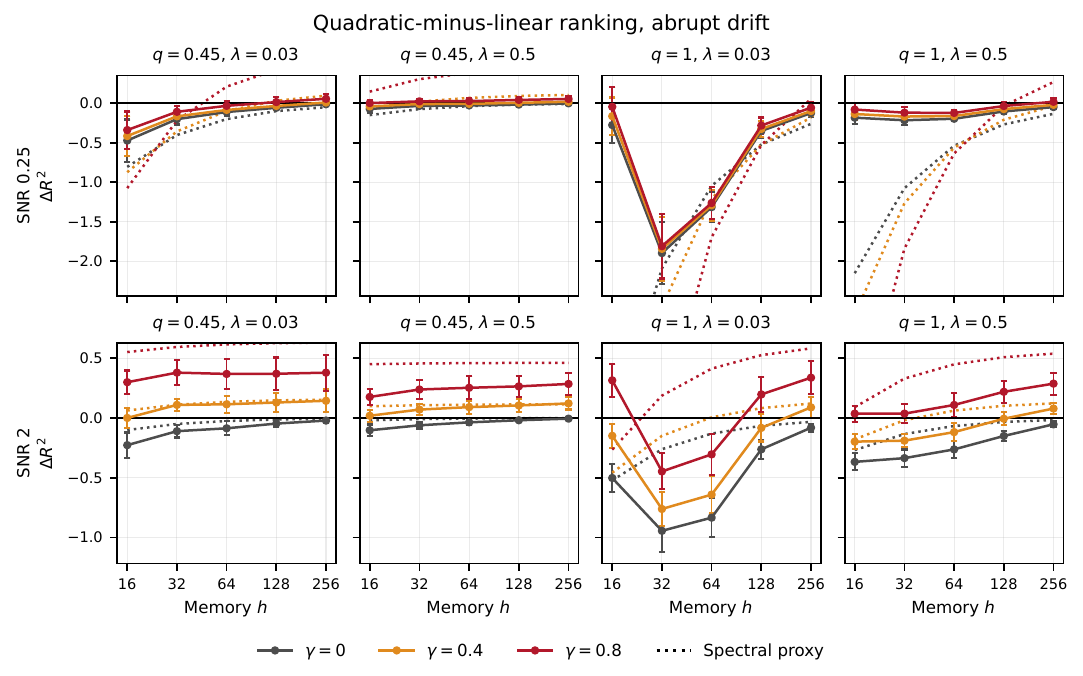}
\caption{Quadratic-minus-linear $R^2$ gap versus memory under abrupt drift. Rows fix SNR and columns fix covariance decay $q$ and penalty $\lambda$. Solid curves with bars are the observed gap and its paired 95\% interval; colors give nonlinear strength $\gamma$. Dotted curves are the leading spectral proxy. Positive values favor the quadratic model. The vertical range of each row covers the observed intervals, so a proxy curve that leaves the panel has missed the observed gap by a large margin. The intervals are descriptive and unadjusted for multiplicity.}
\label{fig:sim-phase-abrupt}
\end{figure}

%% file: appendix_main_experiment.tex
\graphicspath{{figures/stage5/}{figures/controlled_simulations/}{figures/}}

\clearpage
\section{Empirical dynamics of forecast components}
\label{app:stage5-results}

\subsection{Experimental method}
\label{app:stage5-method}

\paragraph{Targets and sample.}
Let $t=1,\dots,T$ index the $T=7{,}374$ trading days from 1 September 1987 through 30 November 2016, in the calendar of the Kenneth French daily files.
Let $j=1,\dots,17$ index the portfolios of the Kenneth French 17-industry classification: Food, Mines, Oil, Clths, Durbl, Chems, Cnsum, Cnstr, Steel, FabPr, Machn, Cars, Trans, Utils, Rtail, Finan, and Other.
Let $R_{tj}$ be the value-weighted daily return of portfolio $j$ in percent, and let $r_t$ be the daily risk-free rate in percent.
We define the target $y_{tj}=(R_{tj}-r_t)/100$, the daily excess return in decimals.
Both series come from the Kenneth French data library, in the vintage built from the 202607 CRSP database.
This vintage can differ slightly from the one that the source paper used.
We fit and score each industry separately, so no parameter is shared across industries.

\paragraph{Forecast origins and test days.}
Let $m$ run over the 323 calendar months from January 1990 through November 2016.
Let $\mathcal D_m$ be the set of trading days in month $m$.
Each $\mathcal D_m$ holds 15 to 23 days, with a median of 21, and the 323 sets hold 6,784 days.
We fit each model once at the start of month $m$, using only days before month $m$.
Preprocessing, validation, early stopping, and tuning also use only days before month $m$.
We use that fit to forecast every day in $\mathcal D_m$.
The 590 trading days before 1990 serve only as training days.
Each configuration thus yields $6{,}784\times17=115{,}328$ out-of-sample forecasts per seed.

\paragraph{Benchmark and score.}
Let $S$ be a set of (day, industry) pairs, and let $\mathbf y_S$ be the vector of the targets $y_{tj}$ in $S$.
Let $\hat{\mathbf y}_S$ be a vector of forecasts for the same pairs.
We score the forecasts against the zero forecast by
\begin{equation*}
R^2(\hat{\mathbf y}_S)=1-\frac{\|\mathbf y_S-\hat{\mathbf y}_S\|^2}{\|\mathbf y_S\|^2}.
\end{equation*}
The score is 1 for a perfect forecast, 0 for the zero forecast, and negative for a forecast that is worse than zero.
The norms pool all pairs in $S$, so $R^2$ is a ratio of sums and not an average of finer scores.
The zero benchmark suits daily returns, because their mean is small.
Over 1990--2016 the mean of $y_{tj}$ is 0.03\% per day, against a root mean square of 1.4\%.
The source paper uses the same benchmark.

\paragraph{Inputs of the two branches.}
In the tabular branch, a model maps the vector $x_t$ to the target $y_{tj}$.
In the temporal branch, a model maps the \emph{context} $X^{(L)}_t=(x_{t-L+1},\dots,x_t)\in\mathbb R^{L\times129}$ to $y_{tj}$.
The context holds the predictors of the last $L$ trading days, including day $t$.
We use $L=20$ as the primary context length and $L=60$ as the secondary one, about one and three months of trading days.
We fixed both lengths in advance.
Ridge, XGBoost, and the feed-forward networks receive the \emph{flattened} context, a vector with $129L$ entries (2,580 for $L=20$ and 7,740 for $L=60$).
The recurrent and attention networks receive the ordered context.
Thus all temporal models use the same numbers within a context length, and their differences reflect the model type.
By contrast, a comparison across branches or across context lengths changes the information set, so we do not read it as an effect of the model alone.

\paragraph{Training windows.}
Let $w\in\{16,64,\infty\}$ be the window length in months.
We write $w=\infty$ for the expanding window.
For forecast month $m$, the window holds the trading days of the $w$ calendar months before $m$, or all days before $m$ if $w=\infty$.
No window holds a day of month $m$ or later.
In the temporal branch, a day $t$ may train a model only if all $L$ days of its context lie inside the window.
Hence a short window never uses predictors from before its start.
Test contexts may reach back into the previous month, because those days are observed when we forecast.
The numbers of days appear in Table~\ref{tab:stage5-sizes}.
The sample starts in September 1987, so the 64-month window holds the same days as the expanding window for the 37 forecast months from January 1990 through January 1993.
The 16-month window is complete from the first forecast month.
The memory $h$ of the main text is the number of days in the window before subsampling.

\begin{table}[t]
\centering
\footnotesize
\begin{tabular}{@{}lccccc@{}}
\toprule
Window & Months & Window days & Subsampled days & Network fit days & Validation days \\
\midrule
16 months & 16 & 328--340 & 256--265 & 192--199 & 79--87 \\
64 months & 28--64 & 590--1,352 & 460--1,052 & 263--856 & 246--255 \\
Expanding & 28--350 & 590--7,353 & 460--5,718 & 263--5,522 & 246--255 \\
\bottomrule
\end{tabular}
\caption{Trading days per fit in the tabular branch, from the minimum to the maximum over the 323 forecast months. Window days are the days of the window before subsampling. Network fit days are the subsampled days before the validation block, which train the networks. The counts do not depend on the seed. A temporal fit has $L-1$ fewer window days.}
\label{tab:stage5-sizes}
\end{table}

\paragraph{Training subsamples and seeds.}
Let $s\in\{0,1,\dots,5\}$ be the seed.
Let $n_c$ be the number of window days in calendar month $c$.
For every seed $s$, forecast month $m$, and month $c$ of the window, we draw $\lfloor0.8\,n_c\rfloor$ of these days without replacement.
The draw uses a NumPy generator seeded by $(s,m,c)$.
We call the union of the draws over $c$ the \emph{subsample} of the fit.
All models of a branch, and all context lengths, use the same draws for the same $(s,m,c)$, up to the first month of a window, where the context rule removes some days.
Thus the replicates are paired across models and windows.
The seed $s$ also determines, together with the model, the month, and the window, the random initialization, the minibatch order, and the dropout masks of a network.
Ridge and XGBoost use no random numbers once the subsample is fixed, because we set the XGBoost row and column sampling rates to one.
Hence their variation across seeds comes from the subsample alone. Random forest adds bootstrap randomness, and networks add their own training randomness.
We use six seeds, fewer than the 20 repetitions of the source paper, to limit compute.
Each seed defines one replicate in the sense of the main text.

\paragraph{Preprocessing.}
Let $\mathcal I$ be the set of days that train a fit.
For each predictor $i$, we compute its mean $\mu_i$ and standard deviation $\sigma_i$ over $\mathcal I$, with divisor $|\mathcal I|$.
In the temporal branch, we compute them over all days of all contexts of $\mathcal I$.
We replace $\sigma_i$ by 1 if $\sigma_i<10^{-8}$.
We standardize every predictor to $z_i=(x_i-\mu_i)/\sigma_i$, and we give the same standardized values to all models, including the boosted trees.
Networks also standardize each target with its training mean and standard deviation.
We map their forecasts back to decimal returns before scoring.
Ridge, XGBoost, and random forest fit the unstandardized targets.
Hence every score and every validation criterion uses return units.
We store the predictors and targets in single precision.

\paragraph{Linear models.}
Let $z_t$ be the standardized input of day $t$.
For industry $j$, ridge regression solves
\begin{equation*}
\min_{b_0,\,b}\ \sum_{t\in\mathcal I}\bigl(y_{tj}-b_0-z_t^\top b\bigr)^2+\alpha\|b\|^2 .
\end{equation*}
The intercept $b_0$ is not penalized.
The penalty multiplies the sum of squared errors and not their mean, so $\alpha=1$ shrinks little when $|\mathcal I|$ is large.
We use $\alpha=1$ as a fixed anchor, as in the source paper, and we tune $\alpha$ on a grid as a second ridge model.
The temporal ridge solves the same problem for the flattened context.
Scikit-learn solves the tabular problem directly and the flattened problem with its LSQR solver.

\paragraph{Boosted trees.}
XGBoost fits gradient-boosted regression trees by squared-error loss.
It provides a boosted-tree comparison with the random forest of the source paper.
For industry $j$ we fit 150 rounds with maximum depth $D$ and learning rate $\eta$.
We fix the minimum child weight at 5, the $L_2$ leaf penalty at 1, and the $L_1$ penalty at 0.
We use no row or column subsampling, and we find splits from histograms on the GPU.
All other settings keep their library defaults.
We tune $D$ and $\eta$.
The temporal version uses the flattened context.

\paragraph{Random forest.}
The tabular forest fits a separate regressor for each of the 17 industries using cuML on an NVIDIA A100.
Each regressor grows 200 bootstrap trees, with maximum depth 5, minimum leaf size 1, all 129 predictors eligible at a split, and 128 histogram bins.
The forest settings are fixed across windows and years; there is no validation grid for this anchor.
The same six subsample seeds and monthly forecast origins used by the other tabular models determine its training sets.

\paragraph{Feed-forward networks.}
For industry $j$, a network maps the standardized input to a scalar through one or two ReLU hidden layers.
Dropout with rate $\delta$ follows each hidden layer.
The small network has one hidden layer of width 32.
The medium network has two hidden layers of widths 128 and 64.
The 17 networks of a model train as one batched tensor program, but they share no parameters.
We initialize them with the PyTorch defaults for linear layers.
We keep the networks small, because the samples are small: a 16-month fit has about 260 subsampled days.

\paragraph{Recurrent and attention networks.}
These networks read the ordered context.
The recurrent network (RNN) applies $h_\tau=\tanh(W_x z_\tau+W_h h_{\tau-1}+b)$ for $\tau=t-L+1,\dots,t$, with $h_{t-L}=0$.
The long short-term memory network (LSTM) uses the standard input, forget, and output gates and a memory cell.
Both predict from the last state through dropout and a linear map.
The Transformer embeds the 129 predictors of each day linearly into width $d$ and adds sinusoidal position codes.
It applies two pre-normalization blocks.
Each block has multi-head self-attention with 4 heads and a residual connection, then a feed-forward layer of width $2d$ with a GELU activation and a residual connection.
Dropout acts on the attention weights and inside each residual branch.
A final layer normalization and a linear map turn the state of day $t$ into the forecast.
Each industry again has its own network.
We list the settings, the candidate grids, and the parameter counts of all models in Table~\ref{tab:stage5-models}.

\begin{table}[t]
\centering
\footnotesize
\begin{tabular}{@{}>{\raggedright\arraybackslash}p{3.0cm}>{\raggedright\arraybackslash}p{1.2cm}>{\raggedright\arraybackslash}p{5.0cm}>{\raggedright\arraybackslash}p{6.3cm}@{}}
\toprule
Model & Input & Fixed settings & Candidate grid (eight candidates) \\
\midrule
\multicolumn{4}{@{}l}{\emph{Tabular branch}} \\
Ridge, fixed alpha & $x_t$ & $\alpha=1$ & none \\
Tuned ridge & $x_t$ & none & $\alpha\in\{10^{-4},10^{-2},0.1,1,10,100,10^{4},10^{6}\}$ \\
XGBoost & $x_t$ & 150 rounds; min.\ child weight 5; $L_2$ penalty 1; no subsampling & $D\in\{2,3,5,7\}$ and $\eta\in\{0.03,0.1\}$ \\
Random forest & $x_t$ & cuML; 200 bootstrap trees; depth 5; min.\ leaf 1; all features; 128 bins & none \\
Small MLP & $x_t$ & width (32); 4,193 parameters; batch 256; at most 30 epochs; patience 5 & learning rate $\{0.001,0.003\}$, weight decay $\{0,0.01\}$, dropout $\{0,0.1\}$ \\
Medium MLP & $x_t$ & widths (128, 64); 24,961 parameters; batch 256; at most 30 epochs; patience 5 & same as the small MLP \\
\midrule
\multicolumn{4}{@{}l}{\emph{Temporal branch}} \\
Flattened ridge & flat & $\alpha=1$ & none \\
Flattened XGBoost & flat & as XGBoost & as XGBoost \\
Small flattened MLP & flat & width (32); learning rate 0.002; weight decay 0.01; dropout 0; batch 128; at most 20 epochs; patience 4; 82,625 parameters ($L=20$), 247,745 ($L=60$) & none \\
Tuned flattened MLP & flat & weight decay 0; batch 128; at most 20 epochs; patience 4; 61,969 or 82,625 parameters ($L=20$), 185,809 or 247,745 ($L=60$) & width $\{(24),(32)\}$, learning rate $\{0.001,0.003\}$, dropout $\{0,0.1\}$ \\
Vanilla RNN & ordered & weight decay 0; batch 128; at most 20 epochs; patience 4; 3,721 or 5,217 parameters & hidden size $\{24,32\}$, learning rate $\{0.001,0.003\}$, dropout $\{0,0.1\}$ \\
LSTM & ordered & same training settings; 6,829 or 9,361 parameters & hidden size $\{12,16\}$, learning rate $\{0.001,0.003\}$, dropout $\{0,0.1\}$ \\
Small Transformer & ordered & 2 blocks; 4 heads; feed-forward width $2d$; same training settings; 6,577 or 12,937 parameters & width $d\in\{16,24\}$, learning rate $\{0.001,0.003\}$, dropout $\{0,0.1\}$ \\
\bottomrule
\end{tabular}
\caption{Model specifications. The input is the predictor vector $x_t$, the flattened context (flat), or the ordered context. Each tuned model has eight candidates, which are all combinations of the listed values. Parameter counts are trainable parameters per industry. When a cell lists two counts, they belong to the two grid values in the listed order. All networks use AdamW and clip the gradient norm at 1.0.}
\label{tab:stage5-models}
\end{table}

\paragraph{Hyperparameter selection.}
Ridge with $\alpha=1$, the tabular random forest, and the small flattened network have fixed settings, so no tuning choice affects these anchors.
We tune every other model once per calendar year, and separately for each context length.
Each tuned model has eight candidates (Table~\ref{tab:stage5-models}).
For window $w$ and forecast month $m$, the \emph{validation block} is the last $v_w$ months before $m$, with $v_{16}=4$ and $v_{64}=v_\infty=12$.
Let $\mathcal V_w$ be the days of the block.
We use all of them, without subsampling.
Let $\mathcal I_{w,s}$ be the subsampled days of seed $s$ that precede the block.
For the January refit of the year, we fit each candidate $\theta$ on $\mathcal I_{w,s}$ for the three windows and the seeds $s\in\{0,1\}$, and we predict $\mathcal V_w$.
For each industry we compute the relative squared error
\begin{equation*}
\mathrm{RSE}_j(\theta,w,s)=\frac{\sum_{t\in\mathcal V_w}\bigl(y_{tj}-\hat y_{tj}\bigr)^2}{\sum_{t\in\mathcal V_w}y_{tj}^2}.
\end{equation*}
The score of candidate $\theta$ is the mean of $\mathrm{RSE}_j(\theta,w,s)$ over the 17 industries, the three windows, and the two seeds.
We select the candidate with the smallest score.
The selection uses only days before 1 January of that year.
It then holds for all monthly refits of the year (eleven in 2016), for all three windows, and for all seeds.
Thus every window uses the same hyperparameters, and a contrast between windows holds the model capacity fixed.
The division by $\sum y^2$ keeps volatile industries from dominating the choice, and it keeps the zero benchmark.
The validation days always follow the training days, so no step uses random cross-validation.
The frozen configuration files fix the grids and the validation lengths for all runs.

\paragraph{Network training.}
We minimize the mean squared error of the standardized targets with AdamW.
We clip the gradient norm at 1.0, and we use the batch sizes in Table~\ref{tab:stage5-models}.
After each epoch we compute the validation score: the mean over industries of $\mathrm{RSE}_j$ on $\mathcal V_w$, in return units.
Training stops after the maximum number of epochs, or when the validation score has not improved for the patience number of epochs.
We keep the weights of the epoch with the best validation score.
A forecast fit of a network trains on $\mathcal I_{w,s}$, and it uses the validation block only to pick the epoch.
By contrast, ridge, XGBoost, and random forest train on all subsampled days of the window, including the last $v_w$ months.
Consequently, the gradient updates of a network see only 12 of the 16 months in the shortest window, 52 of 64 months in the 64-month window, and all but the last 12 months in the expanding window (Table~\ref{tab:stage5-sizes}).

\paragraph{Replicate statistics and exact decomposition.}
Fix a configuration (a model, a context length, and a window) and a set $S$.
Let $\hat{\mathbf y}^{(s)}_S$ be the forecasts of seed $s$ for the pairs in $S$, and let $M=6$ be the number of seeds.
The score of seed $s$ is $R^2_s=R^2(\hat{\mathbf y}^{(s)}_S)$.
The \emph{mean seed score} is $\overline{R^2}=M^{-1}\sum_sR^2_s$.
We also report the sample standard deviation $\hat\sigma$ of the six scores, with divisor $M-1$, and the interval $\overline{R^2}\pm t_{0.975,5}\,\hat\sigma/\sqrt M$ with $t_{0.975,5}=2.5706$.
Let $\bar{\mathbf y}_S=M^{-1}\sum_s\hat{\mathbf y}^{(s)}_S$ be the mean forecast.
The \emph{ensemble score} is $R^2(\bar{\mathbf y}_S)$.
Write $Y_2=\|\mathbf y_S\|^2$, and define
\begin{equation*}
a=\frac{\langle\mathbf y_S,\bar{\mathbf y}_S\rangle}{\|\mathbf y_S\|\,\|\bar{\mathbf y}_S\|},\qquad
g=\frac{\|\bar{\mathbf y}_S\|}{\|\mathbf y_S\|},\qquad
u=\frac{M^{-1}\sum_s\|\hat{\mathbf y}^{(s)}_S-\bar{\mathbf y}_S\|^2}{Y_2}.
\end{equation*}
We call $a$ the alignment, $g$ the aggressiveness, and $u$ the instability.
The alignment is the cosine between the targets and the mean forecast.
The aggressiveness is the size of the mean forecast relative to the targets.
The instability is the mean squared distance of the seed forecasts from their mean, relative to $Y_2$.
Expanding the squares gives the exact identity
\begin{equation*}
\overline{R^2}=2ag-g^2-u,\qquad R^2(\bar{\mathbf y}_S)=2ag-g^2=\overline{R^2}+u .
\end{equation*}
We call $2ag$ the alignment reward, $g^2$ the aggressiveness cost, and $u$ the instability cost.
We compute all three from unrounded forecasts, and the identity holds to within $10^{-9}$ in all 351,540 scope rows.
Because the seeds vary both the subsample and the fitting randomness, $u$ mixes both sources of variation.

\paragraph{Scopes and uncertainty.}
We compute these statistics for five scopes $S$.
The \emph{overall} scope holds all 6,784 days and 17 industries.
The \emph{industry} scope holds all days of one industry.
The \emph{year} scope holds all industries and days of one calendar year (2016 has 11 months).
The \emph{month} scope holds all industries and days of one forecast month.
The \emph{month-by-industry} scope holds the days of one forecast month for one industry.
Each scope pools its sums of squares before it forms a score, so the score of a coarse scope is not the average of the scores of finer scopes.
This appendix displays the overall scope in tables, and the month and month-by-industry scopes in curve plots.
The industry and year scopes remain in the performance files.
The interval above describes the variation of the training seeds and subsamples only.
It ignores serial dependence across days, dependence across industries, changes of regime, and the choice of hyperparameters.
This appendix reports no test of equal predictive accuracy.
For the contrast between the two context lengths, we compute $d_s=R^2_s(L=20)-R^2_s(L=60)$ for each seed $s$ of one model and window.
We report the mean $\bar d$ of the six values and the interval $\bar d\pm t_{0.975,5}\,\mathrm{sd}(d_s)/\sqrt M$.
The two contexts share the seed indices, but their subsamples differ slightly at the start of each window.

\paragraph{Computation and reproducibility.}
We ran the experiment as resumable jobs on the Vessl cloud service, each with one NVIDIA A100 GPU.
There is one fitting job for the classical models and one for the neural models of each branch, and one aggregation job per branch.
Each job splits its work into resumable tasks by calendar year and model group.
XGBoost uses CUDA, and ridge uses the CPU.
The jobs install PyTorch 2.6.0 (CUDA 12.4), NumPy 2.2.6, pandas 2.2.3, scikit-learn 1.7.2, and XGBoost 3.0.5, and they request deterministic kernels where PyTorch has them.
Before a job starts, it checks the predictor panel against a SHA-256 hash.
Every fit writes a checkpoint whose hash covers the configuration, the design, the training, validation, and test rows, the seeds, and the forecasts.
The aggregation step checks every hash and refuses to continue when a checkpoint is missing or does not match.

\subsection{Results of tabular models}
\begin{table}[h]
\centering
\scriptsize
\setlength{\tabcolsep}{4pt}
\begin{tabular}{@{}llccc@{}}
\toprule
Model & & 16 months & 64 months & Expanding \\
\midrule
Ridge, fixed alpha & Ensemble & 0.0706 & 0.3481 & 0.3477 \\
\cmidrule[0.15pt](l){2-5}
 & Mean [95\% CI] & $-$0.0909 [$-$0.1033, $-$0.0786] & 0.3256 [0.3235, 0.3277] & 0.3386 [0.3374, 0.3397] \\
\midrule
Tuned ridge & Ensemble & 0.3671 & 0.3767 & 0.3644 \\
\cmidrule[0.15pt](l){2-5}
 & Mean [95\% CI] & 0.3515 [0.3487, 0.3543] & 0.3668 [0.3658, 0.3678] & 0.3602 [0.3593, 0.3611] \\
\midrule
XGBoost & Ensemble & 0.3621 & 0.3619 & 0.3392 \\
\cmidrule[0.15pt](l){2-5}
 & Mean [95\% CI] & 0.3316 [0.3284, 0.3348] & 0.3419 [0.3380, 0.3457] & 0.3232 [0.3202, 0.3262] \\
\midrule
Small MLP & Ensemble & 0.3096 & 0.3414 & 0.3423 \\
\cmidrule[0.15pt](l){2-5}
 & Mean [95\% CI] & 0.2869 [0.2841, 0.2898] & 0.3231 [0.3212, 0.3251] & 0.3301 [0.3274, 0.3328] \\
\midrule
Medium MLP & Ensemble & 0.3018 & 0.3383 & 0.3425 \\
\cmidrule[0.15pt](l){2-5}
 & Mean [95\% CI] & 0.2786 [0.2756, 0.2816] & 0.3189 [0.3151, 0.3228] & 0.3270 [0.3232, 0.3308] \\
\midrule
Random forest & Ensemble & 0.3705 & 0.3494 & 0.3079 \\
\cmidrule[0.15pt](l){2-5}
 & Mean [95\% CI] & 0.3549 [0.3514, 0.3583] & 0.3374 [0.3341, 0.3407] & 0.2953 [0.2826, 0.3080] \\
\bottomrule
\end{tabular}
\caption{Tabular: pooled out-of-sample $R^2$ for every reported model and training window. In each cell, the upper line gives the $R^2$ of the six-seed ensemble, which scores the mean forecast. The lower line gives the mean of the six seed scores, with its seed-only 95\% $t$ interval in brackets.}
\label{tab:stage5-overall-tabular}
\end{table}

\noindent The tabular random forest has mean seed $R^2$ of 0.3549, 0.3374, and 0.2953 for the 16-month, 64-month, and expanding windows, respectively. Its corresponding ensemble scores are 0.3705, 0.3494, and 0.3079. The 16-month window scores highest for this model on both measures.
\subsubsection{Month-by-industry curves}
\noindent We show one model per figure, with one panel per training window: 16 months, 64 months, and expanding. Each panel draws 17 curves, one per industry. A curve gives the six-seed mean of the pooled $R^2$ for that industry in each month. All panels share the vertical axis $[-2.0,1.0]$, and each industry keeps one color throughout. Values below $-2.0$ fall outside the frame, and each caption counts them.

\begin{figure}[h]
\centering
\includegraphics[width=0.9\textwidth]{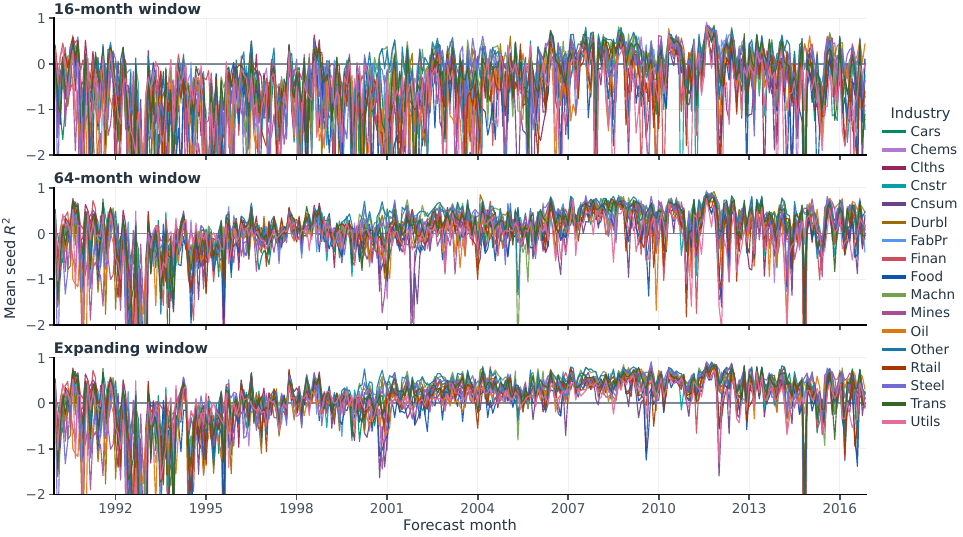}
\caption{Tabular, Ridge, fixed alpha: monthly pooled $R^2$ for the 17 industries, one curve each, under the 16-month, 64-month, and expanding windows. Values below $-2.0$ fall outside the frame: 393, 83, and 90 of 5,491 in the three panels.}
\label{fig:stage5-month-industry-tabular-ridge_alpha1}
\end{figure}

\begin{figure}[h]
\centering
\includegraphics[width=0.9\textwidth]{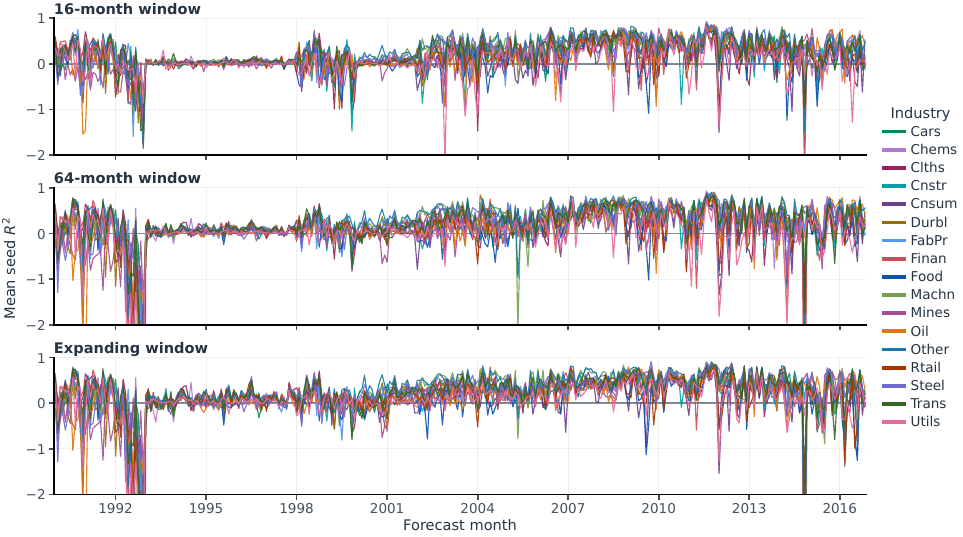}
\caption{Tabular, Tuned ridge: monthly pooled $R^2$ for the 17 industries, one curve each, under the 16-month, 64-month, and expanding windows. Values below $-2.0$ fall outside the frame: 2, 51, and 52 of 5,491 in the three panels.}
\label{fig:stage5-month-industry-tabular-ridge_tuned}
\end{figure}

\begin{figure}[h]
\centering
\includegraphics[width=0.9\textwidth]{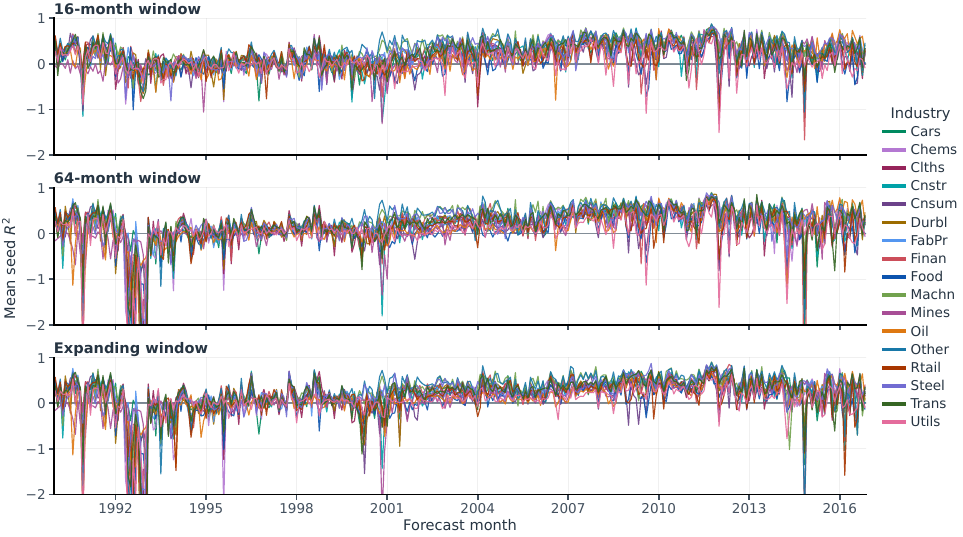}
\caption{Tabular, XGBoost: monthly pooled $R^2$ for the 17 industries, one curve each, under the 16-month, 64-month, and expanding windows. Values below $-2.0$ fall outside the frame: 0, 36, and 38 of 5,491 in the three panels.}
\label{fig:stage5-month-industry-tabular-xgboost}
\end{figure}

\begin{figure}[h]
\centering
\includegraphics[width=0.9\textwidth]{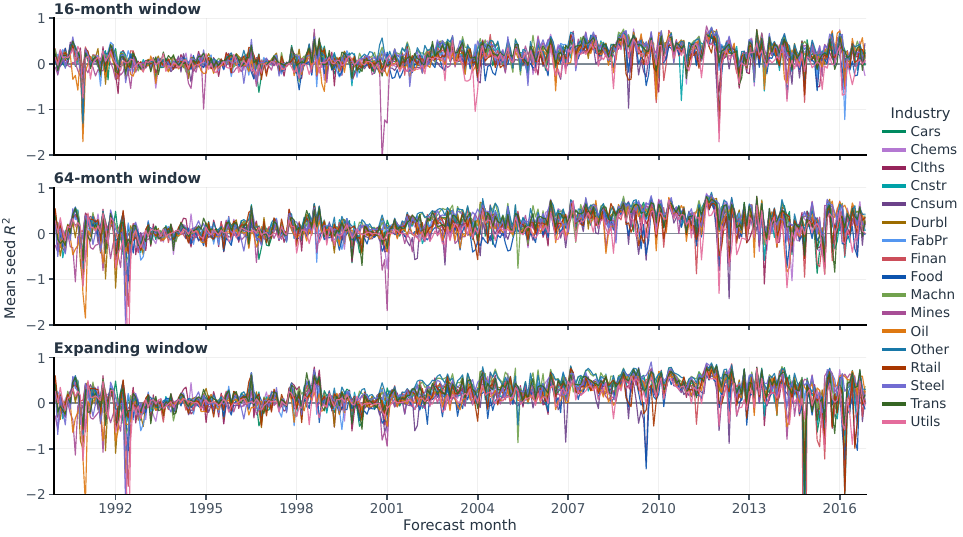}
\caption{Tabular, Small MLP: monthly pooled $R^2$ for the 17 industries, one curve each, under the 16-month, 64-month, and expanding windows. Values below $-2.0$ fall outside the frame: 1, 2, and 10 of 5,491 in the three panels.}
\label{fig:stage5-month-industry-tabular-mlp_small_tuned}
\end{figure}

\begin{figure}[h]
\centering
\includegraphics[width=0.9\textwidth]{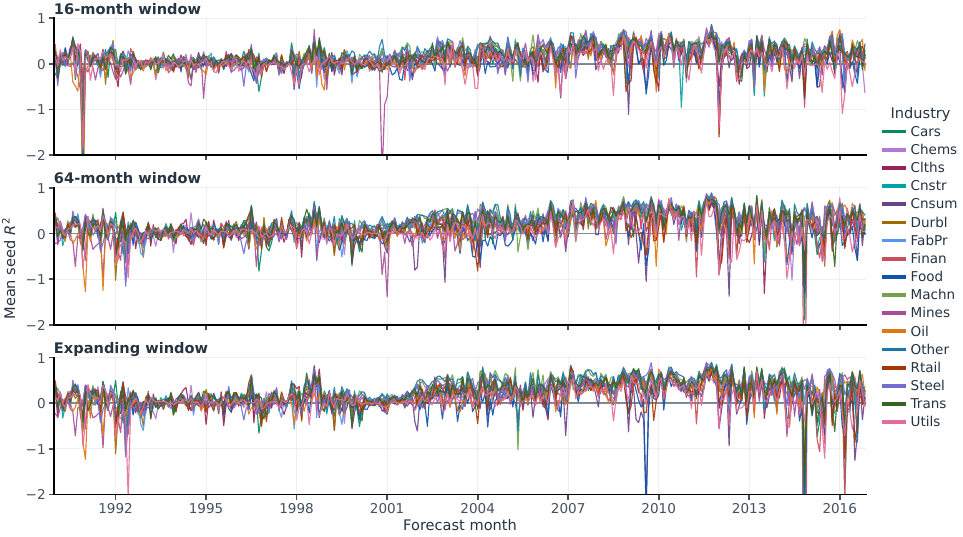}
\caption{Tabular, Medium MLP: monthly pooled $R^2$ for the 17 industries, one curve each, under the 16-month, 64-month, and expanding windows. Values below $-2.0$ fall outside the frame: 6, 1, and 7 of 5,491 in the three panels.}
\label{fig:stage5-month-industry-tabular-mlp_medium_tuned}
\end{figure}

\begin{figure}[h]
\centering
\includegraphics[width=0.9\textwidth]{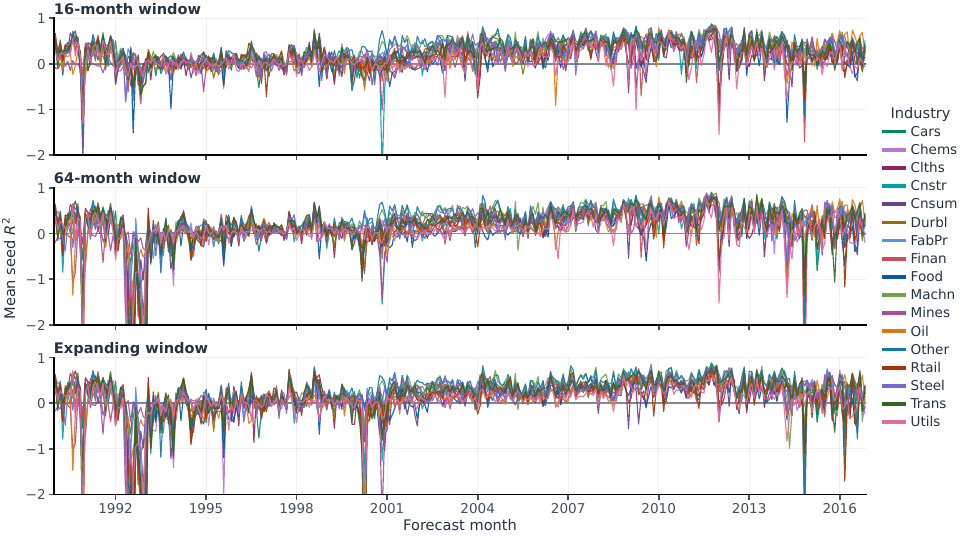}
\caption{Tabular, Random forest: monthly pooled $R^2$ for the 17 industries, one curve each, under the 16-month, 64-month, and expanding windows. Values below $-2.0$ fall outside the frame: 2, 55, and 63 of 5,491 in the three panels.}
\label{fig:stage5-month-industry-tabular-random_forest_anchor}
\end{figure}

\clearpage
\subsection{Results of temporal models, 20-day context}
\begin{table}[h]
\centering
\scriptsize
\setlength{\tabcolsep}{4pt}
\begin{tabular}{@{}llccc@{}}
\toprule
Model & & 16 months & 64 months & Expanding \\
\midrule
Flattened ridge & Ensemble & 0.1915 & $-$0.0139 & $-$0.3308 \\
\cmidrule[0.15pt](l){2-5}
 & Mean [95\% CI] & 0.1397 [0.1346, 0.1447] & $-$0.2128 [$-$0.2208, $-$0.2049] & $-$1.1533 [$-$1.1871, $-$1.1194] \\
\midrule
Flattened XGBoost & Ensemble & 0.3341 & 0.3485 & 0.3379 \\
\cmidrule[0.15pt](l){2-5}
 & Mean [95\% CI] & 0.3026 [0.2992, 0.3061] & 0.3287 [0.3258, 0.3315] & 0.3208 [0.3170, 0.3247] \\
\midrule
Small flattened MLP & Ensemble & 0.0200 & 0.1544 & 0.2485 \\
\cmidrule[0.15pt](l){2-5}
 & Mean [95\% CI] & $-$0.0841 [$-$0.0926, $-$0.0756] & 0.0574 [0.0532, 0.0617] & 0.1623 [0.1552, 0.1694] \\
\midrule
Tuned flattened MLP & Ensemble & 0.0406 & 0.1715 & 0.2607 \\
\cmidrule[0.15pt](l){2-5}
 & Mean [95\% CI] & $-$0.0350 [$-$0.0559, $-$0.0141] & 0.1160 [0.1112, 0.1209] & 0.2066 [0.2038, 0.2095] \\
\midrule
Vanilla RNN & Ensemble & 0.3166 & 0.3166 & 0.3315 \\
\cmidrule[0.15pt](l){2-5}
 & Mean [95\% CI] & 0.2860 [0.2834, 0.2887] & 0.2987 [0.2970, 0.3005] & 0.3184 [0.3161, 0.3206] \\
\midrule
LSTM & Ensemble & 0.2317 & 0.2673 & 0.3071 \\
\cmidrule[0.15pt](l){2-5}
 & Mean [95\% CI] & 0.2087 [0.2059, 0.2114] & 0.2491 [0.2463, 0.2519] & 0.2905 [0.2886, 0.2925] \\
\midrule
Small Transformer & Ensemble & 0.3218 & 0.3191 & 0.3361 \\
\cmidrule[0.15pt](l){2-5}
 & Mean [95\% CI] & 0.2828 [0.2789, 0.2866] & 0.2995 [0.2972, 0.3018] & 0.3193 [0.3172, 0.3213] \\
\bottomrule
\end{tabular}
\caption{Temporal, 20-day context: pooled out-of-sample $R^2$ for every completed model and training window. In each cell, the upper line gives the $R^2$ of the six-seed ensemble, which scores the mean forecast. The lower line gives the mean of the six seed scores, with its seed-only 95\% $t$ interval in brackets.}
\label{tab:stage5-overall-temporal20}
\end{table}

\begin{figure}[h]
\centering
\includegraphics[width=0.9\textwidth]{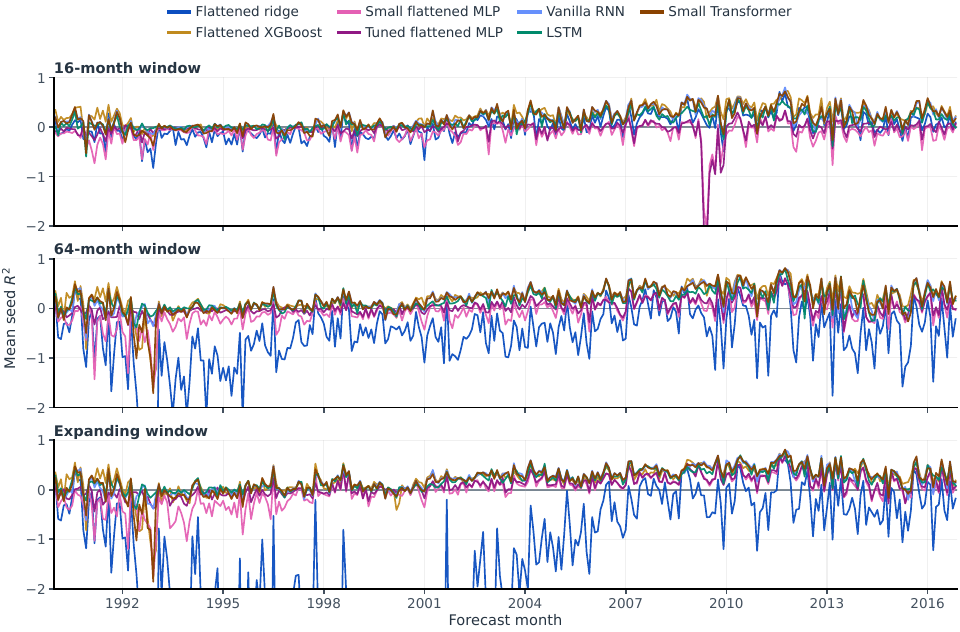}
\caption{Temporal, 20-day context: monthly pooled $R^2$ of every completed model, one curve each, under the 16-month, 64-month, and expanding windows. Each month pools all 17 industries and the trading days of that month. Values below $-2.0$ fall outside the frame: 3, 11, and 112 of 2,261 model-months in the three panels.}
\label{fig:stage5-month-temporal20}
\end{figure}

\subsubsection{Month-by-industry curves}
\noindent We show one model per figure, with one panel per training window: 16 months, 64 months, and expanding. Each panel draws 17 curves, one per industry. A curve gives the six-seed mean of the pooled $R^2$ for that industry in each month. All panels share the vertical axis $[-2.0,1.0]$, and each industry keeps one color throughout. Values below $-2.0$ fall outside the frame, and each caption counts them.
\clearpage
\begin{figure}[h]
\centering
\includegraphics[width=0.9\textwidth]{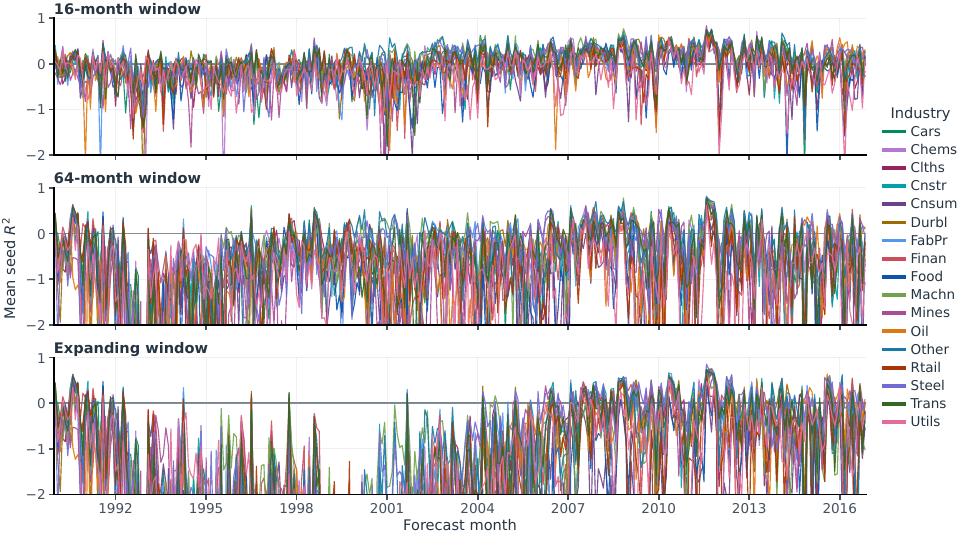}
\caption{Temporal, 20-day context, Flattened ridge: monthly pooled $R^2$ for the 17 industries, one curve each, under the 16-month, 64-month, and expanding windows. Values below $-2.0$ fall outside the frame: 14, 552, and 1,962 of 5,491 in the three panels.}
\label{fig:stage5-month-industry-temporal20-ridge_alpha1_flat}
\end{figure}

\begin{figure}[h]
\centering
\includegraphics[width=0.9\textwidth]{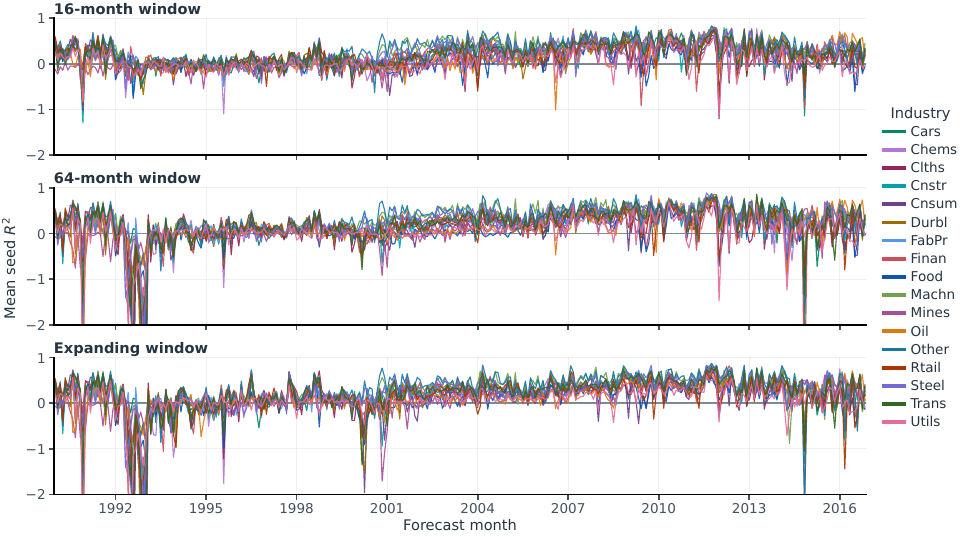}
\caption{Temporal, 20-day context, Flattened XGBoost: monthly pooled $R^2$ for the 17 industries, one curve each, under the 16-month, 64-month, and expanding windows. Values below $-2.0$ fall outside the frame: 0, 29, and 29 of 5,491 in the three panels.}
\label{fig:stage5-month-industry-temporal20-xgboost_flat}
\end{figure}

\begin{figure}[h]
\centering
\includegraphics[width=0.9\textwidth]{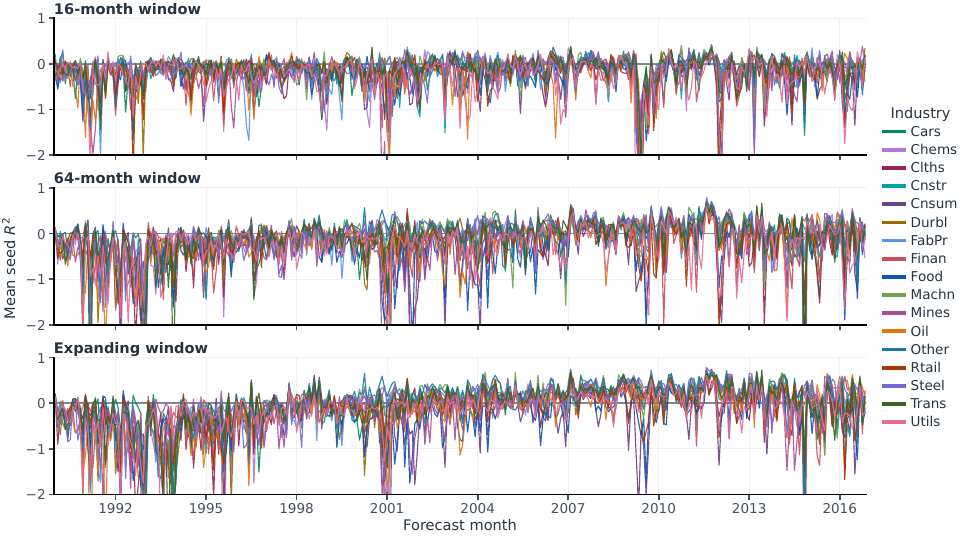}
\caption{Temporal, 20-day context, Small flattened MLP: monthly pooled $R^2$ for the 17 industries, one curve each, under the 16-month, 64-month, and expanding windows. Values below $-2.0$ fall outside the frame: 31, 46, and 42 of 5,491 in the three panels.}
\label{fig:stage5-month-industry-temporal20-mlp_small_anchor_flat}
\end{figure}

\begin{figure}[h]
\centering
\includegraphics[width=0.9\textwidth]{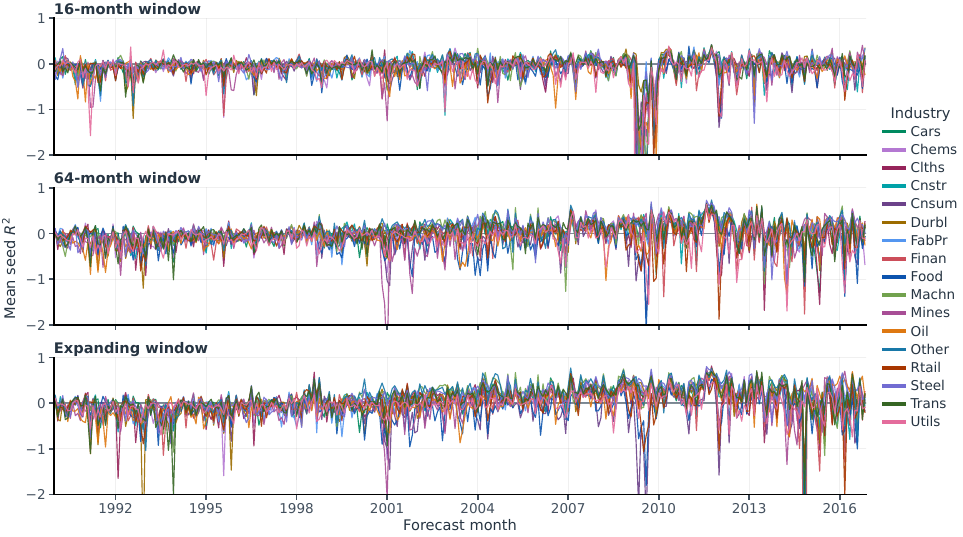}
\caption{Temporal, 20-day context, Tuned flattened MLP: monthly pooled $R^2$ for the 17 industries, one curve each, under the 16-month, 64-month, and expanding windows. Values below $-2.0$ fall outside the frame: 23, 2, and 9 of 5,491 in the three panels.}
\label{fig:stage5-month-industry-temporal20-mlp_tuned_flat}
\end{figure}

\begin{figure}[h]
\centering
\includegraphics[width=0.9\textwidth]{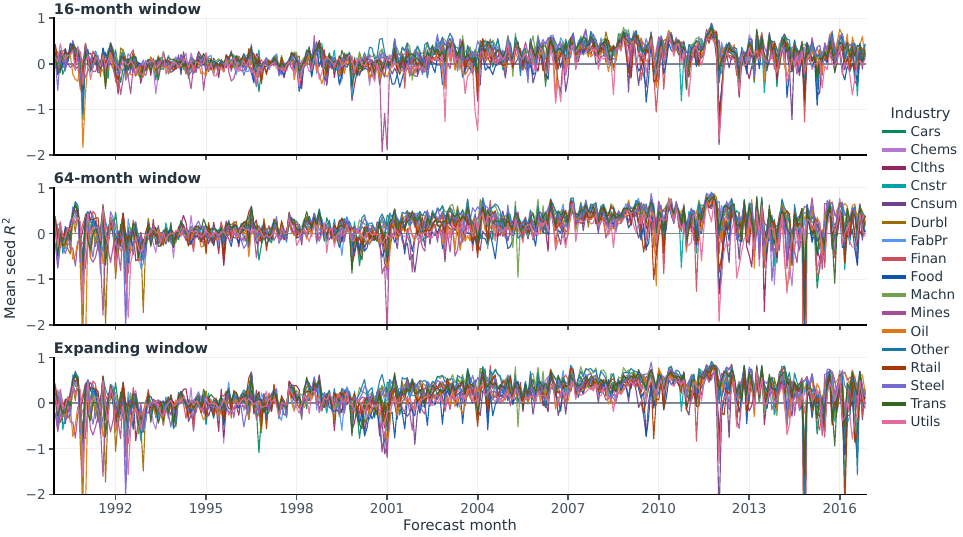}
\caption{Temporal, 20-day context, Vanilla RNN: monthly pooled $R^2$ for the 17 industries, one curve each, under the 16-month, 64-month, and expanding windows. Values below $-2.0$ fall outside the frame: 0, 9, and 9 of 5,491 in the three panels.}
\label{fig:stage5-month-industry-temporal20-rnn_tuned}
\end{figure}

\begin{figure}[h]
\centering
\includegraphics[width=0.9\textwidth]{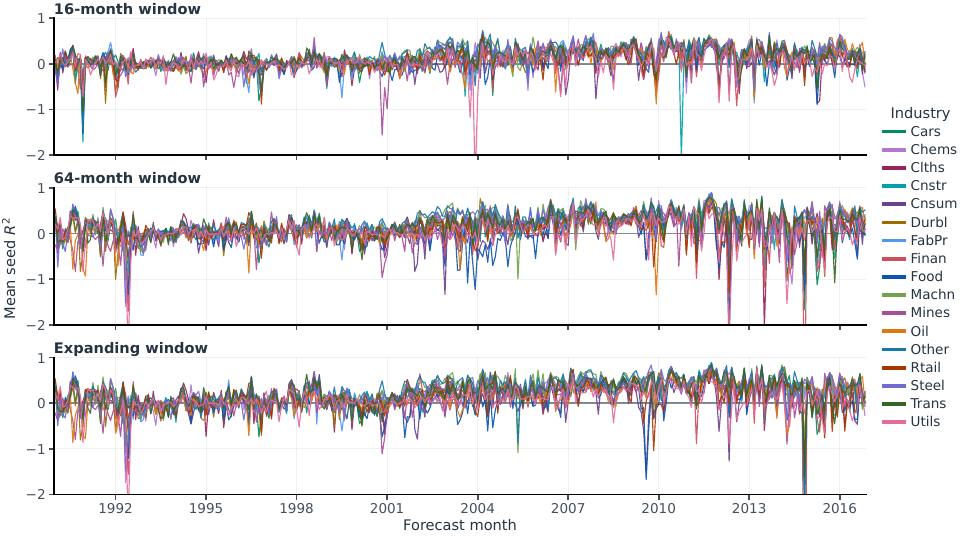}
\caption{Temporal, 20-day context, LSTM: monthly pooled $R^2$ for the 17 industries, one curve each, under the 16-month, 64-month, and expanding windows. Values below $-2.0$ fall outside the frame: 2, 6, and 4 of 5,491 in the three panels.}
\label{fig:stage5-month-industry-temporal20-lstm_tuned}
\end{figure}

\begin{figure}[h]
\centering
\includegraphics[width=0.9\textwidth]{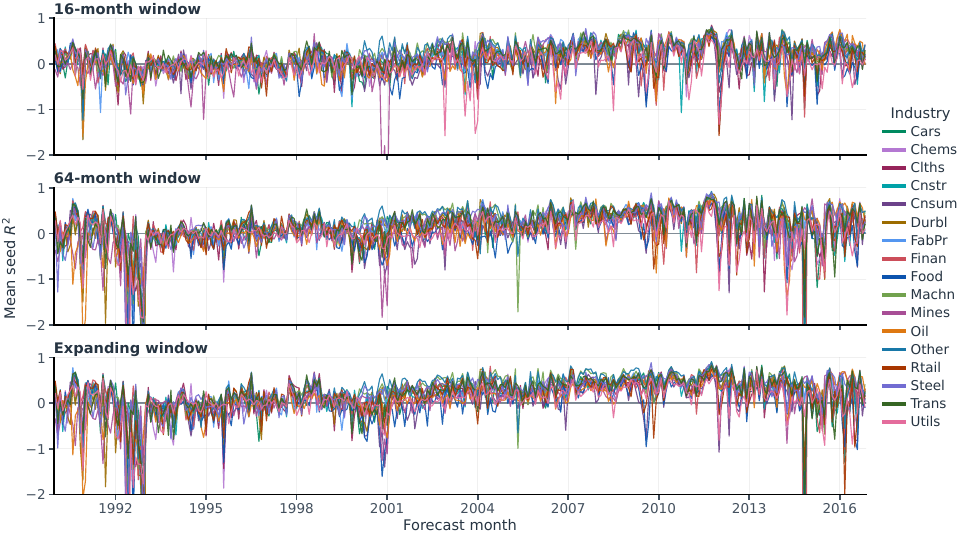}
\caption{Temporal, 20-day context, Small Transformer: monthly pooled $R^2$ for the 17 industries, one curve each, under the 16-month, 64-month, and expanding windows. Values below $-2.0$ fall outside the frame: 2, 21, and 22 of 5,491 in the three panels.}
\label{fig:stage5-month-industry-temporal20-transformer_tuned}
\end{figure}

\clearpage
\subsection{Results of temporal models, 60-day context}
\begin{table}[h]
\centering
\scriptsize
\setlength{\tabcolsep}{4pt}
\begin{tabular}{@{}llccc@{}}
\toprule
Model & & 16 months & 64 months & Expanding \\
\midrule
Flattened ridge & Ensemble & 0.1092 & 0.1654 & $-$0.0295 \\
\cmidrule[0.15pt](l){2-5}
 & Mean [95\% CI] & 0.0772 [0.0712, 0.0831] & 0.1078 [0.1018, 0.1138] & $-$0.2835 [$-$0.2946, $-$0.2724] \\
\midrule
Flattened XGBoost & Ensemble & 0.3161 & 0.3402 & 0.3253 \\
\cmidrule[0.15pt](l){2-5}
 & Mean [95\% CI] & 0.2812 [0.2779, 0.2845] & 0.3223 [0.3188, 0.3257] & 0.3148 [0.3132, 0.3164] \\
\midrule
Small flattened MLP & Ensemble & $-$0.3891 & 0.0765 & 0.1690 \\
\cmidrule[0.15pt](l){2-5}
 & Mean [95\% CI] & $-$0.8216 [$-$0.8845, $-$0.7588] & $-$0.0124 [$-$0.0180, $-$0.0068] & 0.0881 [0.0846, 0.0916] \\
\midrule
Tuned flattened MLP & Ensemble & $-$0.0970 & 0.0804 & 0.1822 \\
\cmidrule[0.15pt](l){2-5}
 & Mean [95\% CI] & $-$0.2223 [$-$0.2347, $-$0.2098] & 0.0002 [$-$0.0034, 0.0039] & 0.1158 [0.1121, 0.1196] \\
\midrule
Vanilla RNN & Ensemble & 0.3123 & 0.3167 & 0.3276 \\
\cmidrule[0.15pt](l){2-5}
 & Mean [95\% CI] & 0.2776 [0.2747, 0.2806] & 0.2989 [0.2969, 0.3009] & 0.3169 [0.3144, 0.3194] \\
\midrule
LSTM & Ensemble & 0.2153 & 0.2661 & 0.3042 \\
\cmidrule[0.15pt](l){2-5}
 & Mean [95\% CI] & 0.1913 [0.1873, 0.1952] & 0.2472 [0.2455, 0.2489] & 0.2885 [0.2876, 0.2895] \\
\midrule
Small Transformer & Ensemble & 0.3191 & 0.3160 & 0.3342 \\
\cmidrule[0.15pt](l){2-5}
 & Mean [95\% CI] & 0.2712 [0.2658, 0.2766] & 0.2969 [0.2961, 0.2976] & 0.3195 [0.3178, 0.3212] \\
\bottomrule
\end{tabular}
\caption{Temporal, 60-day context: pooled out-of-sample $R^2$ for every completed model and training window. In each cell, the upper line gives the $R^2$ of the six-seed ensemble, which scores the mean forecast. The lower line gives the mean of the six seed scores, with its seed-only 95\% $t$ interval in brackets.}
\label{tab:stage5-overall-temporal60}
\end{table}

\begin{figure}[h]
\centering
\includegraphics[width=0.9\textwidth]{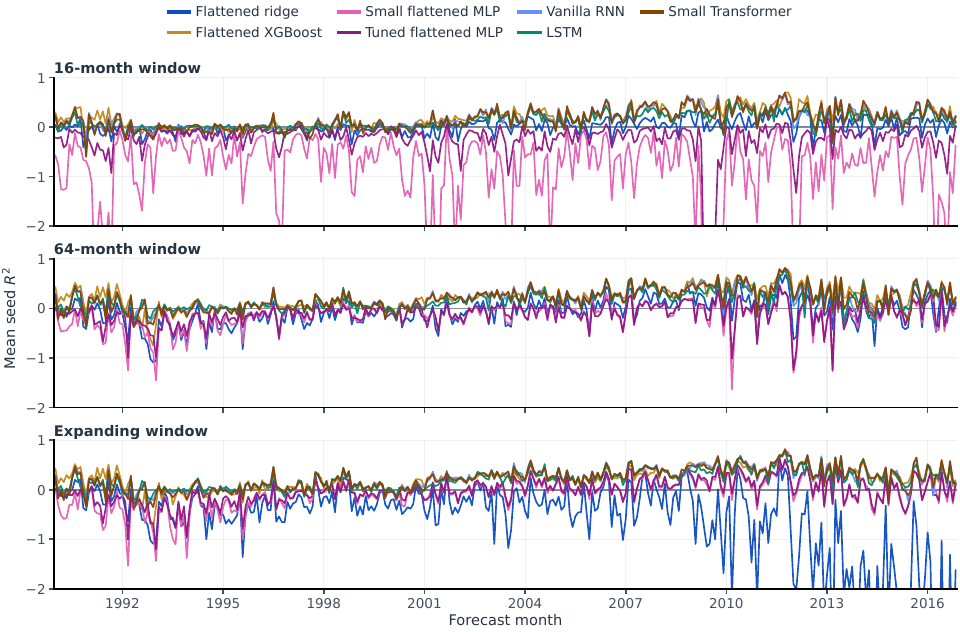}
\caption{Temporal, 60-day context: monthly pooled $R^2$ of every completed model, one curve each, under the 16-month, 64-month, and expanding windows. Each month pools all 17 industries and the trading days of that month. Values below $-2.0$ fall outside the frame: 32, 0, and 27 of 2,261 model-months in the three panels.}
\label{fig:stage5-month-temporal60}
\end{figure}

\subsubsection{Month-by-industry curves}
\noindent We show one model per figure, with one panel per training window: 16 months, 64 months, and expanding. Each panel draws 17 curves, one per industry. A curve gives the six-seed mean of the pooled $R^2$ for that industry in each month. All panels share the vertical axis $[-2.0,1.0]$, and each industry keeps one color throughout. Values below $-2.0$ fall outside the frame, and each caption counts them.
\clearpage
\begin{figure}[h]
\centering
\includegraphics[width=0.9\textwidth]{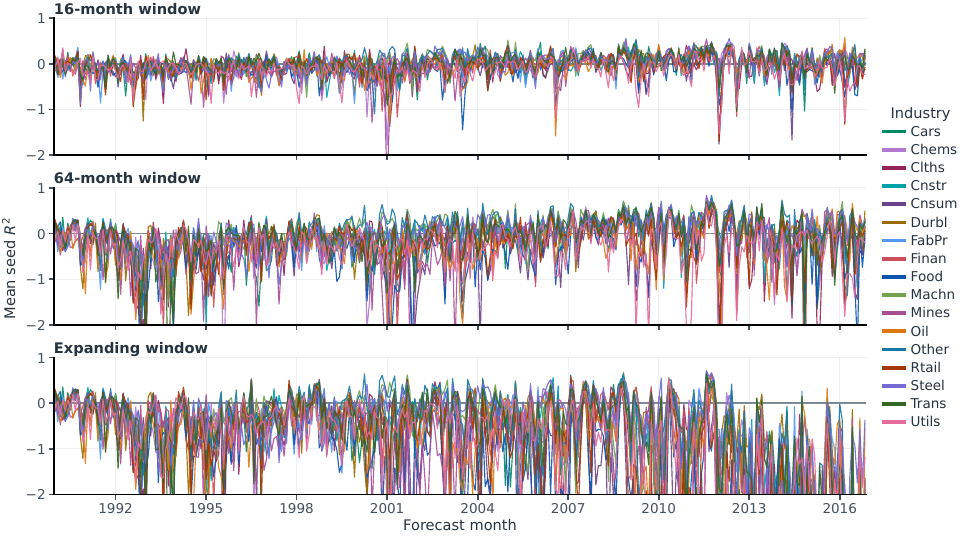}
\caption{Temporal, 60-day context, Flattened ridge: monthly pooled $R^2$ for the 17 industries, one curve each, under the 16-month, 64-month, and expanding windows. Values below $-2.0$ fall outside the frame: 1, 48, and 685 of 5,491 in the three panels.}
\label{fig:stage5-month-industry-temporal60-ridge_alpha1_flat}
\end{figure}

\begin{figure}[h]
\centering
\includegraphics[width=0.9\textwidth]{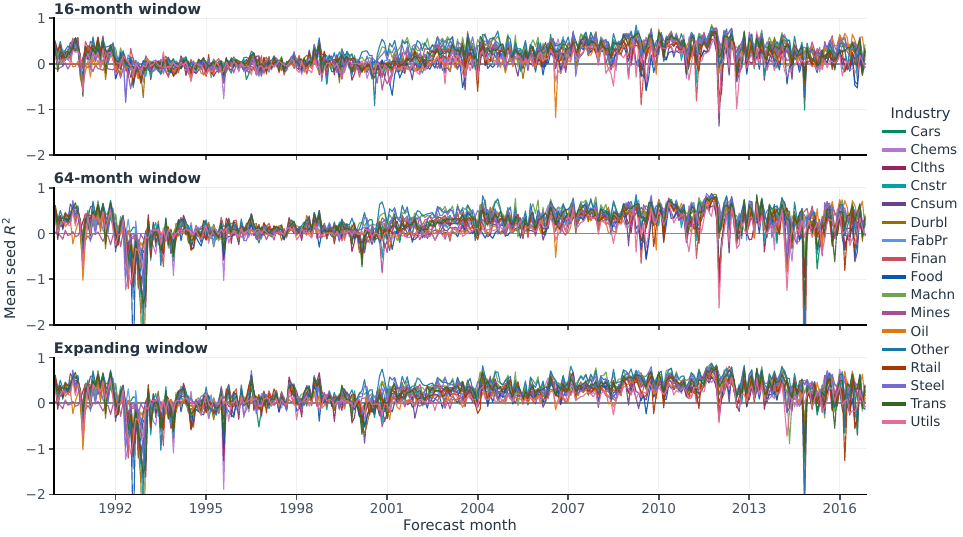}
\caption{Temporal, 60-day context, Flattened XGBoost: monthly pooled $R^2$ for the 17 industries, one curve each, under the 16-month, 64-month, and expanding windows. Values below $-2.0$ fall outside the frame: 0, 5, and 5 of 5,491 in the three panels.}
\label{fig:stage5-month-industry-temporal60-xgboost_flat}
\end{figure}

\begin{figure}[h]
\centering
\includegraphics[width=0.9\textwidth]{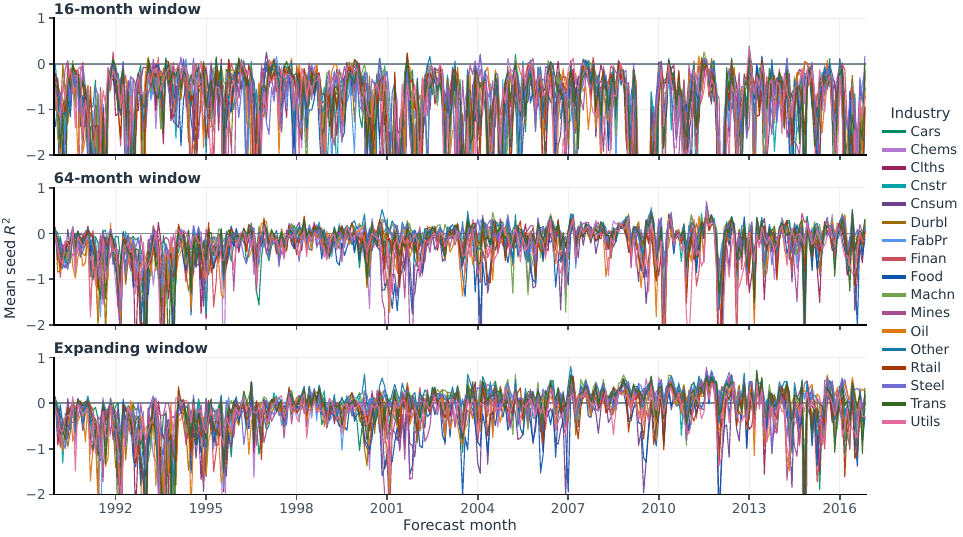}
\caption{Temporal, 60-day context, Small flattened MLP: monthly pooled $R^2$ for the 17 industries, one curve each, under the 16-month, 64-month, and expanding windows. Values below $-2.0$ fall outside the frame: 673, 55, and 53 of 5,491 in the three panels.}
\label{fig:stage5-month-industry-temporal60-mlp_small_anchor_flat}
\end{figure}

\begin{figure}[h]
\centering
\includegraphics[width=0.9\textwidth]{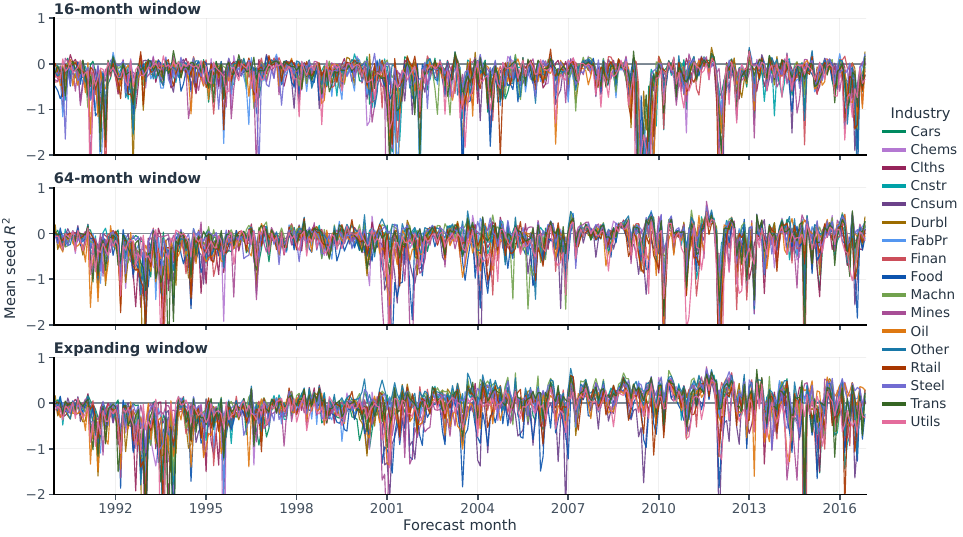}
\caption{Temporal, 60-day context, Tuned flattened MLP: monthly pooled $R^2$ for the 17 industries, one curve each, under the 16-month, 64-month, and expanding windows. Values below $-2.0$ fall outside the frame: 88, 29, and 32 of 5,491 in the three panels.}
\label{fig:stage5-month-industry-temporal60-mlp_tuned_flat}
\end{figure}

\begin{figure}[h]
\centering
\includegraphics[width=0.9\textwidth]{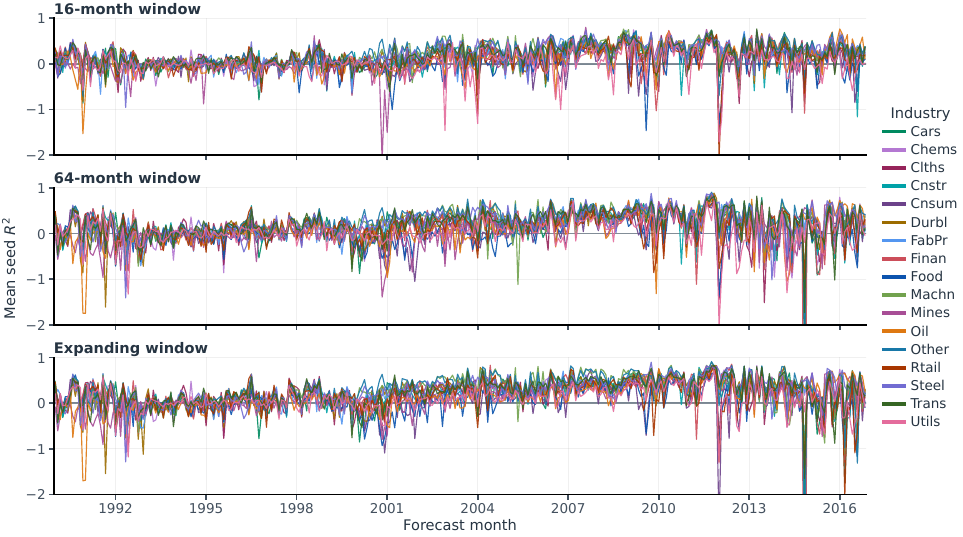}
\caption{Temporal, 60-day context, Vanilla RNN: monthly pooled $R^2$ for the 17 industries, one curve each, under the 16-month, 64-month, and expanding windows. Values below $-2.0$ fall outside the frame: 2, 4, and 5 of 5,491 in the three panels.}
\label{fig:stage5-month-industry-temporal60-rnn_tuned}
\end{figure}

\begin{figure}[h]
\centering
\includegraphics[width=0.9\textwidth]{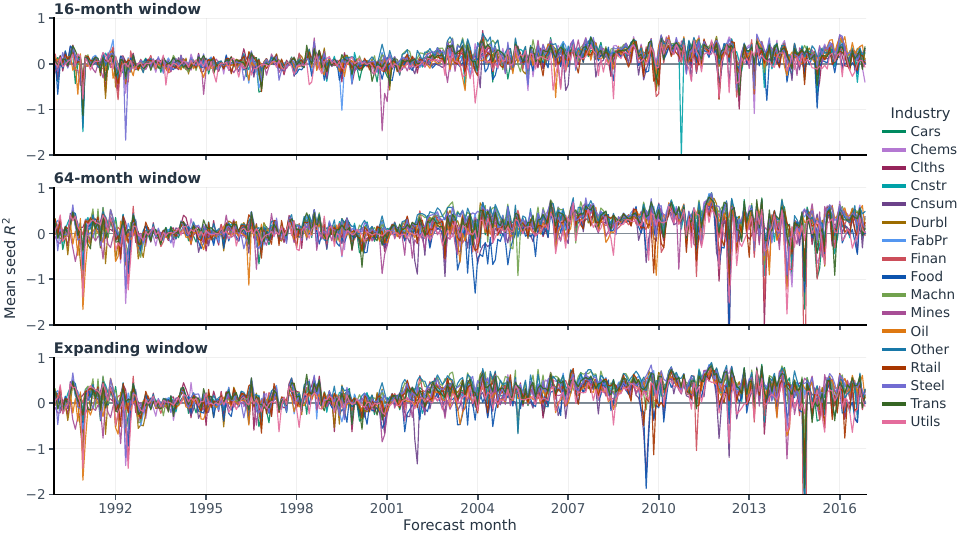}
\caption{Temporal, 60-day context, LSTM: monthly pooled $R^2$ for the 17 industries, one curve each, under the 16-month, 64-month, and expanding windows. Values below $-2.0$ fall outside the frame: 1, 4, and 3 of 5,491 in the three panels.}
\label{fig:stage5-month-industry-temporal60-lstm_tuned}
\end{figure}

\begin{figure}[h]
\centering
\includegraphics[width=0.9\textwidth]{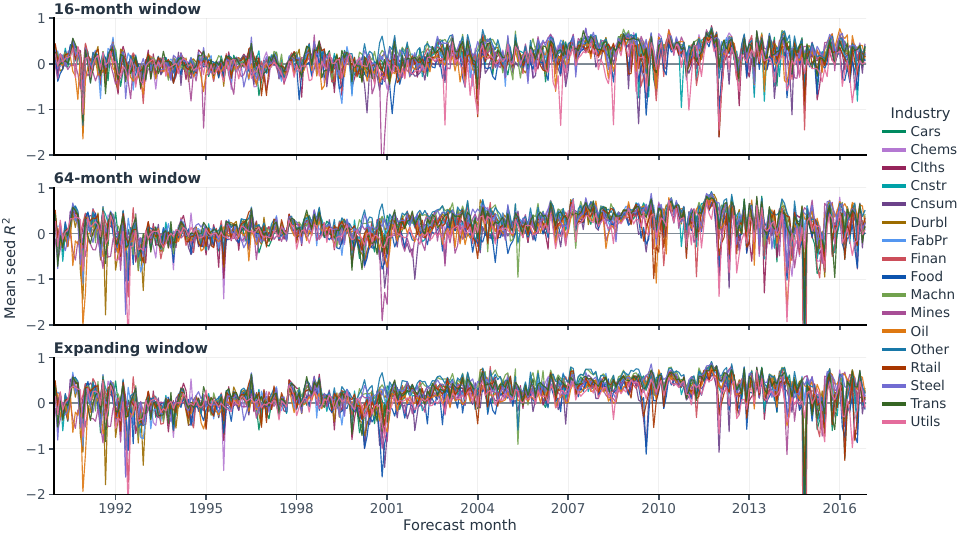}
\caption{Temporal, 60-day context, Small Transformer: monthly pooled $R^2$ for the 17 industries, one curve each, under the 16-month, 64-month, and expanding windows. Values below $-2.0$ fall outside the frame: 1, 6, and 6 of 5,491 in the three panels.}
\label{fig:stage5-month-industry-temporal60-transformer_tuned}
\end{figure}

\clearpage
\subsection{Configuration-level numerical results}
\begingroup\footnotesize\setlength{\tabcolsep}{3.5pt}\renewcommand{\arraystretch}{1.08}
\setlength{\LTcapwidth}{\textwidth}
\begin{longtable}{llrrrrrrrr}
\caption{All 60 reported Stage 5 model-window-context cells. $R^2$ is pooled against zero over 6,784 dates and 17 industries. The interval uses six seed scores; ensemble $R^2$ scores their mean forecast. The final three columns obey $\overline{R^2}=2ag-g^2-u$ up to rounding.}\label{tab:stage5-all}\\
\toprule Model & Window & Mean $R^2$ & SD & CI low & CI high & Ensemble & $2ag$ & $g^2$ & $u$ \\
\midrule\endfirsthead
\toprule Model & Window & Mean $R^2$ & SD & CI low & CI high & Ensemble & $2ag$ & $g^2$ & $u$ \\
\midrule\endhead
\bottomrule\endfoot
\multicolumn{10}{@{}l}{\emph{Tabular}} \\
Ridge, fixed alpha & 16m & $-$0.0909 & 0.0118 & $-$0.1033 & $-$0.0786 & 0.0706 & 0.9726 & 0.9019 & 0.1615 \\
Ridge, fixed alpha & 64m & 0.3256 & 0.0020 & 0.3235 & 0.3277 & 0.3481 & 0.9373 & 0.5892 & 0.0225 \\
Ridge, fixed alpha & Exp. & 0.3386 & 0.0011 & 0.3374 & 0.3397 & 0.3477 & 0.7620 & 0.4143 & 0.0091 \\
Tuned ridge & 16m & 0.3515 & 0.0027 & 0.3487 & 0.3543 & 0.3671 & 0.8296 & 0.4625 & 0.0156 \\
Tuned ridge & 64m & 0.3668 & 0.0010 & 0.3658 & 0.3678 & 0.3767 & 0.8571 & 0.4804 & 0.0099 \\
Tuned ridge & Exp. & 0.3602 & 0.0008 & 0.3593 & 0.3611 & 0.3644 & 0.7084 & 0.3440 & 0.0042 \\
XGBoost & 16m & 0.3316 & 0.0031 & 0.3284 & 0.3348 & 0.3621 & 0.7180 & 0.3559 & 0.0305 \\
XGBoost & 64m & 0.3419 & 0.0037 & 0.3380 & 0.3457 & 0.3619 & 0.7345 & 0.3725 & 0.0201 \\
XGBoost & Exp. & 0.3232 & 0.0029 & 0.3202 & 0.3262 & 0.3392 & 0.6439 & 0.3047 & 0.0160 \\
Small MLP & 16m & 0.2869 & 0.0027 & 0.2841 & 0.2898 & 0.3096 & 0.6003 & 0.2907 & 0.0227 \\
Small MLP & 64m & 0.3231 & 0.0018 & 0.3212 & 0.3251 & 0.3414 & 0.6778 & 0.3364 & 0.0183 \\
Small MLP & Exp. & 0.3301 & 0.0026 & 0.3274 & 0.3328 & 0.3423 & 0.6304 & 0.2881 & 0.0122 \\
Medium MLP & 16m & 0.2786 & 0.0028 & 0.2756 & 0.2816 & 0.3018 & 0.6012 & 0.2994 & 0.0232 \\
Medium MLP & 64m & 0.3189 & 0.0037 & 0.3151 & 0.3228 & 0.3383 & 0.6700 & 0.3317 & 0.0193 \\
Medium MLP & Exp. & 0.3270 & 0.0036 & 0.3232 & 0.3308 & 0.3425 & 0.6272 & 0.2847 & 0.0155 \\
Random forest & 16m & 0.3549 & 0.0033 & 0.3514 & 0.3583 & 0.3705 & 0.7320 & 0.3615 & 0.0156 \\
Random forest & 64m & 0.3374 & 0.0031 & 0.3341 & 0.3407 & 0.3494 & 0.7117 & 0.3623 & 0.0120 \\
Random forest & Exp. & 0.2953 & 0.0121 & 0.2826 & 0.3080 & 0.3079 & 0.6036 & 0.2957 & 0.0126 \\
\midrule
\multicolumn{10}{@{}l}{\emph{Temporal, 20-day context}} \\
Flattened ridge & 16m & 0.1397 & 0.0048 & 0.1346 & 0.1447 & 0.1915 & 0.5584 & 0.3669 & 0.0518 \\
Flattened ridge & 64m & $-$0.2128 & 0.0076 & $-$0.2208 & $-$0.2049 & $-$0.0139 & 0.8094 & 0.8233 & 0.1990 \\
Flattened ridge & Exp. & $-$1.1533 & 0.0323 & $-$1.1871 & $-$1.1194 & $-$0.3308 & 0.8015 & 1.1324 & 0.8225 \\
Flattened XGBoost & 16m & 0.3026 & 0.0033 & 0.2992 & 0.3061 & 0.3341 & 0.6053 & 0.2712 & 0.0315 \\
Flattened XGBoost & 64m & 0.3287 & 0.0027 & 0.3258 & 0.3315 & 0.3485 & 0.6602 & 0.3116 & 0.0198 \\
Flattened XGBoost & Exp. & 0.3208 & 0.0037 & 0.3170 & 0.3247 & 0.3379 & 0.6047 & 0.2668 & 0.0171 \\
Small flattened MLP & 16m & $-$0.0841 & 0.0081 & $-$0.0926 & $-$0.0756 & 0.0200 & 0.2220 & 0.2020 & 0.1041 \\
Small flattened MLP & 64m & 0.0574 & 0.0040 & 0.0532 & 0.0617 & 0.1544 & 0.4273 & 0.2729 & 0.0969 \\
Small flattened MLP & Exp. & 0.1623 & 0.0068 & 0.1552 & 0.1694 & 0.2485 & 0.5360 & 0.2875 & 0.0862 \\
Tuned flattened MLP & 16m & $-$0.0350 & 0.0199 & $-$0.0559 & $-$0.0141 & 0.0406 & 0.1764 & 0.1357 & 0.0756 \\
Tuned flattened MLP & 64m & 0.1160 & 0.0046 & 0.1112 & 0.1209 & 0.1715 & 0.3652 & 0.1937 & 0.0554 \\
Tuned flattened MLP & Exp. & 0.2066 & 0.0027 & 0.2038 & 0.2095 & 0.2607 & 0.5001 & 0.2394 & 0.0540 \\
Vanilla RNN & 16m & 0.2860 & 0.0025 & 0.2834 & 0.2887 & 0.3166 & 0.5951 & 0.2784 & 0.0306 \\
Vanilla RNN & 64m & 0.2987 & 0.0017 & 0.2970 & 0.3005 & 0.3166 & 0.6320 & 0.3154 & 0.0178 \\
Vanilla RNN & Exp. & 0.3184 & 0.0021 & 0.3161 & 0.3206 & 0.3315 & 0.6067 & 0.2752 & 0.0132 \\
LSTM & 16m & 0.2087 & 0.0026 & 0.2059 & 0.2114 & 0.2317 & 0.3958 & 0.1641 & 0.0230 \\
LSTM & 64m & 0.2491 & 0.0027 & 0.2463 & 0.2519 & 0.2673 & 0.5130 & 0.2457 & 0.0182 \\
LSTM & Exp. & 0.2905 & 0.0019 & 0.2886 & 0.2925 & 0.3071 & 0.5439 & 0.2368 & 0.0166 \\
Small Transformer & 16m & 0.2828 & 0.0037 & 0.2789 & 0.2866 & 0.3218 & 0.6275 & 0.3058 & 0.0390 \\
Small Transformer & 64m & 0.2995 & 0.0022 & 0.2972 & 0.3018 & 0.3191 & 0.6437 & 0.3247 & 0.0196 \\
Small Transformer & Exp. & 0.3193 & 0.0020 & 0.3172 & 0.3213 & 0.3361 & 0.6132 & 0.2772 & 0.0168 \\
\midrule
\multicolumn{10}{@{}l}{\emph{Temporal, 60-day context}} \\
Flattened ridge & 16m & 0.0772 & 0.0057 & 0.0712 & 0.0831 & 0.1092 & 0.2775 & 0.1684 & 0.0320 \\
Flattened ridge & 64m & 0.1078 & 0.0057 & 0.1018 & 0.1138 & 0.1654 & 0.5507 & 0.3853 & 0.0576 \\
Flattened ridge & Exp. & $-$0.2835 & 0.0106 & $-$0.2946 & $-$0.2724 & $-$0.0295 & 0.6561 & 0.6856 & 0.2541 \\
Flattened XGBoost & 16m & 0.2812 & 0.0031 & 0.2779 & 0.2845 & 0.3161 & 0.5640 & 0.2479 & 0.0350 \\
Flattened XGBoost & 64m & 0.3223 & 0.0033 & 0.3188 & 0.3257 & 0.3402 & 0.6283 & 0.2880 & 0.0180 \\
Flattened XGBoost & Exp. & 0.3148 & 0.0015 & 0.3132 & 0.3164 & 0.3253 & 0.5388 & 0.2135 & 0.0105 \\
Small flattened MLP & 16m & $-$0.8216 & 0.0599 & $-$0.8845 & $-$0.7588 & $-$0.3891 & 0.1340 & 0.5231 & 0.4325 \\
Small flattened MLP & 64m & $-$0.0124 & 0.0053 & $-$0.0180 & $-$0.0068 & 0.0765 & 0.2813 & 0.2048 & 0.0889 \\
Small flattened MLP & Exp. & 0.0881 & 0.0033 & 0.0846 & 0.0916 & 0.1690 & 0.4152 & 0.2462 & 0.0809 \\
Tuned flattened MLP & 16m & $-$0.2223 & 0.0119 & $-$0.2347 & $-$0.2098 & $-$0.0970 & 0.0760 & 0.1729 & 0.1253 \\
Tuned flattened MLP & 64m & 0.0002 & 0.0035 & $-$0.0034 & 0.0039 & 0.0804 & 0.2552 & 0.1748 & 0.0802 \\
Tuned flattened MLP & Exp. & 0.1158 & 0.0036 & 0.1121 & 0.1196 & 0.1822 & 0.4150 & 0.2328 & 0.0663 \\
Vanilla RNN & 16m & 0.2776 & 0.0028 & 0.2747 & 0.2806 & 0.3123 & 0.5788 & 0.2665 & 0.0347 \\
Vanilla RNN & 64m & 0.2989 & 0.0019 & 0.2969 & 0.3009 & 0.3167 & 0.6275 & 0.3108 & 0.0178 \\
Vanilla RNN & Exp. & 0.3169 & 0.0024 & 0.3144 & 0.3194 & 0.3276 & 0.5841 & 0.2565 & 0.0107 \\
LSTM & 16m & 0.1913 & 0.0038 & 0.1873 & 0.1952 & 0.2153 & 0.3651 & 0.1498 & 0.0241 \\
LSTM & 64m & 0.2472 & 0.0016 & 0.2455 & 0.2489 & 0.2661 & 0.5065 & 0.2404 & 0.0189 \\
LSTM & Exp. & 0.2885 & 0.0009 & 0.2876 & 0.2895 & 0.3042 & 0.5301 & 0.2259 & 0.0156 \\
Small Transformer & 16m & 0.2712 & 0.0052 & 0.2658 & 0.2766 & 0.3191 & 0.6203 & 0.3011 & 0.0479 \\
Small Transformer & 64m & 0.2969 & 0.0007 & 0.2961 & 0.2976 & 0.3160 & 0.6315 & 0.3155 & 0.0191 \\
Small Transformer & Exp. & 0.3195 & 0.0016 & 0.3178 & 0.3212 & 0.3342 & 0.5948 & 0.2606 & 0.0147 \\
\end{longtable}
\endgroup

%% file: references.bib
@article{welch2008comprehensive,
  author = {Welch, Ivo and Goyal, Amit},
  title = {A Comprehensive Look at the Empirical Performance of Equity Premium Prediction},
  journal = {Review of Financial Studies},
  volume = {21}, number = {4}, pages = {1455--1508}, year = {2008},
  doi = {10.1093/rfs/hhm014}
}

@article{campbell2008predicting,
  author = {Campbell, John Y. and Thompson, Samuel B.},
  title = {Predicting Excess Stock Returns Out of Sample: Can Anything Beat the Historical Average?},
  journal = {Review of Financial Studies},
  volume = {21}, number = {4}, pages = {1509--1531}, year = {2008},
  doi = {10.1093/rfs/hhm055}
}

@article{gu2020empirical,
  author = {Gu, Shihao and Kelly, Bryan and Xiu, Dacheng},
  title = {Empirical Asset Pricing via Machine Learning},
  journal = {Review of Financial Studies},
  volume = {33}, number = {5}, pages = {2223--2273}, year = {2020},
  doi = {10.1093/rfs/hhaa009}
}

@misc{zhang2026benchmarking,
  author = {Zhang, Aoxin and Cheng, Yuhan and Leung, Kwanting},
  title = {Benchmarking Deep Time Series Models for Equity Portfolios},
  year = {2026}, howpublished = {arXiv:2606.09420}
}

@misc{goktas2026tempusbench,
  author = {Goktas, Denizalp and Ria{\~n}o-Brice{\~n}o, Gerardo and Abdullah, Alif and others},
  title = {{TempusBench}: An Evaluation Framework for Time-Series Forecasting},
  year = {2026}, howpublished = {arXiv:2604.11529}
}

@misc{capponi2025nonstationarity,
  author = {Capponi, Agostino and Huang, Chengpiao and Sidaoui, J. Antonio and Wang, Kaizheng and Zou, Jiacheng},
  title = {The Nonstationarity-Complexity Tradeoff in Return Prediction},
  year = {2025}, howpublished = {arXiv:2512.23596}
}

@article{dahlhaus1997fitting,
  author = {Dahlhaus, Rainer},
  title = {Fitting Time Series Models to Nonstationary Processes},
  journal = {Annals of Statistics},
  volume = {25}, number = {1}, pages = {1--37}, year = {1997},
  doi = {10.1214/aos/1034276620}
}

@article{pesaran2007selection,
  author = {Pesaran, M. Hashem and Timmermann, Allan},
  title = {Selection of Estimation Window in the Presence of Breaks},
  journal = {Journal of Econometrics},
  volume = {137}, number = {1}, pages = {134--161}, year = {2007},
  doi = {10.1016/j.jeconom.2006.03.010}
}

@misc{huang2023stability,
  author = {Huang, Chengpiao and Wang, Kaizheng},
  title = {A Stability Principle for Learning under Non-Stationarity},
  year = {2023}, howpublished = {arXiv:2310.18304}
}

@inproceedings{han2024assessment,
  author = {Han, Elise and Huang, Chengpiao and Wang, Kaizheng},
  title = {Model Assessment and Selection under Temporal Distribution Shift},
  booktitle = {Proceedings of the 41st International Conference on Machine Learning},
  series = {Proceedings of Machine Learning Research},
  volume = {235}, pages = {17374--17392}, year = {2024}
}

@article{caponnetto2007optimal,
  author = {Caponnetto, Andrea and De Vito, Ernesto},
  title = {Optimal Rates for the Regularized Least-Squares Algorithm},
  journal = {Foundations of Computational Mathematics},
  volume = {7}, pages = {331--368}, year = {2007},
  doi = {10.1007/s10208-006-0196-8}
}

@inproceedings{hsu2012random,
  author = {Hsu, Daniel and Kakade, Sham M. and Zhang, Tong},
  title = {Random Design Analysis of Ridge Regression},
  booktitle = {Proceedings of the 25th Annual Conference on Learning Theory},
  series = {Proceedings of Machine Learning Research},
  volume = {23}, pages = {9.1--9.24}, year = {2012}
}

@article{dobriban2018high,
  author = {Dobriban, Edgar and Wager, Stefan},
  title = {High-Dimensional Asymptotics of Prediction: Ridge Regression and Classification},
  journal = {Annals of Statistics},
  volume = {46}, number = {1}, pages = {247--279}, year = {2018},
  doi = {10.1214/17-AOS1549}
}

@article{bousquet2002stability,
  author = {Bousquet, Olivier and Elisseeff, Andr{\'e}},
  title = {Stability and Generalization},
  journal = {Journal of Machine Learning Research},
  volume = {2}, pages = {499--526}, year = {2002}
}

@article{gneiting2023regression,
  author = {Gneiting, Tilmann and Resin, Johannes},
  title = {Regression Diagnostics Meets Forecast Evaluation: Conditional Calibration, Reliability Diagrams, and Coefficient of Determination},
  journal = {Electronic Journal of Statistics},
  volume = {17}, number = {2}, pages = {3226--3286}, year = {2023},
  doi = {10.1214/23-EJS2180}
}

@article{diebold1995comparing,
  author = {Diebold, Francis X. and Mariano, Roberto S.},
  title = {Comparing Predictive Accuracy},
  journal = {Journal of Business \& Economic Statistics},
  volume = {13}, number = {3}, pages = {253--263}, year = {1995},
  doi = {10.1080/07350015.1995.10524599}
}

@article{hansen2011model,
  author = {Hansen, Peter R. and Lunde, Asger and Nason, James M.},
  title = {The Model Confidence Set},
  journal = {Econometrica},
  volume = {79}, number = {2}, pages = {453--497}, year = {2011},
  doi = {10.3982/ECTA5771}
}

@misc{hewamalage2021look,
  author = {Hewamalage, Hansika and Montero-Manso, Pablo and Bergmeir, Christoph and Hyndman, Rob J.},
  title = {A Look at the Evaluation Setup of the {M5} Forecasting Competition},
  year = {2021}, howpublished = {arXiv:2108.03588}
}

@article{smith2021break,
  author = {Smith, Simon C. and Timmermann, Allan},
  title = {Break Risk},
  journal = {Review of Financial Studies},
  volume = {34}, number = {4}, pages = {2045--2100}, year = {2021},
  doi = {10.1093/rfs/hhaa084}
}

@misc{bejarano2025open,
  author = {Bejarano, Jeremiah and Desai, Viren and Keshava, Kausthub and Kumar, Arsh and Wang, Zixiao and Xu, Vincent Hanyang and Xu, Yangge},
  title = {An Open Benchmark for Evaluating Time Series Forecasting Methods across Financial Markets},
  year = {2025}, howpublished = {Working paper},
  url = {https://jeremybejarano.com/downloads/draft_ftsfr.pdf}
}

@misc{salykaufmann2026deep,
  author = {Saly-Kaufmann, Adir and Wood, Kieran and Peter-Calliess, Jan and Zohren, Stefan},
  title = {Deep Learning for Financial Time Series: A Large-Scale Benchmark of Risk-Adjusted Performance},
  year = {2026}, howpublished = {arXiv:2603.01820}
}

@article{vogt2012nonparametric,
  author = {Vogt, Michael},
  title = {Nonparametric Regression for Locally Stationary Time Series},
  journal = {Annals of Statistics},
  volume = {40}, number = {5}, pages = {2601--2633}, year = {2012},
  doi = {10.1214/12-AOS1043}
}

@article{giraitis2013adaptive,
  author = {Giraitis, Liudas and Kapetanios, George and Price, Simon},
  title = {Adaptive Forecasting in the Presence of Recent and Ongoing Structural Change},
  journal = {Journal of Econometrics},
  volume = {177}, number = {2}, pages = {153--170}, year = {2013},
  doi = {10.1016/j.jeconom.2013.04.003}
}

@article{giacomini2015forecasting,
  author = {Giacomini, Raffaella and Rossi, Barbara},
  title = {Forecasting in Nonstationary Environments: What Works and What Doesn't in Reduced-Form and Structural Models},
  journal = {Annual Review of Economics},
  volume = {7}, pages = {207--229}, year = {2015},
  doi = {10.1146/annurev-economics-080614-115338}
}

@article{giacomini2016model,
  author = {Giacomini, Raffaella and Rossi, Barbara},
  title = {Model Comparisons in Unstable Environments},
  journal = {International Economic Review},
  volume = {57}, number = {2}, pages = {369--392}, year = {2016},
  doi = {10.1111/iere.12161}
}

@inproceedings{cheung2019learning,
  author = {Cheung, Wang Chi and Simchi-Levi, David and Zhu, Ruihao},
  title = {Learning to Optimize under Non-Stationarity},
  booktitle = {Proceedings of the 22nd International Conference on Artificial Intelligence and Statistics},
  year = {2019},
  url = {https://arxiv.org/abs/1810.03024}
}

@article{yuan2020trading,
  author = {Yuan, Jianjun and Lamperski, Andrew},
  title = {Trading-Off Static and Dynamic Regret in Online Least-Squares and Beyond},
  journal = {Proceedings of the AAAI Conference on Artificial Intelligence},
  volume = {34}, number = {4}, pages = {6712--6719}, year = {2020},
  doi = {10.1609/aaai.v34i04.6149}
}

@inproceedings{zhao2020simple,
  author = {Zhao, Peng and Zhang, Lijun and Jiang, Yuan and Zhou, Zhi-Hua},
  title = {A Simple Approach for Non-stationary Linear Bandits},
  booktitle = {Proceedings of the Twenty Third International Conference on Artificial Intelligence and Statistics},
  series = {Proceedings of Machine Learning Research},
  volume = {108}, pages = {746--755}, year = {2020}
}

@inproceedings{yang2016tracking,
  author = {Yang, Tianbao and Zhang, Lijun and Jin, Rong and Yi, Jinfeng},
  title = {Tracking Slowly Moving Clairvoyant: Optimal Dynamic Regret of Online Learning with True and Noisy Gradient},
  booktitle = {Proceedings of the 33rd International Conference on Machine Learning},
  series = {Proceedings of Machine Learning Research},
  volume = {48}, pages = {449--457}, year = {2016}
}

@inproceedings{zhao2021improved,
  author = {Zhao, Peng and Zhang, Lijun},
  title = {Improved Analysis for Dynamic Regret of Strongly Convex and Smooth Functions},
  booktitle = {Proceedings of the 3rd Conference on Learning for Dynamics and Control},
  series = {Proceedings of Machine Learning Research},
  volume = {144}, pages = {48--59}, year = {2021}
}

@inproceedings{baby2023second,
  author = {Baby, Dheeraj and Wang, Yu-Xiang},
  title = {Second Order Path Variationals in Non-Stationary Online Learning},
  booktitle = {Proceedings of the 26th International Conference on Artificial Intelligence and Statistics},
  series = {Proceedings of Machine Learning Research},
  volume = {206}, pages = {9024--9075}, year = {2023}
}

@misc{baby2025adaptive,
  author = {Baby, Dheeraj and Tang, Yifei and Nguyen, Hieu Duy and Wang, Yu-Xiang and Pyati, Rohit},
  title = {Adaptive Estimation and Learning under Temporal Distribution Shift},
  year = {2025}, howpublished = {arXiv:2505.15803}
}

@incollection{mincer1969evaluation,
  author = {Mincer, Jacob A. and Zarnowitz, Victor},
  title = {The Evaluation of Economic Forecasts},
  booktitle = {Economic Forecasts and Expectations: Analysis of Forecasting Behavior and Performance},
  publisher = {National Bureau of Economic Research}, year = {1969},
  url = {https://www.nber.org/books-and-chapters/economic-forecasts-and-expectations-analysis-forecasting-behavior-and-performance/evaluation-economic-forecasts}
}

@book{theil1966applied,
  author = {Theil, Henri},
  title = {Applied Economic Forecasting},
  publisher = {North-Holland}, year = {1966}
}

@article{waghmare2026proper,
  author = {Waghmare, Kartik and Ziegel, Johanna},
  title = {Proper Scoring Rules for Estimation and Forecast Evaluation},
  journal = {Annual Review of Statistics and Its Application},
  volume = {13}, pages = {271--296}, year = {2026},
  doi = {10.1146/annurev-statistics-042424-050626}
}

@article{diebold1987structural,
  author = {Diebold, Francis X. and Pauly, Peter},
  title = {Structural Change and the Combination of Forecasts},
  journal = {Journal of Forecasting},
  volume = {6}, number = {1}, pages = {21--40}, year = {1987},
  doi = {10.1002/for.3980060103}
}

@article{elliott2004optimal,
  author = {Elliott, Graham and Timmermann, Allan},
  title = {Optimal Forecast Combinations under General Loss Functions and Forecast Error Distributions},
  journal = {Journal of Econometrics},
  volume = {122}, number = {1}, pages = {47--79}, year = {2004},
  doi = {10.1016/j.jeconom.2003.10.019}
}

@misc{chen2020timevarying,
  author = {Chen, Bin and Maung, Kenwin},
  title = {Time-varying Forecast Combination for High-Dimensional Data},
  year = {2020}, howpublished = {arXiv:2010.10435}
}

@article{white2000reality,
  author = {White, Halbert},
  title = {A Reality Check for Data Snooping},
  journal = {Econometrica},
  volume = {68}, number = {5}, pages = {1097--1126}, year = {2000},
  doi = {10.1111/1468-0262.00152}
}

@article{hansen2005test,
  author = {Hansen, Peter R.},
  title = {A Test for Superior Predictive Ability},
  journal = {Journal of Business \& Economic Statistics},
  volume = {23}, number = {4}, pages = {365--380}, year = {2005},
  doi = {10.1198/073500105000000063}
}

@article{giacomini2006tests,
  author = {Giacomini, Raffaella and White, Halbert},
  title = {Tests of Conditional Predictive Ability},
  journal = {Econometrica},
  volume = {74}, number = {6}, pages = {1545--1578}, year = {2006},
  doi = {10.1111/j.1468-0262.2006.00718.x}
}

@inproceedings{chen2016xgboost,
  author = {Chen, Tianqi and Guestrin, Carlos},
  title = {{XGBoost}: A Scalable Tree Boosting System},
  booktitle = {Proceedings of the 22nd ACM SIGKDD International Conference on Knowledge Discovery and Data Mining},
  pages = {785--794}, year = {2016},
  doi = {10.1145/2939672.2939785}
}

@inproceedings{cortes2019regularized,
  author = {Cortes, Corinna and Mohri, Mehryar and Storcheus, Dmitry},
  title = {Regularized Gradient Boosting},
  booktitle = {Advances in Neural Information Processing Systems},
  volume = {32}, year = {2019}
}

@inproceedings{chen2020recurrent,
  author = {Chen, Minshuo and Li, Xingguo and Zhao, Tuo},
  title = {On Generalization Bounds of a Family of Recurrent Neural Networks},
  booktitle = {Proceedings of the Twenty Third International Conference on Artificial Intelligence and Statistics},
  series = {Proceedings of Machine Learning Research},
  volume = {108}, pages = {1233--1243}, year = {2020}
}

@inproceedings{trauger2024transformers,
  author = {Trauger, Jacob and Tewari, Ambuj},
  title = {Sequence Length Independent Norm-Based Generalization Bounds for Transformers},
  booktitle = {Proceedings of the 27th International Conference on Artificial Intelligence and Statistics},
  series = {Proceedings of Machine Learning Research},
  volume = {238}, pages = {1405--1413}, year = {2024}
}

@article{stambaugh1999predictive,
  author = {Stambaugh, Robert F.},
  title = {Predictive Regressions},
  journal = {Journal of Financial Economics},
  volume = {54}, number = {3}, pages = {375--421}, year = {1999},
  doi = {10.1016/S0304-405X(99)00041-0}
}

@article{chen2024deep,
  author = {Chen, Luyang and Pelger, Markus and Zhu, Jason},
  title = {Deep Learning in Asset Pricing},
  journal = {Management Science}, volume = {70}, pages = {714--750}, year = {2024}
}

@article{fama1993common,
  author = {Fama, Eugene F. and French, Kenneth R.},
  title = {Common Risk Factors in the Returns on Stocks and Bonds},
  journal = {Journal of Financial Economics}, volume = {33}, number = {1}, pages = {3--56}, year = {1993}
}

@article{fama1997industry,
  author = {Fama, Eugene F. and French, Kenneth R.},
  title = {Industry Costs of Equity},
  journal = {Journal of Financial Economics}, volume = {43}, number = {2}, pages = {153--193}, year = {1997}
}
